\documentclass[lettersize,journal]{IEEEtran}
\usepackage{amsmath,amsfonts}
\usepackage{algorithmic}
\usepackage{algorithm}
\usepackage{array}

\usepackage{textcomp}
\usepackage{stfloats}
\usepackage{url}
\usepackage{verbatim}
\usepackage{graphicx}
\usepackage{cite}
\usepackage{multirow}
\usepackage{epsfig}
\usepackage{graphicx}
\usepackage{amssymb}
\usepackage{multirow}
\usepackage{color}
\usepackage[table]{xcolor}
\usepackage{lipsum}
\usepackage{booktabs}
\usepackage{colortbl}
\usepackage{makecell}
\usepackage{subfigure}
\begin{document}

\title{MIMONet: Multi-scale Input and Multi-scale Output Network for Salient Object Detection}

\author{Zhaojian Yao, 
        Wei Gao,~\IEEEmembership{Senior Member,~IEEE,}
        Tiesong Zhao, 
        Hui Yuan,
        and Sam Kwong,~\IEEEmembership{Fellow,~IEEE}

\thanks{This work was supported by Natural Science Foundation of China (62271013, 62031013), Shenzhen Fundamental Research Program (GXWD20201231165807007-20200806163656003), Shenzhen Science and Technology Plan Basic Research Project (JCYJ20190808161805519), and was sponsored by CCF-Lenovo Blue Ocean Research Fund (CCF-Lenovo OF 202208). (\textit{Corresponding author: Wei Gao.})}

\thanks{Zhaojian Yao is with the School of Electronic and Computer Engineering, Shenzhen Graduate School, Peking University, Shenzhen 518055, China, and also with the Peng Cheng Laboratory, Shenzhen 518066, China (Email: yaozj@pku.edu.cn).}

\thanks{Wei Gao is with the School of Electronic and Computer Engineering, Shenzhen Graduate School, Peking University, Shenzhen 518055, China (Email: gaowei262@pku.edu.cn).}

\thanks{Tiesong Zhao is with the Fujian Key Laboratory for Intelligent Processing and Wireless Transmission of Media Information, College of Physics and Information Engineering, Fuzhou University, Fuzhou 350108, China (Email: t.zhao@fzu.edu.cn).}

\thanks{Hui Yuan is with the School of Control Science and Engineering, Shandong University, Jinan, China (Email: huiyuan@sdu.edu.cn).}

\thanks{Sam Kwong is with the Department of Computer Science, City University of Hong Kong, Kowloon, Hong Kong (Email: cssamk@cityu.edu.hk).}


}


\markboth{Journal of \LaTeX\ Class Files,~Vol.~14, No.~8, August~2021}%
{Shell \MakeLowercase{\textit{et al.}}: Bare Demo of IEEEtran.cls for IEEE Journals}


\maketitle

\begin{abstract}
The existing methods for saliency detection task focus on the application of multi-level features, aiming to take advantage of the respective strengths of high- and low-level features. However, because the inputs of these models are single-size images, their multi-level features have difficulty in learning the knowledge of size variations of salient objects. Object-scale variation learning has great potential for detecting multi-scale objects, which has not been fully explored by existing methods. To improve the recognition ability of a model for objects with different sizes, we are inspired by the image pyramid to propose a Multi-scale Input and Multi-scale Output Network (MIMONet). In MIMONet, we extract multi-level features for three images with different resolutions to form three encoder branches, and information will be exchanged between the branches. The advantage of this approach is that the features of one branch can learn the knowledge of target size variation from the features of the other two branches. In addition, we design a Multi-scale Perception (MSP) module, in which the input feature layer is divided into several sub-layers with different resolutions. Capturing the multi-level structure information of the objects in these sub-layers can make the objects more fully perceived. For network training, we propose a Joint Saliency Loss (JSL), which can constrain multiple saliency maps output by the network to identify the same foreground objects, and induce their boundaries to be preserved clearly. Experimental results show that MIMONet has stronger detection capabilities and harvests better evaluation scores on multiple datasets compared to existing models. The code of our model will be released.
\end{abstract}

\begin{IEEEkeywords}
Salient object detection, multi-scale inputs and multi-scale outputs, multi-scale perception, joint saliency loss.
\end{IEEEkeywords}

\section{Introduction}
\IEEEPARstart{T}{he} scale-variant nature of salient objects has always been an obstacle that plagues detection techniques. On the one hand, the objects are in different scenarios, and their scales are prone to be different. On the other hand, even when in the same scenario, their shapes and sizes may change when they move. Because salient objects have various sizes, it is a challenging task to locate them accurately. To improve the performance of an algorithm, some researchers have introduced depth{~\cite{gao2021unified,fu2022siamese} and thermal~\cite{tu2022rgbt,liao2022cross} information into saliency detection in addition to using RGB information. However, due to the limitation of device, not all RGB images are accompanied by their depth maps or thermal maps of the corresponding scenes. Therefore, the saliency detection for a single RGB image is still a research direction worthy of in-depth exploration. 

\begin{figure*}[htbp]
	\centering
	\includegraphics[width=1\linewidth]{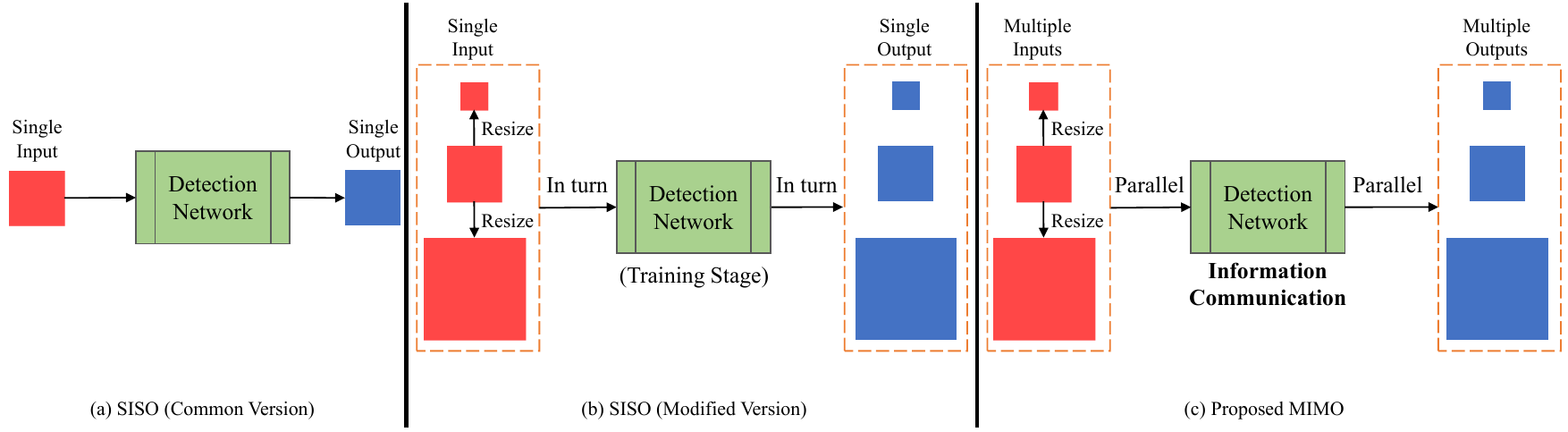}
	\caption{
		Categorisation of saliency detection models according to the number of inputs and outputs. (a) The original image is taken as input, and the detection network outputs its saliency map. (b) In the training stage, the original image is down-sampled and up-sampled, and multiple images with different resolutions are generated. These images are fed into the SISO model in turn and the network processes each image individually. Then, their corresponding saliency maps are output in turn. This framework is simply an improvement on the training strategy of the original SISO model. (c) The original image is resized to produce images of different resolutions. These images are fed into the detection network in parallel. The network simultaneously infers their saliency and outputs their saliency maps in parallel. During the inference process, multi-scale information collaborates with each other.
	}
	\label{fig:version}
\end{figure*}

The existing CNN-based saliency detection models take a single-scale RGB image as input and extract its multi-level features. Among these features, high-level features often represent semantic information, which plays an important role in distinguishing different objects. As for low-level features, they usually contain rich detail information, which is indispensable for object segmentation tasks. Therefore, a common approach of existing work is to exploit the advantages of both types of features and combine them to form complementary features for salient object detection. The end of these models usually outputs a single-scale saliency map as the prediction result. In general, the existing models~\cite{zhao2019egnet,li2022complementarity,chen2020global,liu2022dna,zhou2023edge} can be categorized into a Single-scale Input and Single-scale Output (SISO) framework, as shown in Fig.~\ref{fig:version}(a). However, because the backbone processes a single RGB image, the generated multi-level features can only express the information about single-size objects, and fail to explore the potential change pattern of the object scale. Some researchers~\cite{wei2020f3net,wu2019stacked,wu2022edn,wu2021decomposition} take the scaling of training samples as a fashion of data augmentation. Thus, a modified version can be obtained by training the original SISO model in turn using multi-scale images, as can be seen in Fig.~\ref{fig:version}(b). This training strategy helps the model to identify targets of more sizes to a certain extent. However, as images of different resolutions are fed sequentially into the model, features of different resolutions lack the opportunity to communicate with each other, which makes the model unable to effectively explore the relationship between them. Overall, in the existing SISO models, due to the insufficient exchange between the information of different scales, the experience of object size change has not been fully developed. If a network has learned rich relevant experience, it can still detect similar objects of different sizes when faced with new scenarios. Therefore, it is necessary to consider the exchange and relationship between multi-scale information in network design.

\begin{figure}[htbp]
	\centering
	\subfigure[Input]{
		\begin{minipage}[b]{0.19\linewidth}
			\includegraphics[width=1\linewidth]{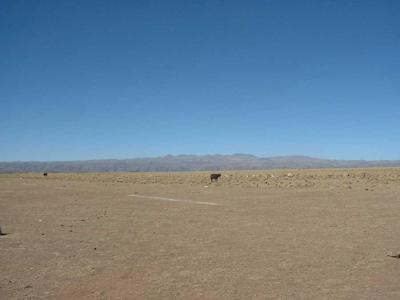}\vspace{1.2pt}
			\includegraphics[width=1\linewidth]{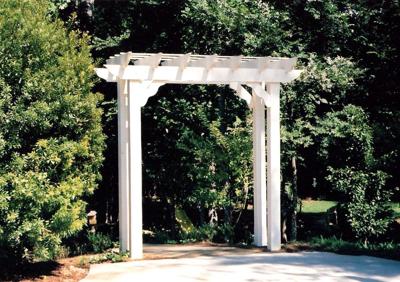}\vspace{1.2pt}
			\includegraphics[width=1\linewidth]{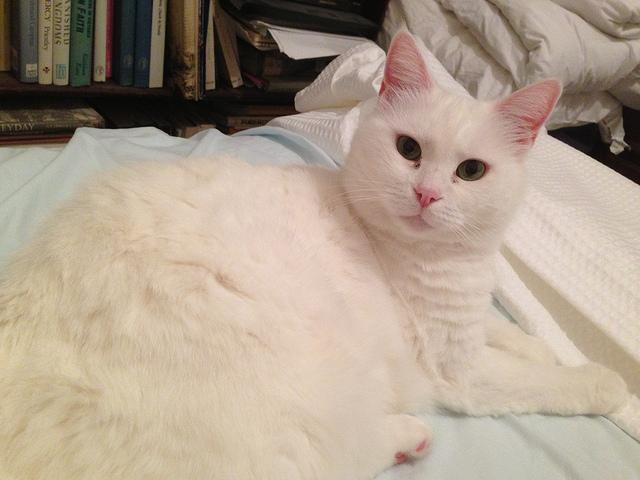}\vspace{1.2pt}
	\end{minipage}}
	\hspace{-2ex}
	\subfigure[GT]{
		\begin{minipage}[b]{0.19\linewidth}
			\includegraphics[width=1\linewidth]{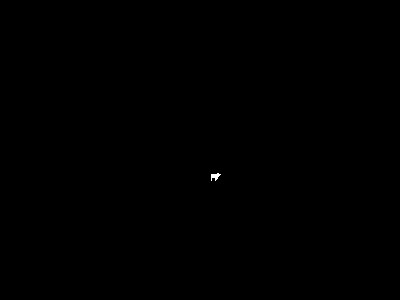}\vspace{1.2pt}
			\includegraphics[width=1\linewidth]{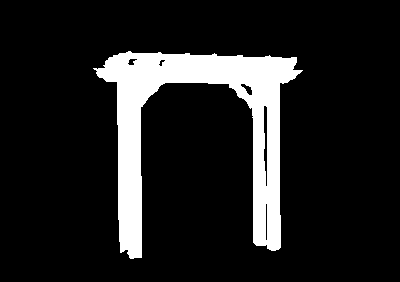}\vspace{1.2pt}
			\includegraphics[width=1\linewidth]{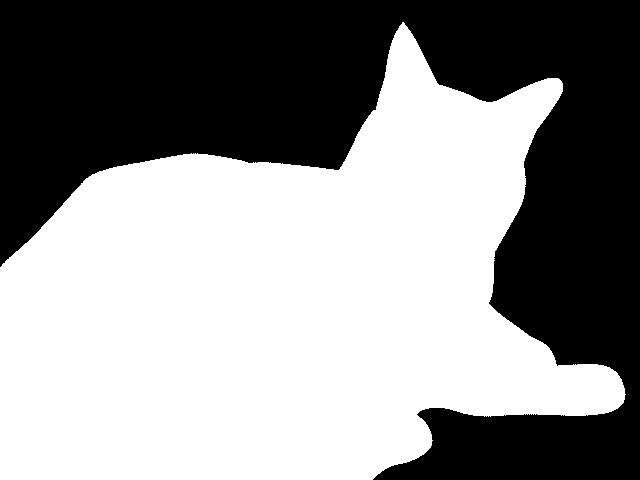}\vspace{1.2pt}
	\end{minipage}}
	\hspace{-2ex}
	\subfigure[MIMONet]{
		\begin{minipage}[b]{0.19\linewidth}
			\includegraphics[width=1\linewidth]{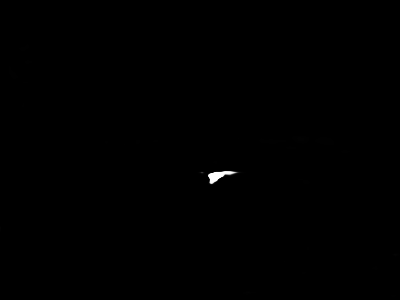}\vspace{1.2pt}
			\includegraphics[width=1\linewidth]{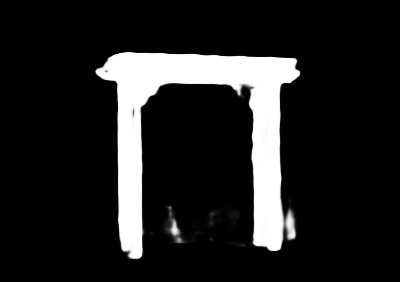}\vspace{1.2pt}
			\includegraphics[width=1\linewidth]{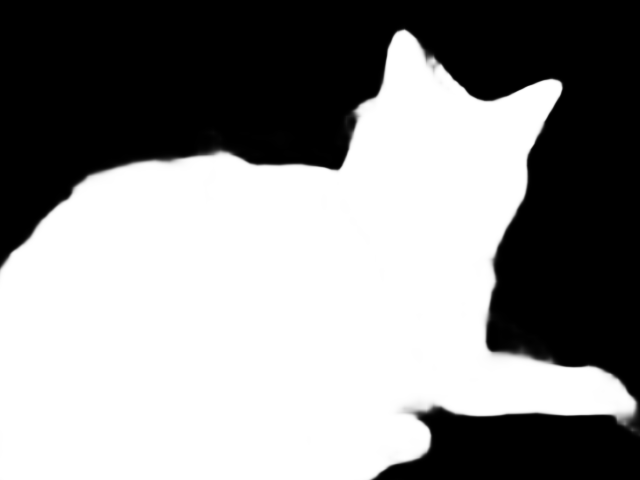}\vspace{1.2pt}
	\end{minipage}}
	\hspace{-2ex}
	\subfigure[EDN~\cite{wu2022edn}]{
		\begin{minipage}[b]{0.19\linewidth}
			\includegraphics[width=1\linewidth]{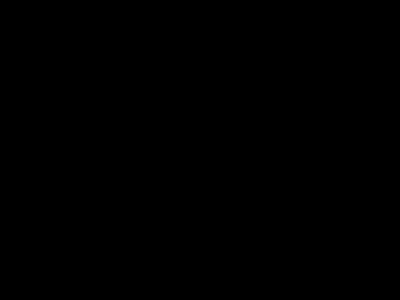}\vspace{1.2pt}
			\includegraphics[width=1\linewidth]{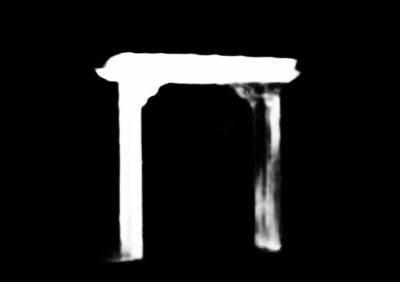}\vspace{1.2pt}
			\includegraphics[width=1\linewidth]{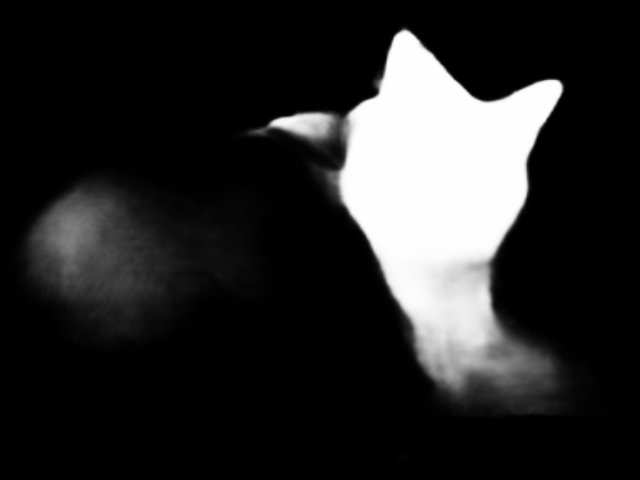}\vspace{1.2pt}
	\end{minipage}}
	\hspace{-2ex}
	\subfigure[EGN~\cite{zhao2019egnet}]{
		\begin{minipage}[b]{0.19\linewidth}
			\includegraphics[width=1\linewidth]{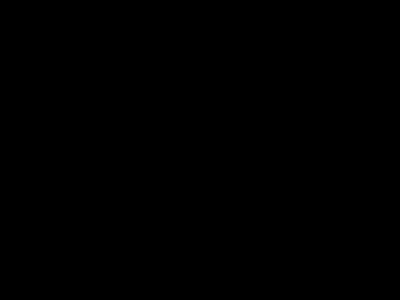}\vspace{1.2pt}
			\includegraphics[width=1\linewidth]{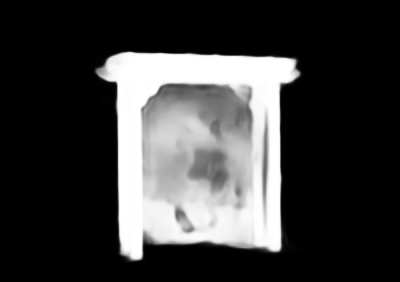}\vspace{1.2pt}
			\includegraphics[width=1\linewidth]{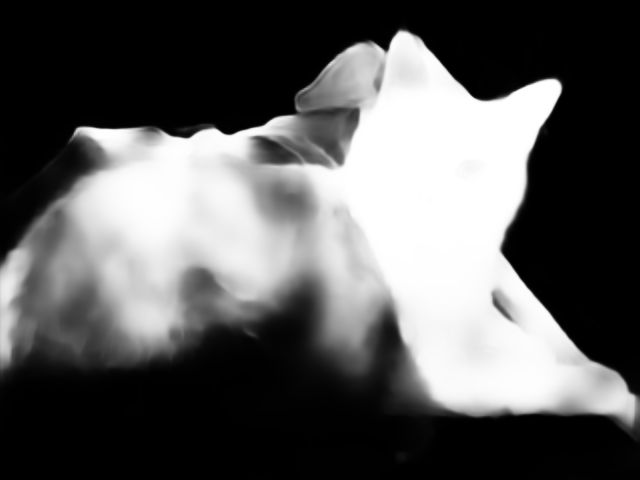}\vspace{1.2pt}
	\end{minipage}}
	\caption{Saliency maps generated by different models. MIMONet is our proposed network. EDN and EGN belong to the modified and original versions of SISO respectively.}
	\label{fig:sample}
\end{figure}

\begin{figure}[htbp]
	\centering
	\includegraphics[width=1\linewidth]{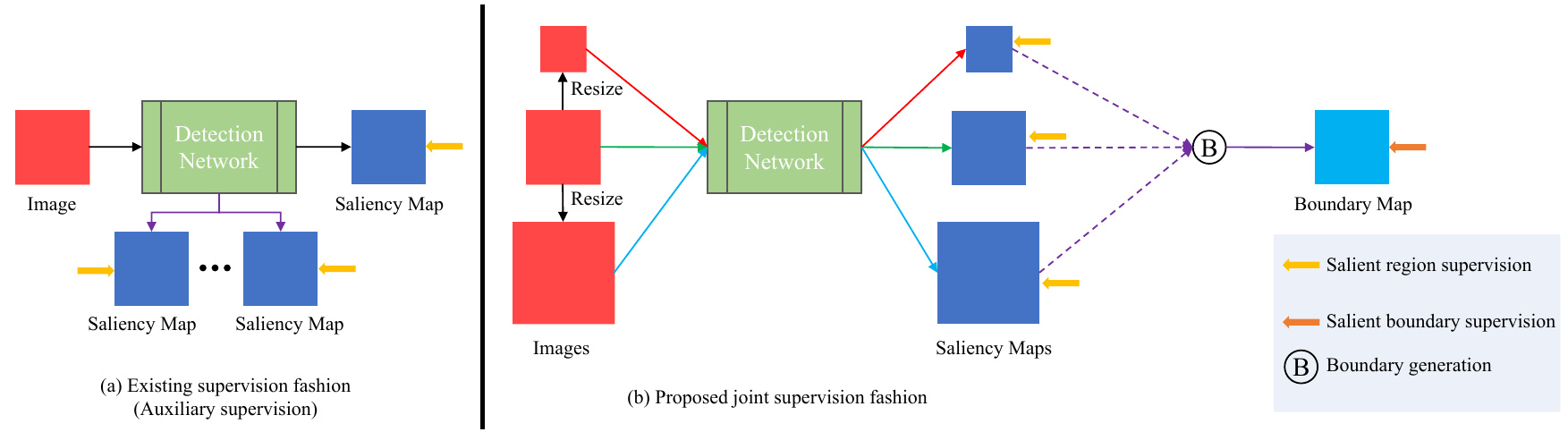}
	\caption{
		Comparison of supervision fashions for saliency map. (a) Some SISO models impose additional constraints on specific feature layers during the training process, and output several saliency maps. There is no correlation between these saliency maps. (b) Our MIMONet outputs multiple saliency maps. Enhancing their correlation can further ensure that the same salient regions are predicted.
	}
	\label{fig:loss}
\end{figure}

To solve the shortcomings of SISO framework, we propose a Multi-scale Input and Multi-scale Output (MIMO) framework, in which multi-scale images are input in parallel, and after feature interaction, their prediction results are output in parallel, as shown in Fig.~\ref{fig:version}(c). Specifically, we create an MIMO network (MIMONet) to identify foreground objects with variable sizes. Ideally, the complementarity and cooperation of multiple data usually yields more reliable inference results than single data. A single RGB image contains limited information, but by scaling its size, more data is generated, which can further provide some promising clues. Therefore, we combine multi-scale images for saliency prediction instead of utilizing single-scale image. The difference from existing models is that the inputs of our model are three images with different resolutions, and the output is the result of foreground prediction for each of them. The comparison of the detection results of various models is shown in Fig.~\ref{fig:sample}. It can be seen that MIMONet segments the salient objects more clearly than the other two models. In the encoding stage of MIMONet, multi-scale images are simultaneously input to the same backbone, where the backbone parameters are shared. After acquiring the multi-level features of each image, the multi-scale features of the same level will exchange information. After complementing information across scales, multi-level features not only contain semantic and detail cues, but also collect cues about the size variation of objects. In the decoding stage, fusing multi-level features can exploit these cues simultaneously to more accurately detect the salient objects in the scenes. To completely segment objects with different sizes, we also design a Multi-Scale Perception (MSP) module to refine a feature layer. It constructs multiple branches to extract the multi-scale structural information of the objects. Combining the processing results of multiple branches, the salient regions will receive more attention. When the MSP module is embedded into MIMONet, the inference capability of the network is further enhanced.

The optimization of a network involves the definition or design of the loss function, and therefore the loss function affects the detection performance of the network. Some existing methods~\cite{wei2020f3net,zhang2021auto} constrain some specific feature layers to predict saliency maps, as shown in Fig.~\ref{fig:loss}. This way is considered as auxiliary supervision. However, all these saliency maps are independent of each other. The existing methods do not impose joint constraints on them, which makes the salient regions in the prediction results may be different. The ability of the network to detect truly salient objects is easily weakened by the lack of effective regional consistency supervision. In MIMONet, multiple saliency maps with different resolution are output. To bootstrap these saliency maps to contain the same salient regions, we introduce a simple Joint Saliency Loss (JSL). We first calculate the common salient regions for all saliency maps. Then several different foreground-background separation maps are generated based on the common salient regions. Combining these foreground-background separation maps results in a common boundary map. Finally, we directly measure the difference between the common boundary map and the boundary label. There are two benefits of using JSL to train the network. On the one hand, constraining multiple saliency maps to predict common regions better enables the network to learn features that are relevant to the true targets. On the other hand, since JSL also calculates the extent of loss of boundary information, it can make the predicted salient regions have clear boundaries.

In general, the contributions of our work to the salient object detection task are summarized as follows:

\begin{itemize}
\item We propose a multi-scale input and multi-scale output network (MIMONet). The parallel input of multi-resolution images provides the basis for learning object scale variations. In the network, we introduce a multi-scale feature exchange mechanism that breaks through the limitations of existing SISO models that cannot establish cooperation between multi-scale features. Multiple outputs are used to further facilitate the network to accurately collect salient cues from multiple inputs.

\item We build a multi-scale perception (MSP) module for feature enhancement. The module is composed of three branches, each of which contains different scale transformations of the feature layer. Therefore, multi-level structural information of the salient regions can be extracted from the three resolution features. We deploy multiple MSP modules in MIMONet to enhance its perception of targets with various sizes.

\item We propose a simple joint saliency loss to supervise multiple outputs of the network, guaranteeing that they predict the same salient regions. In addition to making these outputs locate the consistent targets, this loss also guides them to retain the object boundaries. Therefore, with the aid of JSP, the network can segment accurate and complete foreground objects.

\end{itemize}

\section{Related Work}
Recently, the achievements of salient object detection technology have been gradually improved. The fusion of multi-level features and network supervision are two important technical schemes that affect the performance of the models. We discuss the existing work in view of these two aspects.

\subsection{Multi-level feature fusion} 
For saliency detection, researchers adopt the fully convolutional neural networks (FCN)~\cite{7298965} for pixel-wise prediction. The common method is to combine multi-level features to simultaneously leverage complementary information between high-level and low-level features. Zhang et al.~\cite{zhang2017amulet} directly merge the five-layer features and scale them to different resolutions, and then predict a saliency map based on the combined features. Hou et al.~\cite{hou2017deeply} apply short connections to fuse high-level and low-level features, and the final saliency map is generated by combining several saliency maps. Hu et al.~\cite{hu18recurrently} directly integrate the multi-layer features and then use the fused features to iteratively refine the features at each layer. Zhang et al.~\cite{zhang2018bi} built a bi-directional message passing model to convey information between the high-level and low-level features in different directions. Liu et al.~\cite{liu2018picanet} integrate global context and multiscale local contexts for salient object detection. Wu et al.~\cite{wu2019cascaded} propose a partial decoder to fuse high-level and low-level features and use a holistic attention module to predict saliency maps. In order to identify targets with variable scales, Pang et al.~\cite{pang2020multi} propose an aggregate interaction module to fuse adjacent feature layers. Wu et al.~\cite{wu2022edn} apply an extremely downsampled block to high-level features to enable them to learn global knowledge. Zhuge et al.~\cite{zhuge2022salient} use an integrity channel enhancement to mine the relationship between multi-scale features. Zhang et al.~\cite{zhang2021auto} adopt the strategy of Natural Architecture Search to fuse multi-scale features.

These methods only consider the fusion of low- and high-level features of a single-scale image, and do not pay enough attention to the importance of the multi-size information about the targets. The lack of capturing relevant information reduces the flexibility of the model when the size of the target changes. Unlike them, we introduce multi-size object information at the encoding stage, and impose saliency constraints on each size information at the decoding stage. In this way, the cues of size changes can be mined, and multi-size object recognition is facilitated.

\subsection{Network supervision}
Recently, some boundary-aware methods have been proposed to keep the inherent boundaries of salient objects. Luo et al.~\cite{luo2017non} extract the object boundaries from the detected saliency map, and then calculate the Intersection-over-Union (IoU) loss between the predicted boundaries and the ground truth boundaries. Similarly, Feng et al.~\cite{feng2019attentive} design a boundary loss function to measure the difference between the extracted boundaries and the ground truth boundaries. Su et al.~\cite{su2019selectivity} design a three-stream network where the boundary localization stream is used to detect the boundaries of salient objects. Zhou et al.~\cite{zhou2020interactive} define an adaptive contour loss to guide the network to enhance learning of hard examples. Zhao et al.~\cite{zhao2019egnet} utilize local edge and global location information to generate the salient edge features. Wu et al.~\cite{wu2019stacked} build a cross refinement unit to bi-directionally transfer information from both salient region and boundary detection tasks. In addition to manufacturing boundary labels for boundary constraints, some models also use other labels for training supervision. Liu et al.~\cite{liu2020dynamic} introduce skeleton labels for training the network. Inspired by this work, Wu et al.~\cite{wu2021decomposition} also propose a decomposition and completion Network to predict salient regions, edges, and skeletons. Moreover, Wei et al.~\cite{wei2020label} use saliency, detail, and body labels to train their network. Besides, some specific loss functions are also designed to calculate the difference between the saliency map and the ground truth. For example, Li et al.~\cite{li2021salient} introduce a structural similarity loss to calibrate region-level saliency values. Yang et al.~\cite{yang2021progressive} propose a progressive self-guided loss as the auxiliary training supervision of the network. Wei et al.~\cite{wei2020f3net} use a pixel position aware loss to focus on local details of the salient regions.

During the training process, some models output multiple saliency maps for network supervision. However, the existing methods only focus on the individual optimization of each saliency map, but fail to consider the commonality between saliency maps. Due to the lack of commonality constraints, the output results may not uniformly emphasize the same complete salient regions, resulting in the network not being more effectively optimized. Realizing their shortcomings, we add the commonality constraint of the saliency maps to network supervision, which is beneficial to further motivate them to jointly predict the same salient regions, and enable the network to collect more accurate saliency cues.

\section{Proposed method}

\subsection{Multi-scale Perception Module}
The sizes of salient objects are variable. For a feature layer, it is necessary to have the ability to perceive multi-scale object information. To make the feature layer focus on the potential size change of the targets, we design an MSP module, as shown in Fig.~\ref{fig:MSPM}. This module consists of three branches, each of which handles different scales of features. The first branch deals with low-resolution features and is used to simulate the case where the object size becomes smaller. The second and third branches process raw-resolution features as well as high-resolution features, respectively. Unlike the first branch, the third branch is concerned with the state when the object size becomes larger. Integrating the features of the three branches can extract the low-, mid- and high-level structural information simultaneously, which facilitates a more complete segmentation of the salient regions.

\begin{figure}[htbp]
	\centering
	\includegraphics[width=1\linewidth]{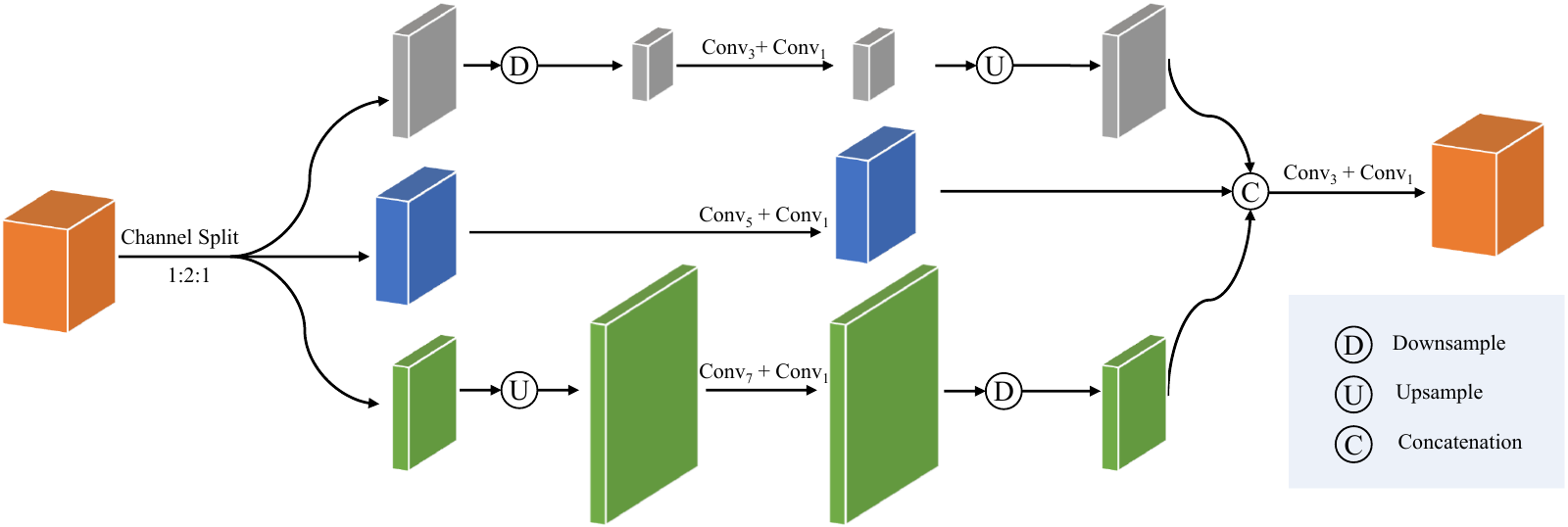}
	\caption{
		The construction of MSP module. The feature layer is first separated into 3 sub-layers at a specific proportion. For the first sub-layer, it is down-sampled and its features are extracted using $3\times3$ convolution. The second sub-layer is processed with $5\times5$ convolution. After the third sub-layer is up-sampled, the larger $7\times7$ convolution is adopted to collect its useful information. By aggregating the processing results of the three sub-layers, the saliency information is enhanced.
	}
	\label{fig:MSPM}
\end{figure}

Specifically, given a feature layer $K$, we first split it along the channel dimension to obtain three sub-layers:
\begin{equation}
\label{eq:msp}
\{K_1, K_2, K_3\}=\operatorname{Split}(K),
\end{equation}
where the ratio of the number of feature maps between $K_1$, $K_2$ and $K_3$ is 1:2:1, aiming to learn more structural information of the original scale, while using the low- and high-level information as auxiliary information. Then, we perform feature extraction for $K_1$, $K_2$ and $K_3$ in parallel:
\begin{equation}
\begin{aligned}
& \tilde{K}_1=\operatorname{Conv}_1(\operatorname{Conv}_3(\downarrow(K_1))), \\
& \tilde{K}_2=\operatorname{Conv}_1(\operatorname{Conv}_5(K_2)), \\
& \tilde{K}_3=\operatorname{Conv}_1(\operatorname{Conv}_7(\uparrow(K_3))),
\end{aligned}
\end{equation}
where $\operatorname{Conv}_1$, $\operatorname{Conv}_3$, $\operatorname{Conv}_5$ and $\operatorname{Conv}_7$ successively represent convolution operations with kernel sizes of $1\times1$, $3\times3$, $5\times5$, and $7\times7$, and $\downarrow(*)$ and $\uparrow(*)$ are down-sampling and up-sampling operations, respectively. Here, after the feature layer is down-sampled, its resolution becomes half of the original one, while after up-sampling, its resolution is twice of the original one. Note that we use convolution kernels whit different sizes in three branches, where $3\times3$ convolution kernels for $K_1$ with low resolution, and $5\times5$ and $7\times7$ convolution kernels for $K_2$ and $K_3$ with higher resolution, respectively. Utilizing small convolution kernels on low-resolution features help to perceive small-sized objects, while applying relatively larger convolution kernels on high-resolution features can make it easier to perceive a large range of complete regions.

After extracting structural information at all levels, we integrate them to obtain the enhanced features $\tilde{K}$:
\begin{equation}
\tilde{K}=\operatorname{Conv}_1(\operatorname{Conv}_3(\operatorname{Cat}(\uparrow(\tilde{K}_1), \tilde{K}_2, \downarrow(\tilde{K}_3)))),
\end{equation}
where $\operatorname{Cat}(*)$ indicates cross-channel concatenation and $\tilde{K}$ is regarded as the output of MSP module. On the whole, MSP can be considered as a further refinement of the internal information of $K$ to make it more adapted to express accurate object information. Some feature maps of $K$ and $\tilde{K}$ are visualized in Fig.~\ref{fig:feat}. It can be seen that the $\tilde{K}$ generated by MSP is able to highlight the complete salient regions more than the original $K$.

\begin{figure}[htbp]
	\centering
	\subfigure[Input]{
		\begin{minipage}[b]{0.21\linewidth}
			\includegraphics[width=1\linewidth]{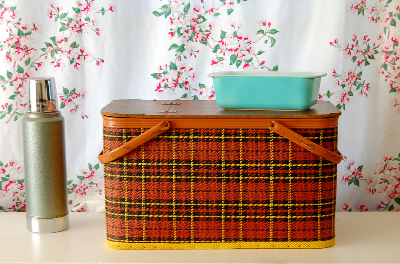}\vspace{1.2pt}
			\includegraphics[width=1\linewidth]{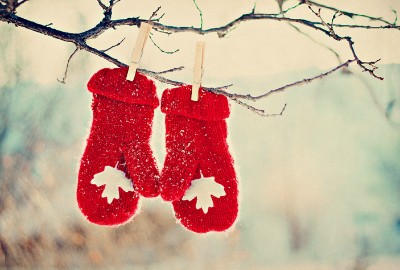}\vspace{1.2pt}
			\includegraphics[width=1\linewidth]{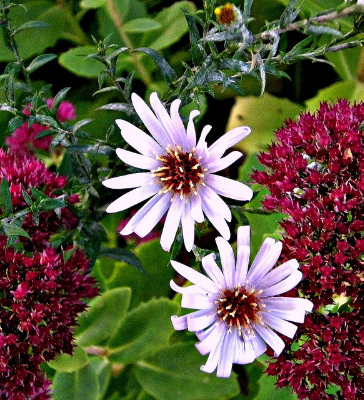}\vspace{1.2pt}
	\end{minipage}}
	\hspace{-2ex}
	\subfigure[GT]{
		\begin{minipage}[b]{0.21\linewidth}
			\includegraphics[width=1\linewidth]{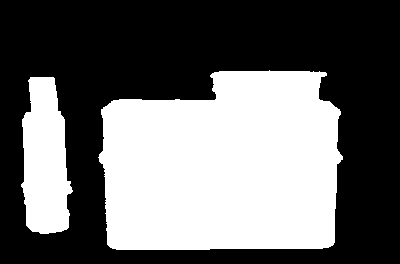}\vspace{1.2pt}
			\includegraphics[width=1\linewidth]{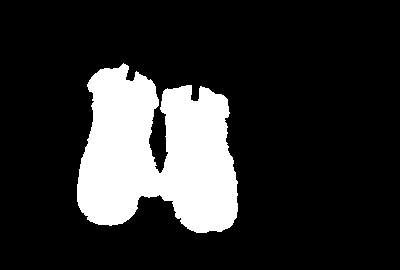}\vspace{1.2pt}
			\includegraphics[width=1\linewidth]{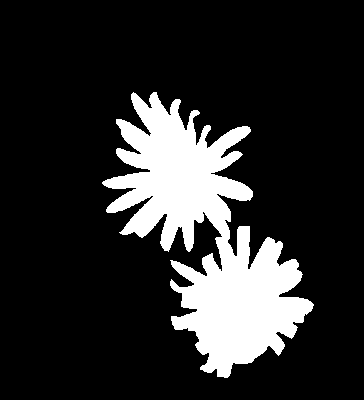}\vspace{1.2pt}
	\end{minipage}}
	\hspace{-2ex}
	\subfigure[$K$]{
		\begin{minipage}[b]{0.21\linewidth}
			\includegraphics[width=1\linewidth]{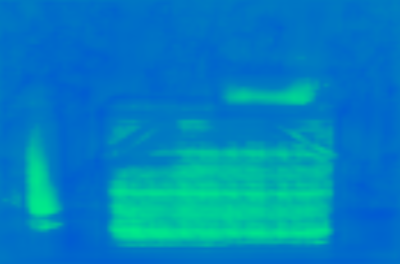}\vspace{1.2pt}
			\includegraphics[width=1\linewidth]{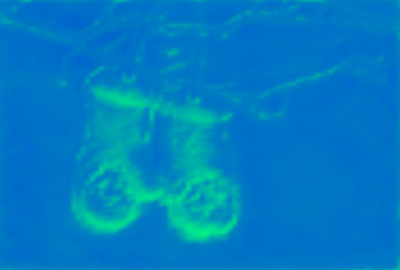}\vspace{1.2pt}
			\includegraphics[width=1\linewidth]{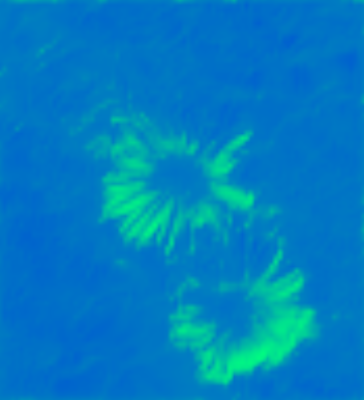}\vspace{1.2pt}
	\end{minipage}}
	\hspace{-2ex}
	\subfigure[$\tilde{K}$]{
		\begin{minipage}[b]{0.21\linewidth}
			\includegraphics[width=1\linewidth]{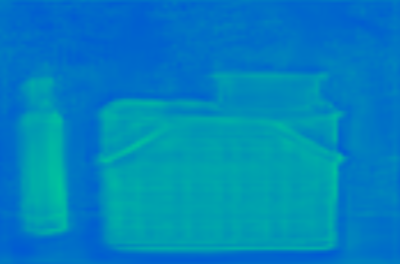}\vspace{1.2pt}
			\includegraphics[width=1\linewidth]{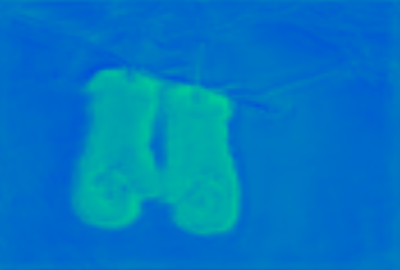}\vspace{1.2pt}
			\includegraphics[width=1\linewidth]{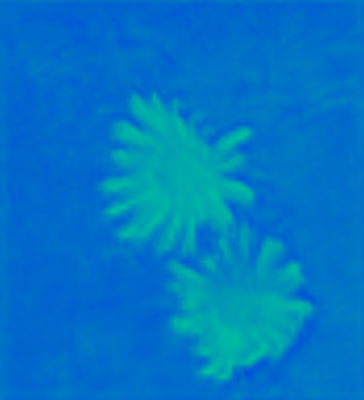}\vspace{1.2pt}
	\end{minipage}}
	\caption{Feature map visualization. Obviously, salient regions receive more attention in $K$.}
	\label{fig:feat}
\end{figure}

\begin{figure*}[htbp]
	\centering
	\includegraphics[width=1\linewidth]{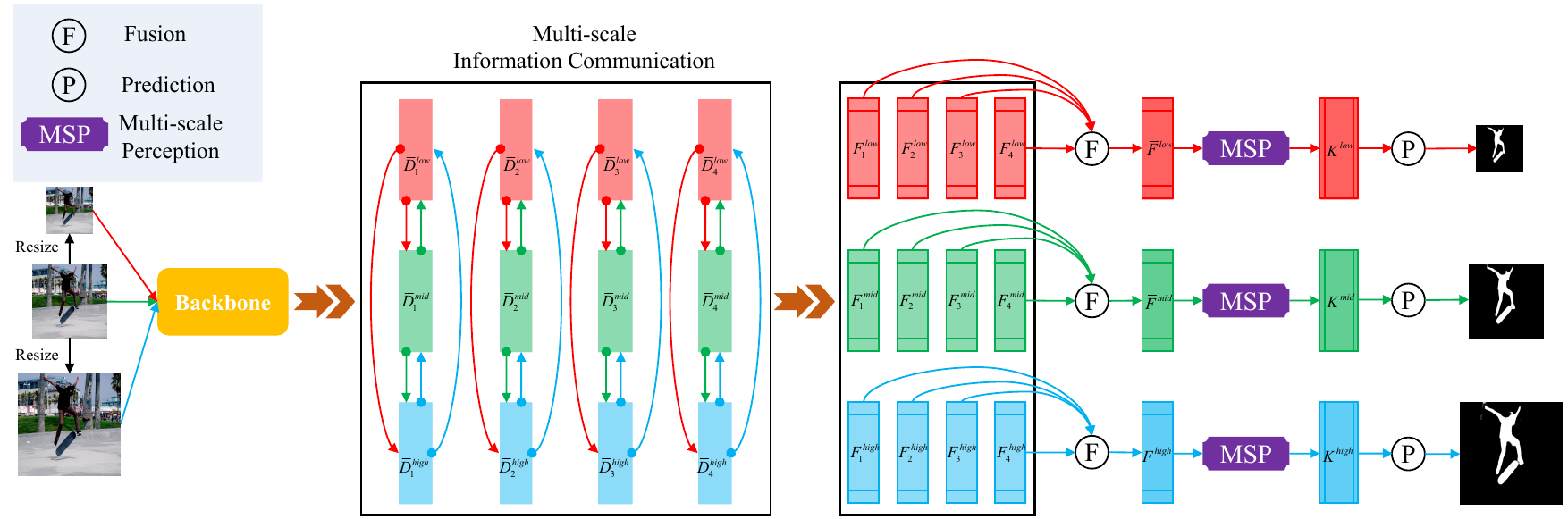}
	\caption{
		The saliency inference steps of MIMONet. First, an original image is down-sampled and up-sampled, with factors of 0.5 and 1.5, respectively, to obtain three images of different resolutions. These images are fed into an identical backbone (shared parameters), and then four feature layers will be generated. For each branch feature layer, it absorbs scale change information from the features of the other two branches. After the exchange of scale information, the multi-level features of each branch are fused and fed into an MSP module for feature enhancement. Finally, the optimized features are applied to detect salient regions. Thus, each branch outputs a saliency map.
	}
	\label{fig:net}
\end{figure*}

\subsection{Multi-scale Input and Multi-scale Output Network}
An RGB image contains limited information, and in some cases it may not provide sufficient clues to determine the foreground objects. In order to introduce additional information to assist the network in further judging the objects, we resize the RGB image to generate both low- and high-resolution images. The original images and images with different resolutions will be fed into the network simultaneously. After extracting the multi-level features of these images, we integrate the features with different resolutions at each level to obtain the size-wise complementary features. These size-wise complementary features aggregate cues for changing the object size from small to large. Employing these features for inference, the network can robustly identify object with different sizes. For different resolution images, MIMONet will output their corresponding saliency maps. The overall framework of MIMONet is shown in Fig.~\ref{fig:net}.

Specifically, given an RGB image $I$, we resize it (with scaling factors of 0.5 and 1.5, respectively) to obtain a low-resolution image $I^\text {low}$ and a high-resolution image $I^\text {high}$. To further distinguish the various images, we define the original $I$ as $I^\text {mid}$. Images of the three resolutions are fed simultaneously into an identical backbone (ResNet50~\cite{he2016deep}). Taking $I^\text {low}$ as an example, its corresponding multi-level feature set is defined as $D^\text {low}=\left\{D_1^\text {low}, D_2^\text {low}, D_3^\text {low}, D_4^\text {low}\right\}$, each level feature has different resolutions and number of feature maps, $\left(\frac{H^\text {low}}{2^{i+1}}, \frac{W^\text {low}}{2^{i+1}}\right)$ and $2^{7+i}$, respectively, where $i=1,2,3,4$, and $H^\text {low}$ and $W^\text {low}$ are the height and width of $I^\text {low}$. Similarly, $D^\text {mid}=\left\{D_1^\text {mid}, D_2^\text {mid}, D_3^\text {mid}, D_4^\text {mid}\right\}$ and $D^\text {high}=\left\{D_1^\text {high}, D_2^\text {high}, D_3^\text {high}, D_4^\text {high}\right\}$. To facilitate the subsequent feature fusion process, a $3\times3$ convolution layer is used for all feature layers in the three feature sets, so that they all have 64 feature maps. This process generates new feature sets, $\bar{D}^\text {low}$, $\bar{D}^\text {mid}$ and $\bar{D}^\text {high}$. Taking processing $D^\text {low}$ as an example, its corresponding new feature set $\bar{D}^\text {low}$ is calculated as
\begin{equation}
\bar{D}_i^\text {low}=\operatorname{Conv}_3(D_i^\text {low}),i=1,2,3,4.
\end{equation}

It is worth noting that these four $3\times3$ convolution operations do not share parameters. In addition, the parameters of the backbone are shared when the three images are processed with the backbone, as well as the parameters of $3\times3$ convolution operations used for generations of $\bar{D}_i^\text {low}$, $\bar{D}_i^\text {mid}$ and $\bar{D}_i^\text {high}$ are also shared.

For a fixed-resolution RGB image, the extracted multi-level features can only represent the information of single-size objects. However, in realistic scenes, salient object sizes are frequently changing, and it is difficult for limited multi-level features to learn the size change pattern accurately. Therefore, given a feature layer, we project its same-level feature layers of different scales into it. This process is expressed as follows:
\begin{equation}
\small
\begin{aligned}
& 
F_i^{\text {low}}=\operatorname{Conv}_3(\bar{D}_i^{\text {low}}\oplus\operatorname{Conv}_1(\downarrow(\bar{D}_i^{\text {mid}}))\oplus\operatorname{Conv}_1(\downarrow(\bar{D}_i^{\text {high }}))), \\
&
F_i^{\text {mid}}=\operatorname{Conv}_3(\bar{D}_i^{\text {mid}}\oplus\operatorname{Conv}_1(\uparrow(\bar{D}_i^{\text {low}}))\oplus\operatorname{Conv}_1(\downarrow(\bar{D}_i^{\text {high }}))), \\
&
F_i^{\text {high }}=\operatorname{Conv}_3(\bar{D}_i^{\text {high }}\oplus\operatorname{Conv}_1(\uparrow(\bar{D}_i^{\text {low}}))\oplus\operatorname{Conv}_1(\uparrow(\bar{D}_i^{mid}))),
\end{aligned}
\end{equation}
where $F_i^{\text {low}}$, $F_i^{\text {mid}}$ and $F_i^{\text {high}}$ are complementary features of different sizes, and $\oplus$ stands for element-wise addition. Taking the generation of $F_i^{\text {low}}$ as an example, $\bar{D}_i^{\text {low}}$ can absorb additional scale cues from $\bar{D}_i^{\text {mid}}$ and $\bar{D}_i^{\text {high}}$ to compensate for the limitations of its own information representation. By exchanging information across scales, the single-scale feature layer gathers cues of object size changes. Because high- and low-level features have different roles in detection, we integrate multi-level features to produce level-wise complementary features $\bar{F}^{\text {low }}$, $\bar{F}^{\text {mid}}$ and $\bar{F}^{\text {high}}$:
\begin{equation}
\small
\begin{aligned}
& 
\bar{F}^{\text {low}}=\operatorname{Conv}_3(\operatorname{Cat}(F_1^{\text {low}}, \uparrow(F_2^{\text {low}}), \uparrow(F_3^{\text {low}}), \uparrow(F_4^{\text {low}}))), \\
& 
\bar{F}^{\text {mid }}=\operatorname{Conv}_3(\operatorname{Cat}(F_1^{\text {mid}}, \uparrow(F_2^{\text {mid}}), \uparrow(F_3^{\text {mid}}), \uparrow(F_4^{\text {mid}}))), \\
& 
\bar{F}^{\text {high}}=\operatorname{Conv}_3(\operatorname{Cat}(F_1^{\text {high}}, \uparrow(F_2^{\text {high}}), \uparrow(F_3^{\text {high}}), \uparrow(F_4^{\text {high}}))).
\end{aligned}
\end{equation}

Before exploiting the complementary features for saliency inference, we input them into the MSP modules for feature enhancement:
\begin{equation}
\begin{aligned}
& 
K^{\text {low}} = MSP({{\bar F}^{\text {low}}}),\\
&
K^{\text {mid}} = MSP({{\bar F}^{\text {mid}}}),\\
&
K^{\text {high}} = MSP({{\bar F}^{\text {high}}}),
\end{aligned}
\end{equation}
where $K^{\text {low}}$, $K^{\text {mid}}$ and $K^{\text {high}}$ are all enhanced features. After processing by the MSP modules, the three enhanced features have better representation ability. For MIMONet, resizing the original image is the first scale transformation of information, while the introduction of MSP module realizes the second scale transformation. Through two scale transformations, the change rule of object size is more fully understood. Finally, $K^{\text {low}}$, $K^{\text {mid}}$ and $K^{\text {high}}$ are each used to predict a saliency map as the output of MIMONet:
\begin{equation}
\begin{aligned}
& 
S^{\text {low}} = \delta (\operatorname{Conv}_1(\operatorname{Conv}_3(K^{\text {low}}))),\\
&
S^{\text {mid}} = \delta (\operatorname{Conv}_1(\operatorname{Conv}_3(K^{\text {mid}}))),\\
&
S^{\text {high}} = \delta (\operatorname{Conv}_1(\operatorname{Conv}_3(K^{\text {high}}))).
\end{aligned}
\end{equation}
where $\delta$ indicates a sigmoid function. Therefore, $S^{\text {low}}$, $S^{\text {mid}}$ and $S^{\text {high}}$ are the predictions of $I^{\text {low}}$, $I^{\text {mid}}$ and $I^{\text {high}}$, respectively. So far, a saliency detection network with multi-scale inputs and multi-scale outputs is constructed. Fig.~\ref{fig:io} shows some of the results predicted by MIMONet. It can be seen that when multi-resolution images are input, MIMONet is able to clearly segment the salient objects in the scenes where the scale changes.

\begin{figure}[htbp]
	\centering
	\includegraphics[width=1\linewidth]{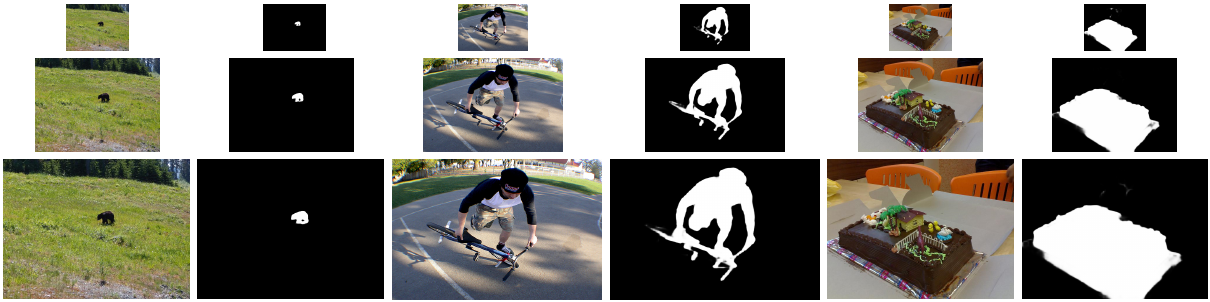}
	\caption{
		Multiple predictions of multi-scale inputs by MIMO.
	}
	\label{fig:io}
\end{figure}

\begin{figure}[htbp]
	\centering
	\includegraphics[width=1\linewidth]{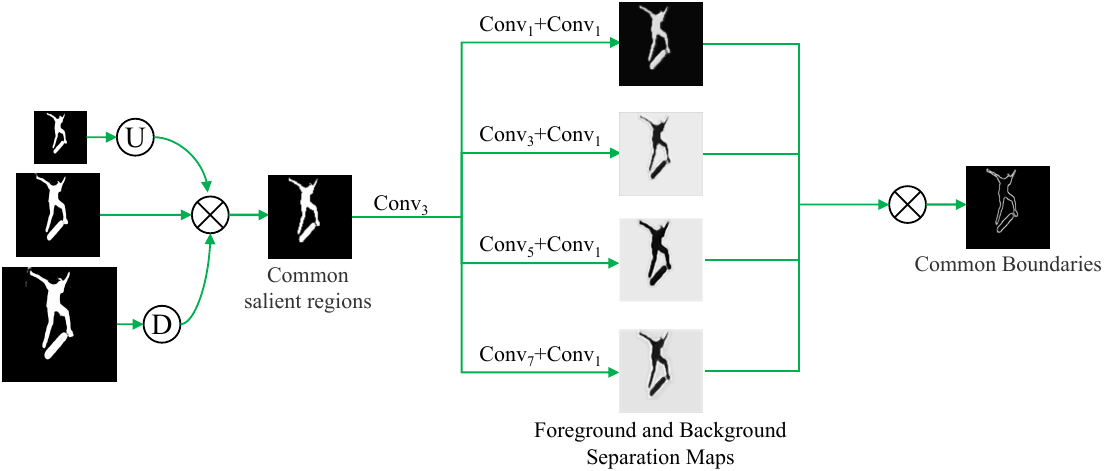}
	\caption{
		The process of generating common boundaries. Specifically, the low- and high-resolution saliency maps are upsampled and downsampled, respectively, and subsequently they are multiplied with the original resolution saliency map. In this way, a common saliency map is generated. We extract its structural information in parallel with different convolution kernels to obtain multiple foreground-background separation maps. By combining these results, a boundary map can be formed.
	}
	\label{fig:jsl}
\end{figure}

\subsection{Loss Function}
For a saliency map $X$, we define the loss between it and the label $Y$ as a combination of binary cross entropy (BCE) and intersection-over-union (IoU) losses. BCE and IoU losses constrain the generation of $X$ to be consistent with $Y$ in terms of the individual and whole loss of pixels, respectively. They are defined as follows:
\begin{small}
\begin{equation}
	L_{B C E}(X,Y)=-\sum_{u, v} Y_{u, v} \log \left(X_{u, v}\right)+\left(1-Y_{u, v}\right) \log \left(1-X_{u, v}\right),
\end{equation}
\end{small}
\begin{equation}
\small
	L_{I o U}(X,Y)=1-\frac{\sum\limits_{u=1}^{H} \sum\limits_{v=1}^{W} X(u, v) \cdot Y(u, v)}{\sum\limits_{u=1}^{H} \sum\limits_{v=1}^{W}[X(u, v)+Y(u, v)-X(u, v) \cdot Y(u, v)]},
\end{equation}
where $(u,v)$ denotes the spatial coordinate position. The MIMONet outputs saliency maps of three different scales. Each prediction is independently supervised by:
\begin{equation}
\label{eq:sal}
\begin{aligned}
& 
L^{\text {low}} = L_{B C E}({S^{\text {low}}},GT{^R}) + {\mu _R}L_{I o U}({S^{\text {low}}},GT{^R}),\\
& 
L^{\text {mid}} = L_{B C E}({S^{\text {mid}}},GT{^R}) + {\mu _R}L_{I o U}({S^{\text {mid}}},GT{^R}),\\
& 
L^{\text {high}} = L_{B C E}({S^{\text {high}}},GT{^R}) + {\mu _R}L_{I o U}({S^{\text {high}}},GT{^R}),
\end{aligned}
\end{equation}
where $GT{^R}$ denotes the salient region label and $\mu _R$ is used to balance the importance of the two losses. In addition, if all predictions are not constrained by consistency, it is not effective to ensure that the same salient objects can be identified from other resolution images as well, resulting in size change patterns of specific objects not being analyzed and learned. In order to make these saliency maps contain the same objects, we propose a joint saliency loss as shown in Fig.~\ref{fig:jsl}. We first calculate the common saliency map ($S^c$) consisting of $S^\text {low}$, $S^\text {mid}$ and $S^\text {high}$:
\begin{equation}
S^c =  \uparrow ({S^{\text {low}}})\otimes{S^{\text {mid}}}\otimes \downarrow ({S^{\text {high}}}),
\end{equation}
where $\otimes$ denotes element-wise multiplication. Then, $S^c$ is further segmented by several branches in parallel, and the corresponding foreground-background separation maps ($S_i^{fb},i=1,2,3,4$) are generated dynamically:
\begin{equation}
\label{eq:cs}
\begin{aligned}
& 
S_1^{fb} = \operatorname{Conv}_1(\operatorname{Conv}_1(T)),\\
&
S_2^{fb} = \operatorname{Conv}_1(\operatorname{Conv}_3(T)),\\
&
S_3^{fb} = \operatorname{Conv}_1(\operatorname{Conv}_5(T)),\\
&
S_4^{fb} = \operatorname{Conv}_1(\operatorname{Conv}_7(T)),
\end{aligned}
\end{equation}
where $T = \operatorname{Conv}_3({S^c})$ indicates that the features of $S^c$ are extracted. Eq.~\ref{eq:cs} can capture the structural information of salient regions diversity by using convolution kernels of different sizes. A boundary map ($B$) is formed by fusing these foreground-background separation maps:
\begin{equation}
B = S_1^{fb}\otimes S_2^{fb}\otimes S_3^{fb}\otimes S_4^{fb}.
\end{equation}

\begin{figure}[htbp]
	\centering
	\subfigure[Image]{
		\begin{minipage}[b]{0.158\linewidth}
			\includegraphics[width=1\linewidth]{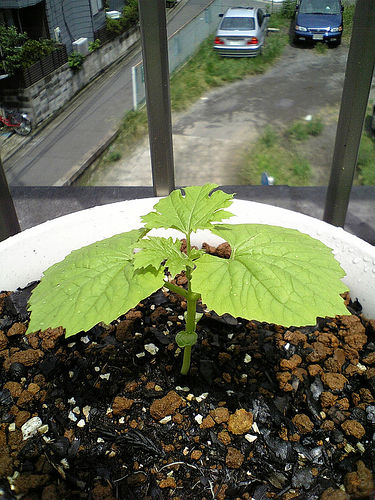}\vspace{1.2pt}
			\includegraphics[width=1\linewidth]{FBS/COCO_train2014_000000054541/img.jpg}\vspace{1.2pt}
			\includegraphics[width=1\linewidth]{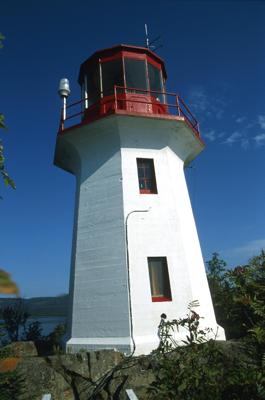}\vspace{1.2pt}
	\end{minipage}}
	\hspace{-2ex}
	\subfigure[FBSM-1]{
		\begin{minipage}[b]{0.158\linewidth}
			\includegraphics[width=1\linewidth]{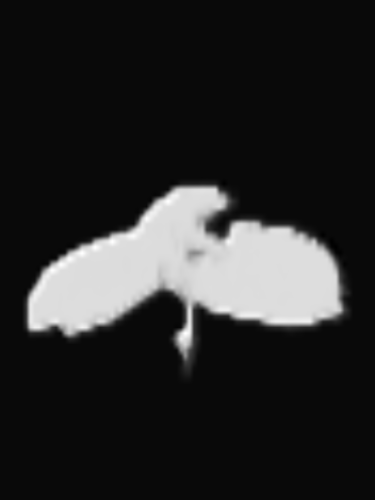}\vspace{1.2pt}
			\includegraphics[width=1\linewidth]{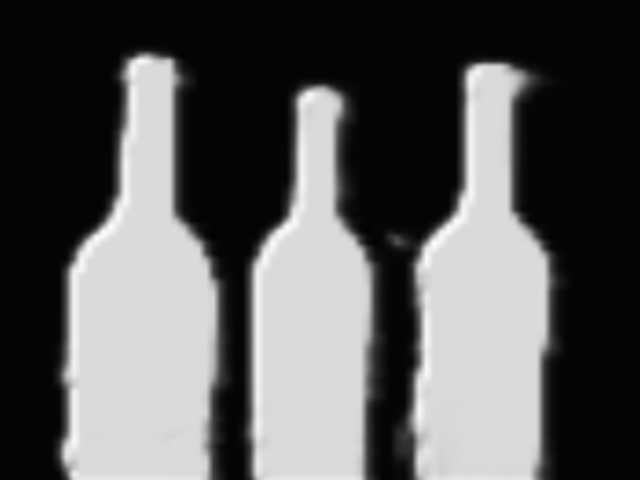}\vspace{1.2pt}
			\includegraphics[width=1\linewidth]{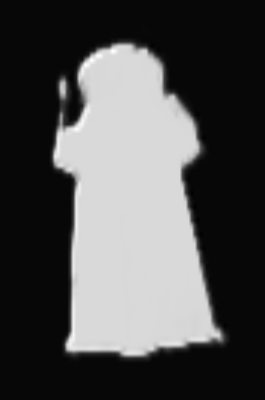}\vspace{1.2pt}
	\end{minipage}}
	\hspace{-2ex}
	\subfigure[FBSM-2]{
		\begin{minipage}[b]{0.158\linewidth}
			\includegraphics[width=1\linewidth]{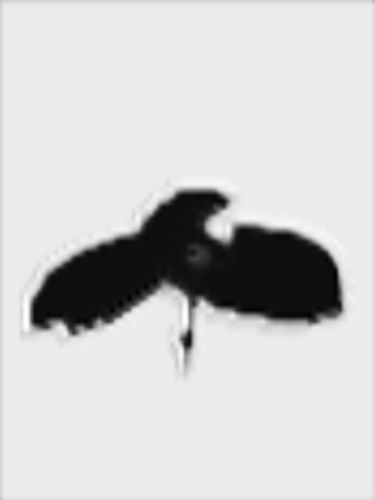}\vspace{1.2pt}
			\includegraphics[width=1\linewidth]{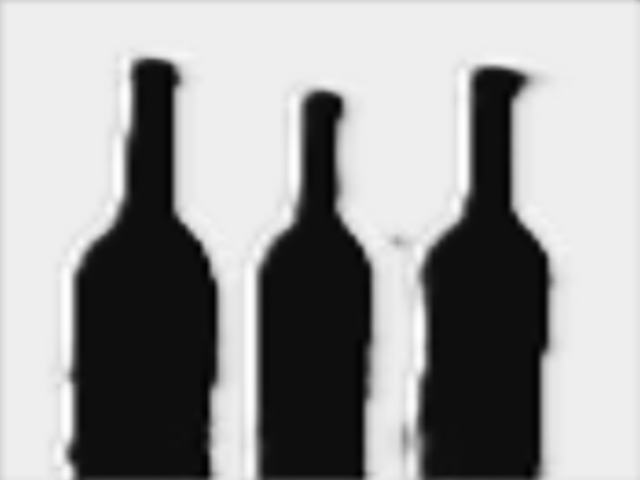}\vspace{1.2pt}
			\includegraphics[width=1\linewidth]{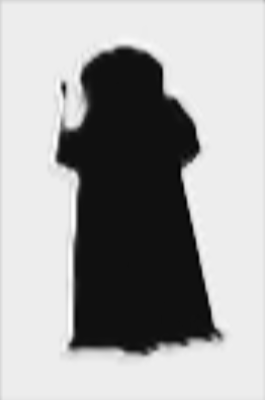}\vspace{1.2pt}
	\end{minipage}}
	\hspace{-2ex}
	\subfigure[FBSM-3]{
		\begin{minipage}[b]{0.158\linewidth}
			\includegraphics[width=1\linewidth]{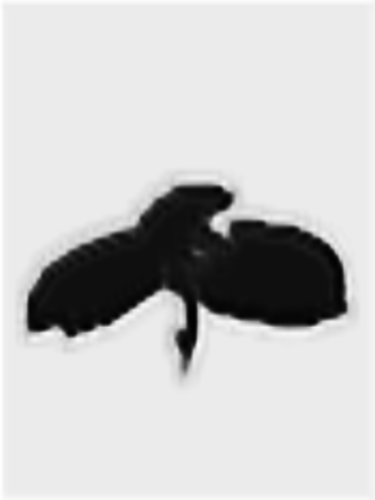}\vspace{1.2pt}
			\includegraphics[width=1\linewidth]{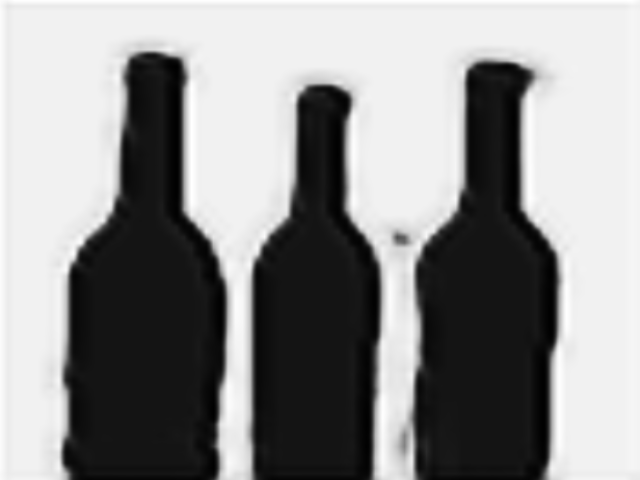}\vspace{1.2pt}
			\includegraphics[width=1\linewidth]{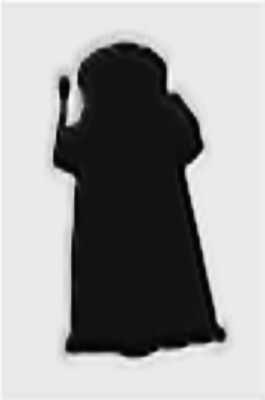}\vspace{1.2pt}
	\end{minipage}}
	\hspace{-2ex}
	\subfigure[FBSM-4]{
		\begin{minipage}[b]{0.158\linewidth}
			\includegraphics[width=1\linewidth]{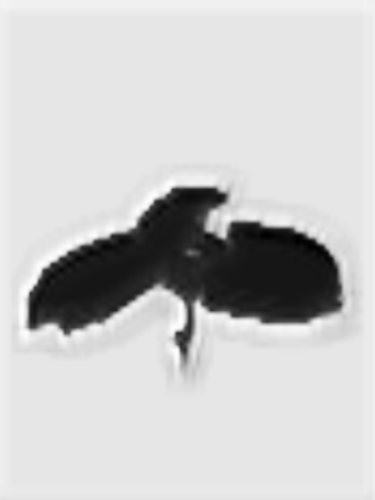}\vspace{1.2pt}
			\includegraphics[width=1\linewidth]{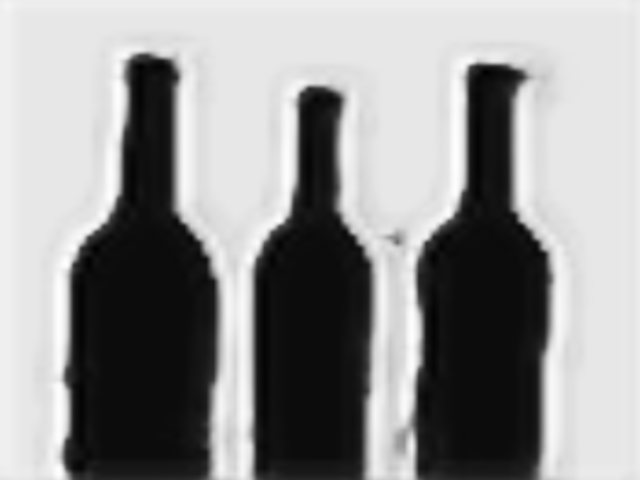}\vspace{1.2pt}
			\includegraphics[width=1\linewidth]{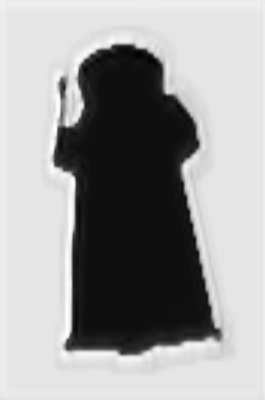}\vspace{1.2pt}
	\end{minipage}}
	\hspace{-2ex}
	\subfigure[Boundary]{
		\begin{minipage}[b]{0.158\linewidth}
			\includegraphics[width=1\linewidth]{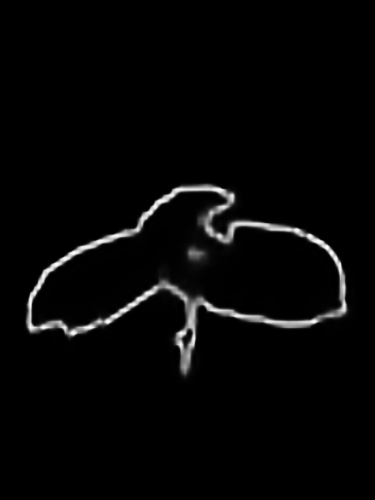}\vspace{1.2pt}
			\includegraphics[width=1\linewidth]{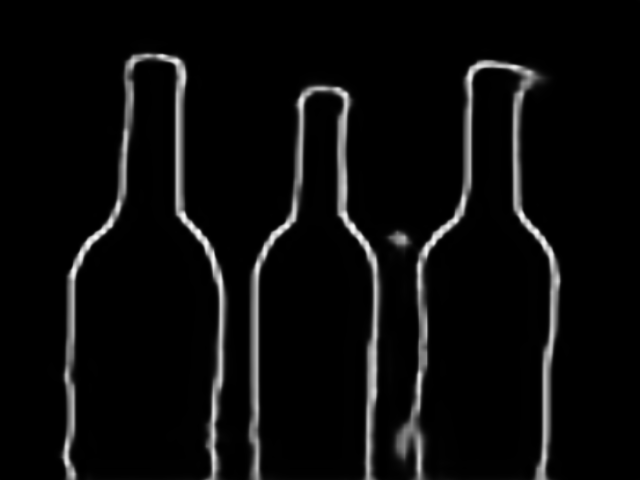}\vspace{1.2pt}
			\includegraphics[width=1\linewidth]{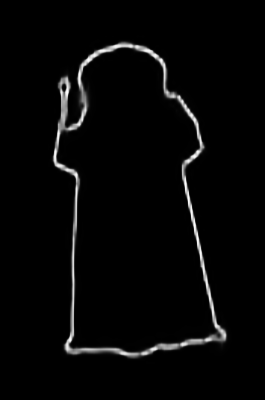}\vspace{1.2pt}
	\end{minipage}}
	\caption{Separation effect between foreground and background. FBSM stands for foreground and background separation map. Numbers 1, 2, 3, and 4 indicate the results processed by different branches, corresponding to the branches of Conv$_1$, Conv$_3$, Conv$_5$, and Conv$_7$, respectively.}
	\label{fig:fbs}
\end{figure}

We also provide some foreground and background separation maps in Fig.~\ref{fig:fbs} along with the calculation results of the common boundaries. We observe that the generation of object boundaries can be determined by fusing four branch structures. Finally, we directly constrain $B$ to align it with the boundary label ($GT^B$), and calculate the loss between them as follows:
\begin{equation}
\label{eq:jsl}
L^{JS} = L_{B C E}(B,G{T^B}) + {\mu _B}L_{I o U}(B,G{T^B}),
\end{equation}
where $\mu _B$ is also a balance factor. Adding the constraint of $L^{JS}$ will enable the saliency maps of three resolutions to predict more accurate salient regions and retain their boundaries. The total loss $L^{total}$ of the network consists of the independent saliency loss and the joint saliency loss of ${S^{\text {low}}}$, ${S^{\text {mid}}}$ and ${S^{\text {high}}}$:
\begin{equation}
{L^{Total}} = {L^{\text {low}}} + {L^{\text {mid}}} + {L^{\text {high}}} + {L^{JS}}.
\end{equation}

\section{Experiments}
\subsection{Experimental Setup}
\subsubsection{Evaluation Datasets} 
The datasets used in our experiments include ECSSD~\cite{yan2013hierarchical}, PASCAL-S~\cite{li2014secrets}, HKU-IS~\cite{li2015visual}, DUT-OMRON~\cite{yang2013saliency}, DUTS~\cite{wang2017learning} and SOC~\cite{fan2018salient}. These six datasets are often adopted by many methods as benchmarks for performance comparisons. ECSSD has 1000 images that contain different scenes, some of which are relatively complex. PASCAL-S is a challenging dataset with 850 images selected from the PASCAL VOC dataset. HKU-IS is composed of 4447 images, and most images have multiple salient objects. DUT-OMRON is a large dataset with 5168 images. Salient objects vary in size and structure, and it is relatively difficult to detect them completely. DUTS consists of a training set (10,533 images) and a test set (5019 images). This dataset covers various natural scenarios and is a difficult dataset to detect salient objects. SOC is a representative and complex dataset. Its test set contains 1200 images that contain salient objects with different attributes.

\begin{table*}[htbp]
	\centering
	
	\caption{
		Quantitative comparisons on six datasets. The best two scores are highlighted in \textcolor[rgb]{.9,.6,.6}{red} and \textcolor[rgb]{.6,.9,.6}{green}, respectively. The subscript for each method indicates its year of publication. ``*'' means using a transformer backbone. $\uparrow$ and $\downarrow$ indicate higher and lower scores, respectively, for better performance
	}\label{Scores1}
	\resizebox{1\textwidth}{!}{
		\begin{tabular}{l|c|c|cccccc|cccccc|cccccc}
			\Xhline{1pt} 
			\multirow{2}[0]{*}{Method} 
			& \#Params
			& Speed
			& \multicolumn{6}{c|}{ECSSD} 
			& \multicolumn{6}{c|}{PASCAL-S}
            & \multicolumn{6}{c}{HKU-IS}
			\\ 
			\cline{4-21} 
			& (M) & (FPS) & $F^{m}_\beta\uparrow$ & $F^{a}_\beta\uparrow$ & $F^{w}_\beta\uparrow$  & $S_m\uparrow$ & $E_m\uparrow$ & $\mathcal{M}\downarrow$
			& $F^{m}_\beta\uparrow$ & $F^{a}_\beta\uparrow$ & $F^{w}_\beta\uparrow$  & $S_m\uparrow$ & $E_m\uparrow$ & $\mathcal{M}\downarrow$& $F^{m}_\beta\uparrow$ & $F^{a}_\beta\uparrow$ & $F^{w}_\beta\uparrow$  & $S_m\uparrow$ & $E_m\uparrow$ & $\mathcal{M}\downarrow$
			\\
			\Xhline{1pt}
			
			CPD$_{2019}$ & 
			47.85&  45& 0.939 & 0.917 & 0.898 & 0.918 & 0.949 & 0.037 &
			0.866 & 0.826 & 0.800 & 0.848 & 0.887 & 0.071  &
            0.925 & 0.891 & 0.875 & 0.906 & 0.950 & 0.034\\
			
			BAS$_{2019}$ & 
			87.06&  42& 0.943 & 0.880 & 0.904 & 0.916 & 0.921 & 0.037 &
			0.860 & 0.777 & 0.799 & 0.838 & 0.853 & 0.075  &
            0.928 & 0.896 & 0.889 & 0.909 & 0.946 & 0.032\\
			
			PoolNet$_{2019}$ & 
			68.26&  38& 0.944 & 0.915 & 0.896 & 0.921 & 0.945 & 0.039 &
			0.869 & 0.821 & 0.799 & 0.849 & 0.876 & 0.074  &
            0.932 & 0.899 & 0.881 & 0.916 & 0.954 & 0.033\\ 
			
			BANet$_{2019}$ & 
			55.92&  10& 0.945 & 0.923 & 0.908 & 0.924 & 0.953 & 0.035 &
			0.870 & 0.830 & 0.808 & 0.852 & 0.886 & 0.070   &
            0.931 & 0.900 & 0.886 & 0.913 & 0.954 & 0.032\\
			
			SCRN$_{2019}$ & 
			\textbf{25.23}&  35& 0.950 & 0.918 & 0.900 & 0.927 & 0.942 & 0.038 &
			0.883 & 0.833 & 0.812 & 0.868 & 0.887 & 0.063  &0.934 & 0.897 & 0.876 & 0.916 & 0.953 & 0.034\\ 
			
			EGNet$_{2019}$ & 
			111.69&  9& 0.947 & 0.920 & 0.903 & 0.925 & 0.947 & 0.037 &
			0.872 & 0.824 & 0.801 & 0.852 & 0.877 & 0.074  &
            0.936 & 0.902 & 0.886 & 0.918 & 0.955 & 0.031\\
			
			F3Net$_{2020}$ & 
			\textbf{25.54}&  37& 0.945 & 0.925 & 0.912 & 0.924 & 0.946 & 0.033 &
			0.878 & 0.841 & 0.822 & 0.860 & 0.894 & 0.062  &0.937 & 0.910 & 0.900 & 0.917 & 0.958 & 0.028\\ 
			
			GCPA$_{2020}$ & 
			67.06&  43& 0.949 & 0.919 & 0.903 & 0.927 & 0.952 & 0.035 &
			0.876 & 0.833 & 0.815 & 0.865 & 0.898 & 0.062  &0.938 & 0.899 & 0.889 & 0.920 & 0.956 & 0.031\\
			
			ITSD$_{2020}$ & 
			\textbf{26.47}&  45& 0.947 & 0.895 & 0.911 & 0.925 & 0.932 & 0.035 &
			0.876 & 0.791 & 0.818 & 0.859 & 0.864 & 0.066  &
            0.934 & 0.899 & 0.894 & 0.917 & 0.953 & 0.031\\ 
			
			GateNet$_{2020}$ & 
			128.63& 48& 0.945 & 0.916 & 0.894 & 0.920 & 0.943 & 0.040 &
			0.875 & 0.825 & 0.802 & 0.857 & 0.884 & 0.068  &
            0.934 & 0.899 & 0.880 & 0.915 & 0.953 & 0.033\\
			
			MINet$_{2020}$ & 
			162.38& 32& 0.948&	0.924&	0.911&	0.925&	0.953&	0.034 &
			0.873&	0.836&	0.815&	0.856&	0.898&	0.063  &
            0.935&	0.909&	0.897&	0.919&	0.960&	0.029\\

			LDF$_{2020}$ & 
			\textbf{25.15}& 30& 0.950&	0.930&	0.915&	0.924&	0.951&	0.034 &
			0.881&	0.850&	0.828&	0.863&	0.905&	\cellcolor[rgb]{.6,.9,.6}0.059  &
            0.939&	0.914&	0.904&	0.920&	0.960&	0.028\\
		
			VST$^*_{2021}$ & 
			44.48&  25& 0.951&	0.920&	0.910&	\cellcolor[rgb]{.6,.9,.6}0.932&	0.957&	0.033 &
			0.882&	0.835&	0.822&	\cellcolor[rgb]{.9,.6,.6}0.872&	0.903&	0.061  &
            0.942&	0.900&	0.897&	\cellcolor[rgb]{.9,.6,.6}0.928&	0.960&	0.029\\

            MSFNet$_{2021}$&
            33.35& 34& 0.941& 0.929 &0.916 &0.915 &0.954 &0.033&
            0.869 &0.850 &0.828 &0.851 &0.904 &0.061&
            0.927 &0.914 &0.903 &0.908 &0.959 &0.027 \\

            PSG$_{2021}$&
            \textbf{25.55}& 47 & 0.949 &0.932 &0.917 &0.925 &0.954 &\cellcolor[rgb]{.6,.9,.6}0.032&
            0.879 &0.848 &0.827 &0.860 &0.902 &0.061&
            0.936 &0.917 &0.902 &0.917 &0.960 &0.028\\

            PurNet$_{2021}$&
            35.53& 31& 0.945 &0.921 &0.907 &0.925 &0.953 &0.035&
            0.873 &0.826 &0.804 &0.848 &0.891 &0.068&
            0.936 &0.902 &0.892 &0.918 &0.956 &0.030\\

            DCN$_{2021}$&
            37.96& 27&\cellcolor[rgb]{.6,.9,.6}0.952 &\cellcolor[rgb]{.6,.9,.6}0.934 &\cellcolor[rgb]{.6,.9,.6}0.920 &0.928 &\cellcolor[rgb]{.6,.9,.6}0.958 &\cellcolor[rgb]{.6,.9,.6}0.032&
            0.879 &0.843 &0.826 &0.861 &0.902 &0.061&
            0.939 &0.914 &0.905 &0.922 &\cellcolor[rgb]{.6,.9,.6}0.962 &0.027\\

            ICON$_{2022}$&
            33.04 & 46&0.950 &0.928 &0.918 &0.929 &0.954 &\cellcolor[rgb]{.6,.9,.6}0.032&
            0.882 &0.839 &0.824 &0.862 &0.893 &0.064&
            0.940 &0.910 &0.902 &0.920 &0.959 &0.029\\

            EDN$_{2022}$&
            42.85& 40& 0.951 &0.933 &0.918 &0.927 &0.955 &\cellcolor[rgb]{.6,.9,.6}0.032&
            \cellcolor[rgb]{.6,.9,.6}0.886 &0.853 &0.831 &0.864 &0.904 &0.062&
            \cellcolor[rgb]{.6,.9,.6}0.941 &0.919 &0.908 &0.924 &\cellcolor[rgb]{.6,.9,.6}0.962 &\cellcolor[rgb]{.6,.9,.6}0.026\\

            RCSBNet$_{2022}$&
            \textbf{27.97}& 13& 0.944 &0.927 &0.916 &0.922 &0.948 &0.033&
            0.882 &\cellcolor[rgb]{.6,.9,.6}0.854 &\cellcolor[rgb]{.6,.9,.6}0.832 &0.859 &\cellcolor[rgb]{.6,.9,.6}0.908 &\cellcolor[rgb]{.6,.9,.6}0.059&
            0.938 &\cellcolor[rgb]{.6,.9,.6}0.924 &\cellcolor[rgb]{.6,.9,.6}0.909 &0.919 &0.959 &0.027\\

            ACCoNet$_{2023}$&
            127.00& 34& 0.945 &0.929 &0.916 &0.924 &0.956 &\cellcolor[rgb]{.6,.9,.6}0.032&
            0.866 &0.833 &0.812 &0.849 &0.891 &0.069&
            0.932 &0.912 &0.900 &0.915 &0.958 &0.029\\

            \Xhline{1pt}
            
            MIMONet&
            \textbf{28.36}& 32&\cellcolor[rgb]{.9,.6,.6}0.956 &\cellcolor[rgb]{.9,.6,.6}0.941 &\cellcolor[rgb]{.9,.6,.6}0.932 &\cellcolor[rgb]{.9,.6,.6}0.933 &\cellcolor[rgb]{.9,.6,.6}0.962 &\cellcolor[rgb]{.9,.6,.6}0.027&
            \cellcolor[rgb]{.9,.6,.6}0.890 &\cellcolor[rgb]{.9,.6,.6}0.859 &\cellcolor[rgb]{.9,.6,.6}0.843 &\cellcolor[rgb]{.6,.9,.6}0.869 &\cellcolor[rgb]{.9,.6,.6}0.913 &\cellcolor[rgb]{.9,.6,.6}0.054&
            \cellcolor[rgb]{.9,.6,.6}0.944 &\cellcolor[rgb]{.9,.6,.6}0.927 &\cellcolor[rgb]{.9,.6,.6}0.918 &\cellcolor[rgb]{.6,.9,.6}0.926 &\cellcolor[rgb]{.9,.6,.6}0.965 &\cellcolor[rgb]{.9,.6,.6}0.024
            \\

			\hline
            \hline 
			\multirow{2}[0]{*}{Method} 
			& \#Params
			& Speed
			& \multicolumn{6}{c|}{DUT-OMRON} 
			& \multicolumn{6}{c|}{DUTS-test}
            & \multicolumn{6}{c}{SOC-test}
			\\ 
			\cline{4-21} 
			& (M) & (FPS) & $F^{m}_\beta\uparrow$ & $F^{a}_\beta\uparrow$ & $F^{w}_\beta\uparrow$  & $S_m\uparrow$ & $E_m\uparrow$ & $\mathcal{M}\downarrow$
			& $F^{m}_\beta\uparrow$ & $F^{a}_\beta\uparrow$ & $F^{w}_\beta\uparrow$  & $S_m\uparrow$ & $E_m\uparrow$ & $\mathcal{M}\downarrow$& $F^{m}_\beta\uparrow$ & $F^{a}_\beta\uparrow$ & $F^{w}_\beta\uparrow$  & $S_m\uparrow$ & $E_m\uparrow$ & $\mathcal{M}\downarrow$
			\\
			\Xhline{1pt}
			
			CPD$_{2019}$ & 
			47.85&  45& 0.797 & 0.747 & 0.719 & 0.825 & 0.873 & 0.056 &  
			0.866 & 0.805 & 0.795 & 0.869 & 0.904 & 0.043 &
            0.752 &0.688 &0.657 &0.768 &0.814 &0.095\\
			
			BAS$_{2019}$ & 
			87.06&  42& 0.805 & 0.756 & 0.751 & 0.836 & 0.869 & 0.057 &  
			0.860 & 0.791 & 0.803 & 0.866 & 0.884 & 0.048 &
            0.754 &0.698 &0.668 &0.761 &0.815 &0.099\\
			
			PoolNet$_{2019}$ & 
			68.26&  38& 0.808 & 0.747 & 0.729 & 0.836 & 0.869 & 0.056 & 
			0.880 & 0.809 & 0.807 & 0.883 & 0.904 & 0.040 &
            0.760 &0.680 &0.651 &0.766 &0.797 &0.104\\
			
			BANet$_{2019}$ & 
			55.92&  10& 0.803 & 0.746 & 0.736 & 0.832 & 0.865 & 0.059 &  
			0.872 & 0.815 & 0.811 & 0.879 & 0.907 & 0.040 &
            0.758 &0.690 &0.665 &0.769 &0.801 &0.099\\
			
			SCRN$_{2019}$ & 
			\textbf{25.23}&  35& 0.812 & 0.746 & 0.720 & 0.837 & 0.869 & 0.056 & 
			0.888 & 0.809 & 0.803 & 0.885 & 0.902 & 0.040 &
            0.763 &0.691 &0.657 &0.779 &0.807 &0.093\\ 
			
			EGNet$_{2019}$ & 
			111.69&  9& 0.815 & 0.756 & 0.738 & 0.841 & 0.874 & 0.053 & 0.889 & 0.815 & 0.816 & 0.887 & 0.907 & 0.039 &
            0.758 &0.661 &0.631 &0.754 &0.775 &0.111\\
			
			F3Net$_{2020}$ & 
			\textbf{25.54}&  37& 0.813 & 0.766 & 0.747 & 0.839 & 0.876 & 0.053 & 
			0.891 & 0.840 & 0.835 & 0.888 & 0.918 & 0.035 &
            0.777 &0.711 &0.684 &0.781 &0.822 &0.089\\ 
			
			GCPA$_{2020}$ & 
			67.06&  43& 0.812 & 0.748 & 0.734 & 0.839 & 0.869 & 0.056 &  
			0.888 & 0.817 & 0.821 & 0.891 & 0.913 & 0.038 &
            0.771 &0.696 &0.673 &0.783 &0.816 &0.091\\
			
			ITSD$_{2020}$ & 
			\textbf{26.47}&  45& 0.821 & 0.756 & 0.750 & 0.840 & 0.867 & 0.061 & 
			0.883 & 0.804 & 0.824 & 0.885 & 0.898 & 0.041 &
            0.772 &0.686 &0.680 &0.781 &0.796 &0.095\\ 
			
			GateNet$_{2020}$ & 
			128.63& 48& 0.818 & 0.746 & 0.729 & 0.838 & 0.868 & 0.055 &  
			0.888 & 0.807 & 0.809 & 0.885 & 0.903 & 0.040 &
            0.773 &0.692 &0.665 &0.780 &0.812 &0.092\\
			
			MINet$_{2020}$ & 
			162.38& 32& 0.810&	0.756&	0.738&	0.833&	0.873&	0.056 &
			0.884&	0.828&	0.825&	0.884&	0.917&	0.037  &
            0.765 &0.693 &0.669 &0.774 &0.816 &0.090\\

			LDF$_{2020}$ & 
			\textbf{25.15}& 30& 0.820&	0.774&	0.752&	0.839&	0.881&	0.051&
			0.898&	0.855&	0.845&	0.892&	0.929&	\cellcolor[rgb]{.6,.9,.6}0.034  &
            \cellcolor[rgb]{.6,.9,.6}0.785 &\cellcolor[rgb]{.6,.9,.6}0.728 &\cellcolor[rgb]{.6,.9,.6}0.693 &0.785 &\cellcolor[rgb]{.6,.9,.6}0.839 &\cellcolor[rgb]{.6,.9,.6}0.086\\
		
			VST$^*_{2021}$ & 
			44.48&  25& \cellcolor[rgb]{.6,.9,.6}0.825&	0.756&	0.755&	\cellcolor[rgb]{.9,.6,.6}0.850&	0.872&	0.058&
			0.890&	0.818&	0.828&	\cellcolor[rgb]{.6,.9,.6}0.896&	0.916&	0.037  &
            0.774 &0.699 &0.678 &\cellcolor[rgb]{.6,.9,.6}0.787 &0.815 &0.095\\

            MSFNet$_{2021}$&
            33.35& 34& 0.799& 0.778 &0.757 &0.832 &0.876 &0.050&
            0.877 &\cellcolor[rgb]{.6,.9,.6}0.856 &0.841 &0.877 &\cellcolor[rgb]{.6,.9,.6}0.931 &\cellcolor[rgb]{.6,.9,.6}0.034&
            0.759 &0.725 &0.688 &0.773 &0.834 &0.087 \\

            PSG$_{2021}$&
            \textbf{25.55}& 47 & 0.811 &0.772 &0.747 &0.834 &0.878 &0.052&
            0.887 &0.849 &0.835 &0.884 &0.926 &0.036&
            0.772 &0.718 &0.686 &0.781 &0.830 &0.088\\

            PurNet$_{2021}$&
            35.53& 31& 0.814 &0.756 &0.747 &0.841 &0.876 &0.051&
            0.879 &0.818 &0.817 &0.881 &0.915 &0.039&
            0.763 &0.696 &0.666 &0.772 &0.822 &0.089\\

            DCN$_{2021}$&
            37.96& 27&0.823 &0.774 &0.760 &\cellcolor[rgb]{.6,.9,.6}0.845 &\cellcolor[rgb]{.6,.9,.6}0.885 &0.051&
            0.894 &0.844 &0.840 &0.892 &0.923 &0.035&
            0.776 &0.714 &0.687 &0.782 &0.826 &0.090\\

            ICON$_{2022}$&
            33.04 & 46&0.825 &0.772 &0.761 &0.844 &0.879 &0.057&
            0.892 &0.838 &0.837 &0.889 &0.919 &0.037&
            0.775 &0.705 &0.676 &0.776 &0.817 &0.094\\

            EDN$_{2022}$&
            42.85& 40& \cellcolor[rgb]{.9,.6,.6}0.828 &\cellcolor[rgb]{.9,.6,.6}0.785 &\cellcolor[rgb]{.9,.6,.6}0.770 &\cellcolor[rgb]{.9,.6,.6}0.850 &\cellcolor[rgb]{.6,.9,.6}0.885 &0.049&
            \cellcolor[rgb]{.6,.9,.6}0.895 &0.852 &\cellcolor[rgb]{.6,.9,.6}0.845 &0.892 &0.928 &0.035&
            0.777 &0.723 &0.691 &0.782 &0.829 &0.089\\

            RCSBNet$_{2022}$&
            \textbf{27.97}& 13& 0.809 &0.778 &0.752 &0.835 &0.870 &\cellcolor[rgb]{.9,.6,.6}0.047&
            0.889 &\cellcolor[rgb]{.6,.9,.6}0.856 &0.840 &0.881 &0.920 &0.035&
            0.759 &0.708 &0.670 &0.766 &0.823 &0.089\\

            ACCoNet$_{2023}$&
            127.00& 34& 0.807 &0.772 &0.751 &0.834 &0.878 &0.056&
            0.874 &0.836 &0.821 &0.874 &0.914 &0.042&
            0.756 &0.697 &0.661 &0.761 &0.808 &0.100\\

            \Xhline{1pt}
            
            MIMONet&
            \textbf{28.36}& 32&\cellcolor[rgb]{.6,.9,.6}0.825 &\cellcolor[rgb]{.6,.9,.6}0.784 &\cellcolor[rgb]{.6,.9,.6}0.769 &\cellcolor[rgb]{.6,.9,.6}0.845 &\cellcolor[rgb]{.9,.6,.6}0.888 &\cellcolor[rgb]{.6,.9,.6}0.048&
            \cellcolor[rgb]{.9,.6,.6}0.901 &\cellcolor[rgb]{.9,.6,.6}0.867 &\cellcolor[rgb]{.9,.6,.6}0.862 &\cellcolor[rgb]{.9,.6,.6}0.899 &\cellcolor[rgb]{.9,.6,.6}0.936 &\cellcolor[rgb]{.9,.6,.6}0.030&
            \cellcolor[rgb]{.9,.6,.6}0.790 &\cellcolor[rgb]{.9,.6,.6}0.736 &\cellcolor[rgb]{.9,.6,.6}0.707 &\cellcolor[rgb]{.9,.6,.6}0.790 &\cellcolor[rgb]{.9,.6,.6}0.846 &\cellcolor[rgb]{.9,.6,.6}0.081
            \\
			
			\Xhline{1pt}
   
		\end{tabular}
	}
\end{table*}

\begin{figure*}[htbp]
	\centering
	\subfigure[Input]{
		\begin{minipage}[b]{0.082\linewidth}
			\includegraphics[width=1\linewidth]{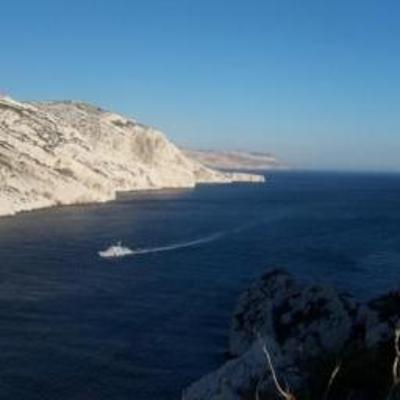}\vspace{1.2pt}
			\includegraphics[width=1\linewidth]{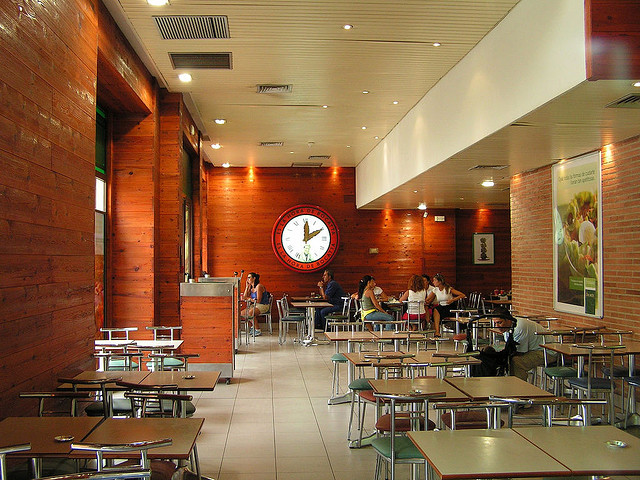}\vspace{1.2pt}
			\includegraphics[width=1\linewidth]{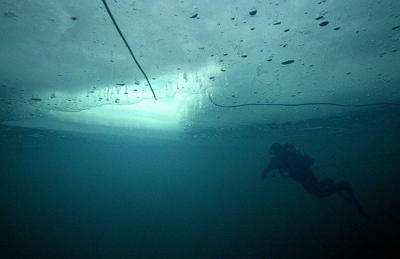}\vspace{1.2pt}
			\includegraphics[width=1\linewidth]{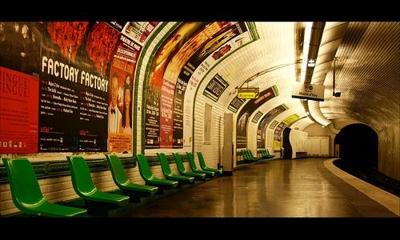}\vspace{1.2pt}
			\includegraphics[width=1\linewidth]{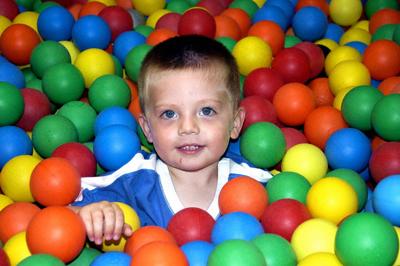}\vspace{1.2pt}
			\includegraphics[width=1\linewidth]{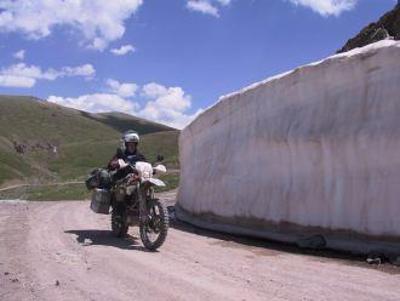}\vspace{1.2pt}
			\includegraphics[width=1\linewidth]{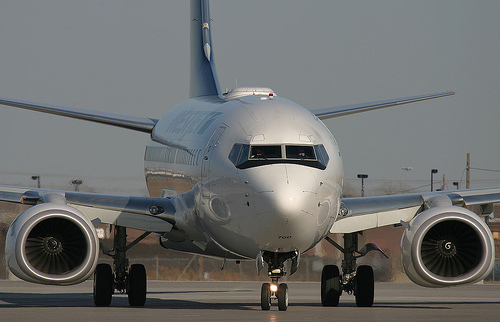}\vspace{1.2pt}
			\includegraphics[width=1\linewidth]{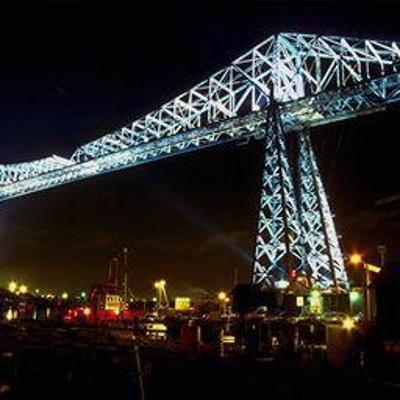}\vspace{1.2pt}
			\includegraphics[width=1\linewidth]{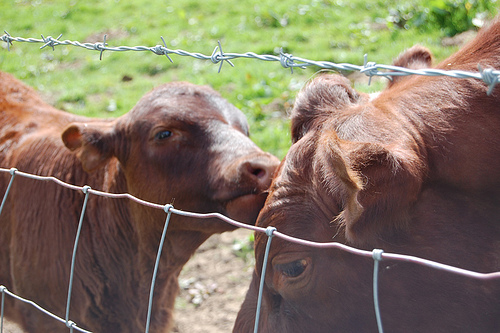}\vspace{1.2pt}
	\end{minipage}}
	\hspace{-2ex}
	\subfigure[GT]{
		\begin{minipage}[b]{0.082\linewidth}
			\includegraphics[width=1\linewidth]{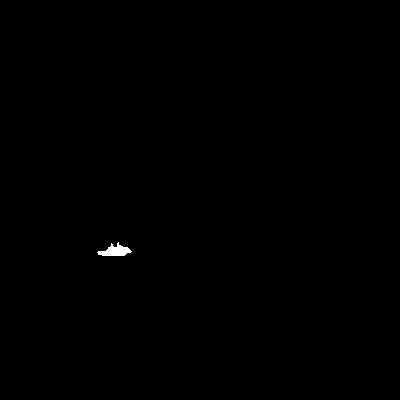}\vspace{1.2pt}
			\includegraphics[width=1\linewidth]{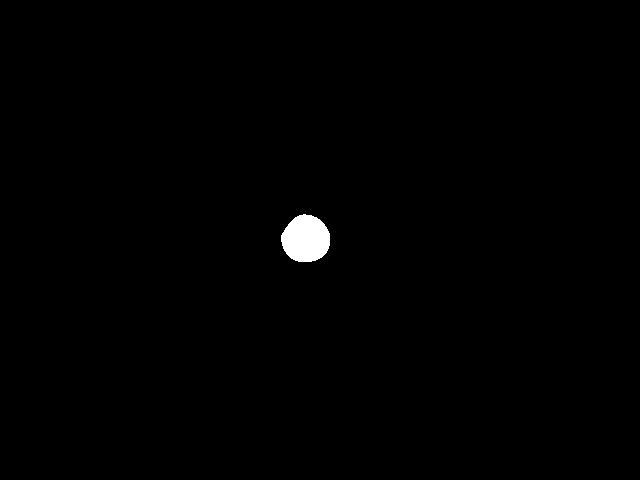}\vspace{1.2pt}
			\includegraphics[width=1\linewidth]{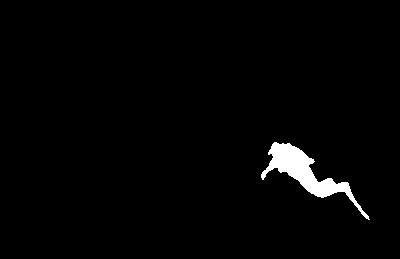}\vspace{1.2pt}
			\includegraphics[width=1\linewidth]{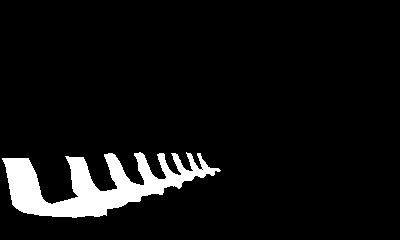}\vspace{1.2pt}
			\includegraphics[width=1\linewidth]{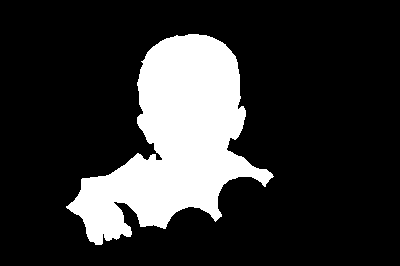}\vspace{1.2pt}
			\includegraphics[width=1\linewidth]{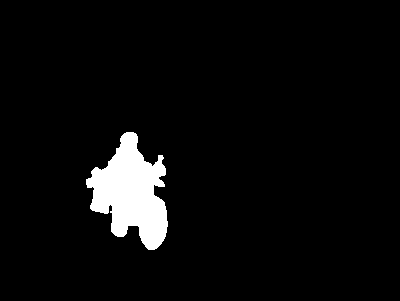}\vspace{1.2pt}
			\includegraphics[width=1\linewidth]{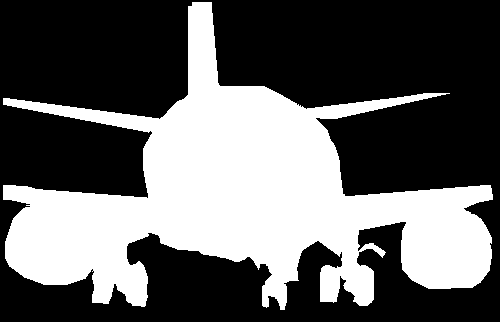}\vspace{1.2pt}
			\includegraphics[width=1\linewidth]{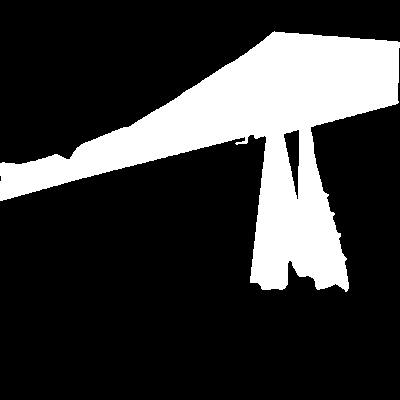}\vspace{1.2pt}
			\includegraphics[width=1\linewidth]{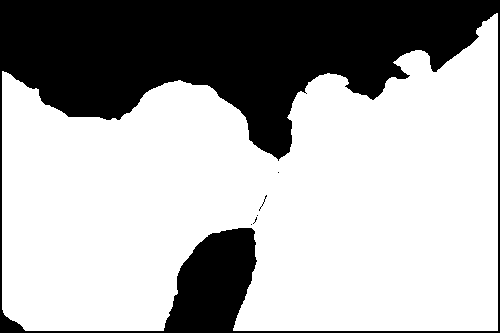}\vspace{1.2pt}
	\end{minipage}}
	\hspace{-2ex}
	\subfigure[Ours]{
		\begin{minipage}[b]{0.082\linewidth}
			\includegraphics[width=1\linewidth]{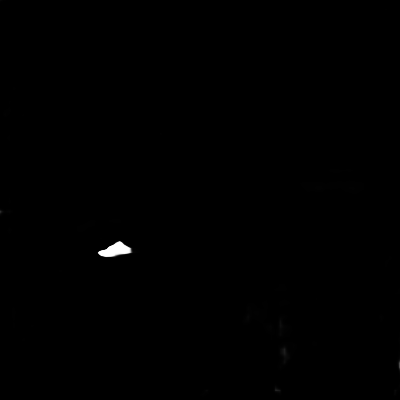}\vspace{1.2pt}
			\includegraphics[width=1\linewidth]{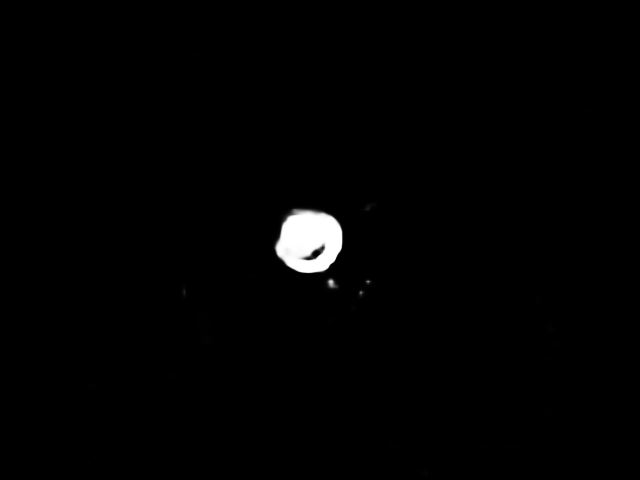}\vspace{1.2pt}
			\includegraphics[width=1\linewidth]{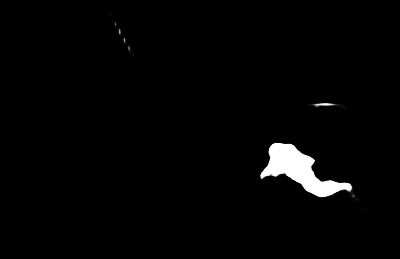}\vspace{1.2pt}
			\includegraphics[width=1\linewidth]{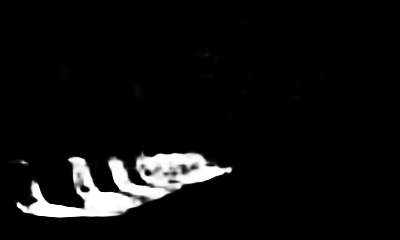}\vspace{1.2pt}
			\includegraphics[width=1\linewidth]{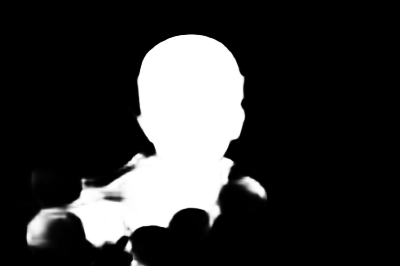}\vspace{1.2pt}
			\includegraphics[width=1\linewidth]{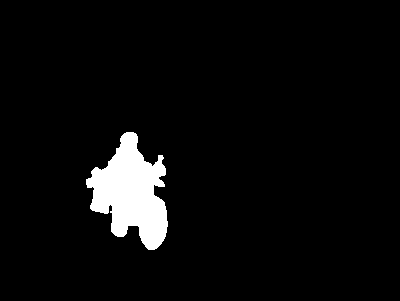}\vspace{1.2pt}
			\includegraphics[width=1\linewidth]{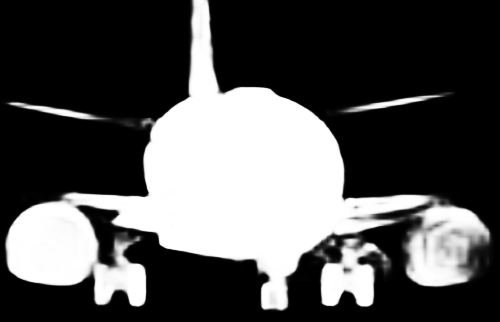}\vspace{1.2pt}
			\includegraphics[width=1\linewidth]{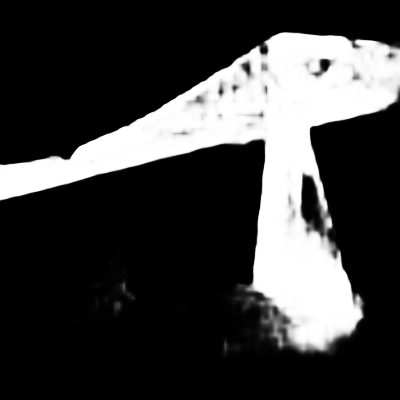}\vspace{1.2pt}
			\includegraphics[width=1\linewidth]{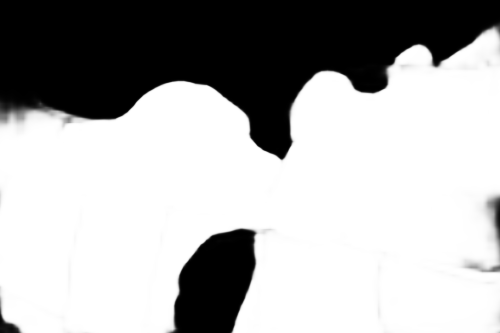}\vspace{1.2pt}
			
	\end{minipage}}
	\hspace{-2ex}
	\subfigure[EDN]{
		\begin{minipage}[b]{0.082\linewidth}
			\includegraphics[width=1\linewidth]{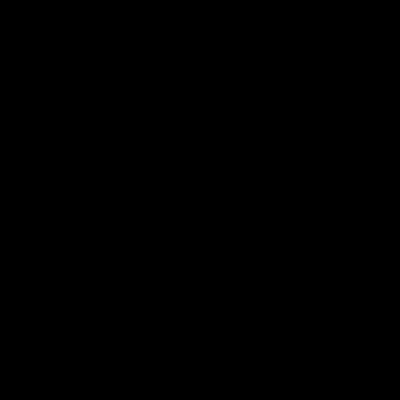}\vspace{1.2pt}
			\includegraphics[width=1\linewidth]{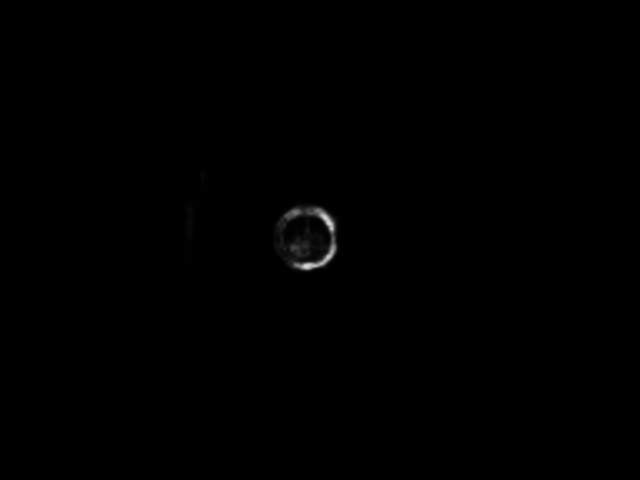}\vspace{1.2pt}
			\includegraphics[width=1\linewidth]{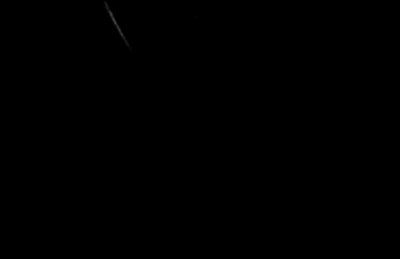}\vspace{1.2pt}
			\includegraphics[width=1\linewidth]{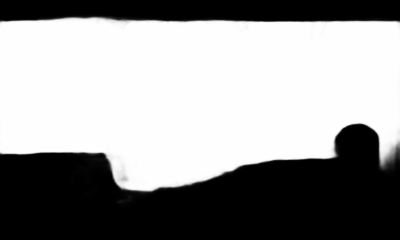}\vspace{1.2pt}
			\includegraphics[width=1\linewidth]{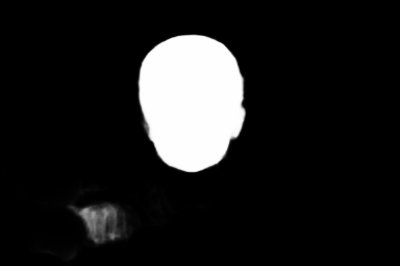}\vspace{1.2pt}
			\includegraphics[width=1\linewidth]{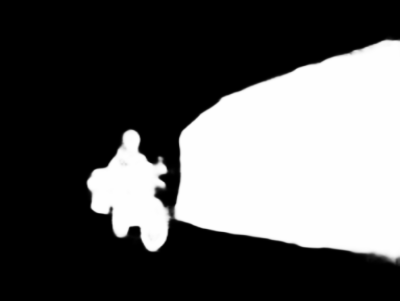}\vspace{1.2pt}
			\includegraphics[width=1\linewidth]{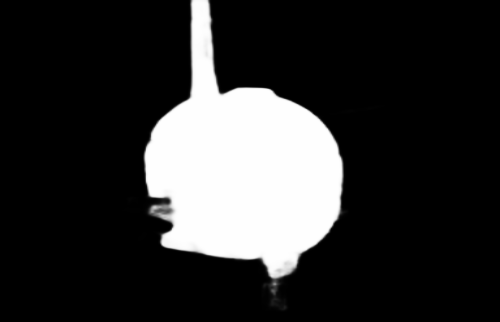}\vspace{1.2pt}
			\includegraphics[width=1\linewidth]{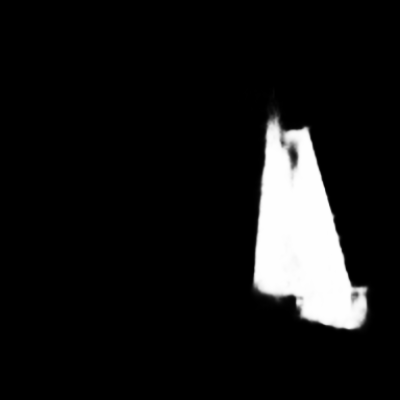}\vspace{1.2pt}
			\includegraphics[width=1\linewidth]{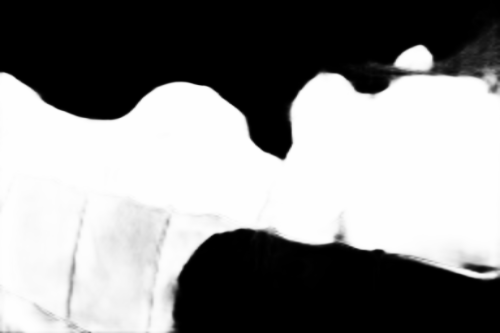}\vspace{1.2pt}
	\end{minipage}}
	\hspace{-2ex}
	\subfigure[ICON]{
		\begin{minipage}[b]{0.082\linewidth}
			\includegraphics[width=1\linewidth]{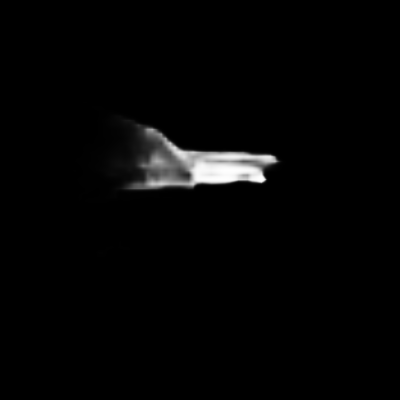}\vspace{1.2pt}
			\includegraphics[width=1\linewidth]{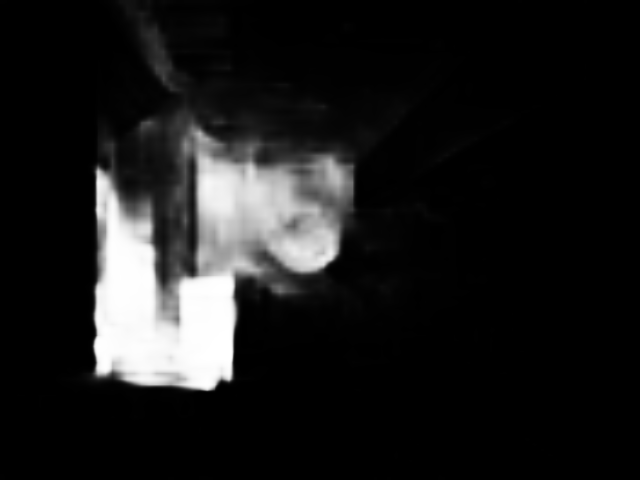}\vspace{1.2pt}
			\includegraphics[width=1\linewidth]{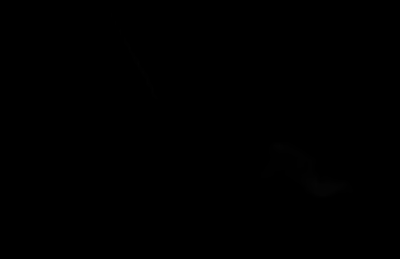}\vspace{1.2pt}
			\includegraphics[width=1\linewidth]{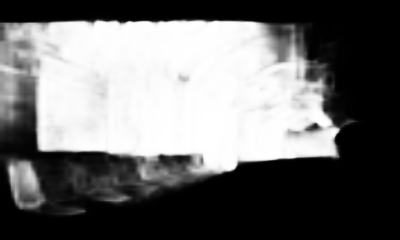}\vspace{1.2pt}
			\includegraphics[width=1\linewidth]{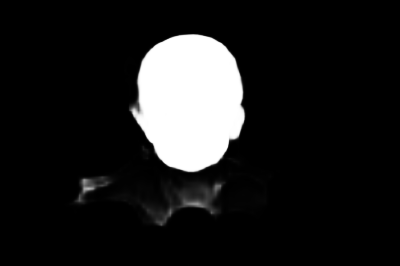}\vspace{1.2pt}
			\includegraphics[width=1\linewidth]{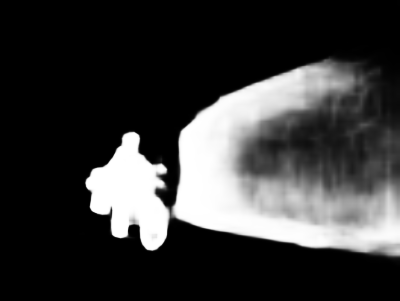}\vspace{1.2pt}
			\includegraphics[width=1\linewidth]{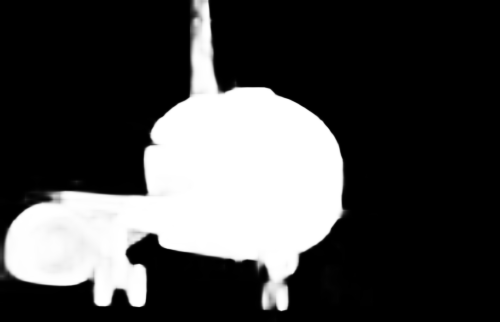}\vspace{1.2pt}
			\includegraphics[width=1\linewidth]{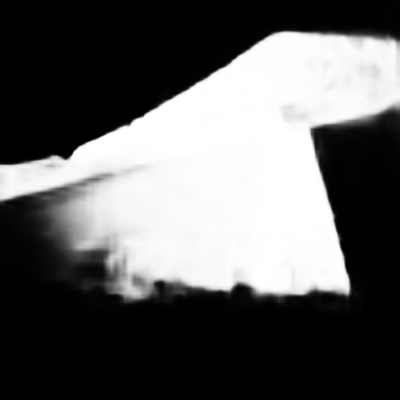}\vspace{1.2pt}
			\includegraphics[width=1\linewidth]{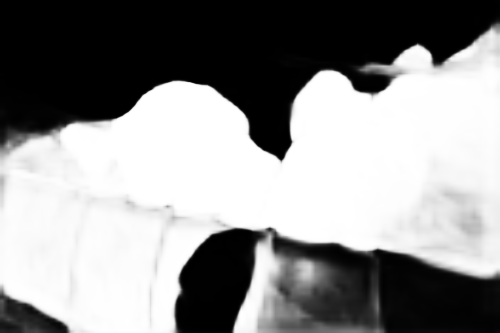}\vspace{1.2pt}
	\end{minipage}}
	\hspace{-2ex}
	\subfigure[DCN]{
		\begin{minipage}[b]{0.082\linewidth}
			\includegraphics[width=1\linewidth]{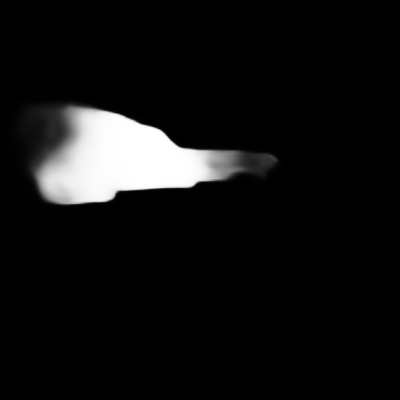}\vspace{1.2pt}
			\includegraphics[width=1\linewidth]{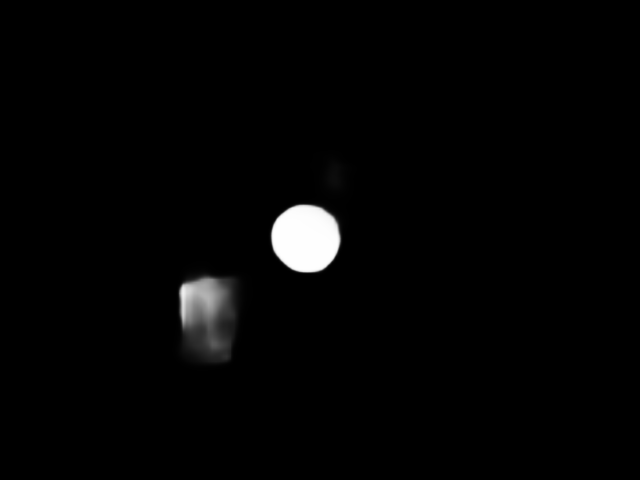}\vspace{1.2pt}
			\includegraphics[width=1\linewidth]{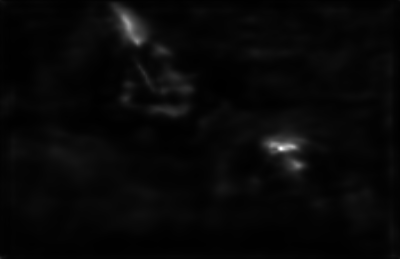}\vspace{1.2pt}
			\includegraphics[width=1\linewidth]{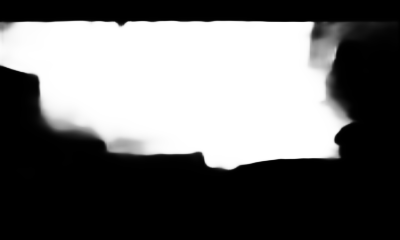}\vspace{1.2pt}
			\includegraphics[width=1\linewidth]{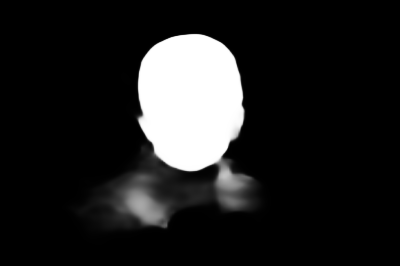}\vspace{1.2pt}
			\includegraphics[width=1\linewidth]{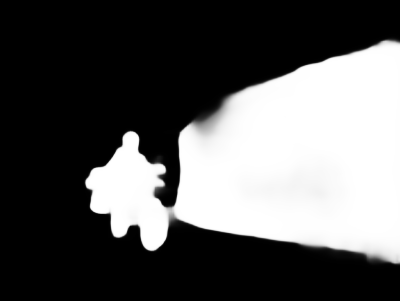}\vspace{1.2pt}
			\includegraphics[width=1\linewidth]{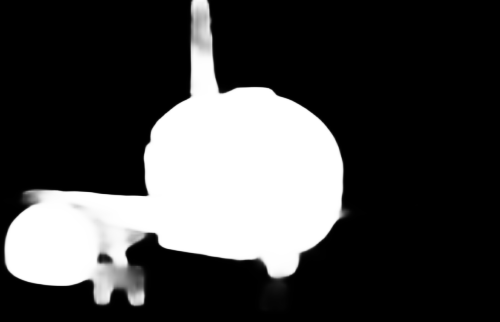}\vspace{1.2pt}
			\includegraphics[width=1\linewidth]{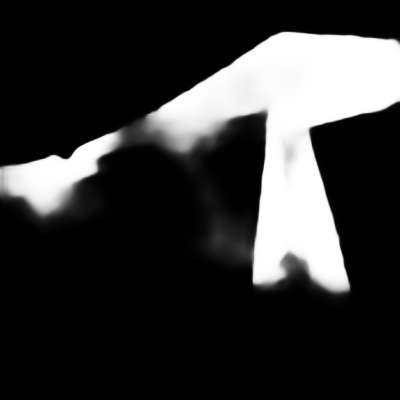}\vspace{1.2pt}
			\includegraphics[width=1\linewidth]{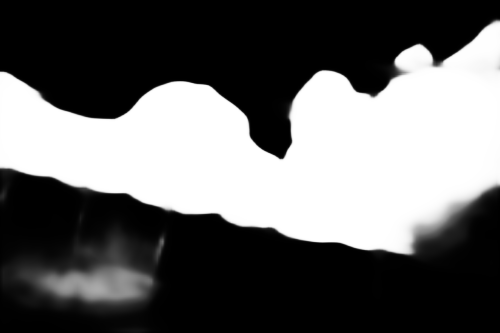}\vspace{1.2pt}
	\end{minipage}}
	\hspace{-2ex}
	\subfigure[MSFN]{
		\begin{minipage}[b]{0.082\linewidth}
			\includegraphics[width=1\linewidth]{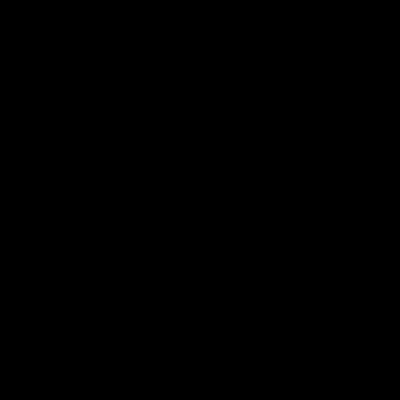}\vspace{1.2pt}
			\includegraphics[width=1\linewidth]{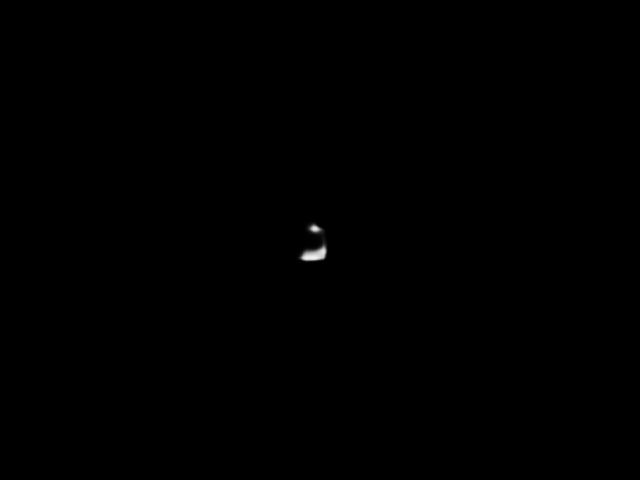}\vspace{1.2pt}
			\includegraphics[width=1\linewidth]{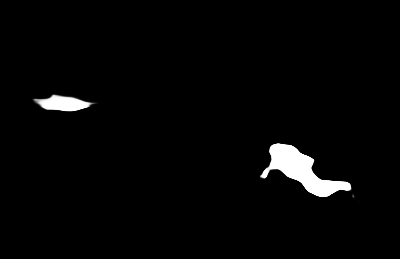}\vspace{1.2pt}
			\includegraphics[width=1\linewidth]{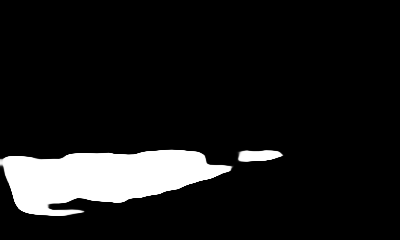}\vspace{1.2pt}
			\includegraphics[width=1\linewidth]{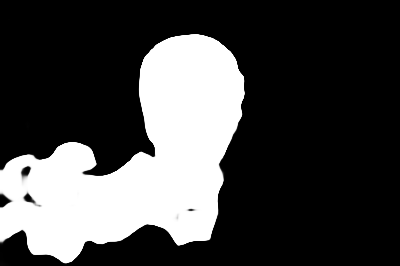}\vspace{1.2pt}
			\includegraphics[width=1\linewidth]{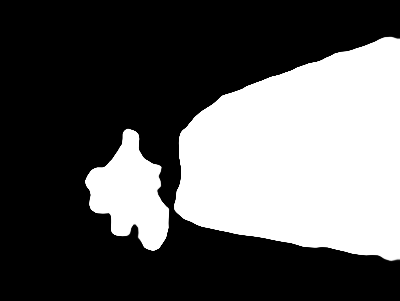}\vspace{1.2pt}
			\includegraphics[width=1\linewidth]{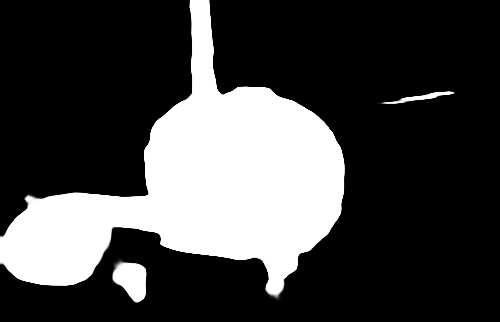}\vspace{1.2pt}
			\includegraphics[width=1\linewidth]{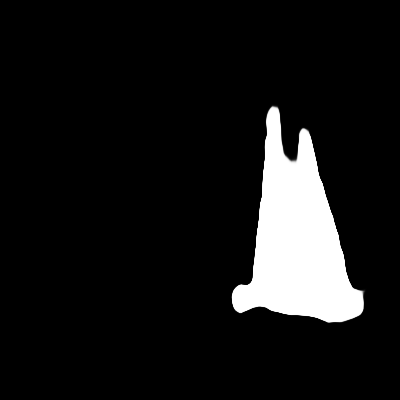}\vspace{1.2pt}
			\includegraphics[width=1\linewidth]{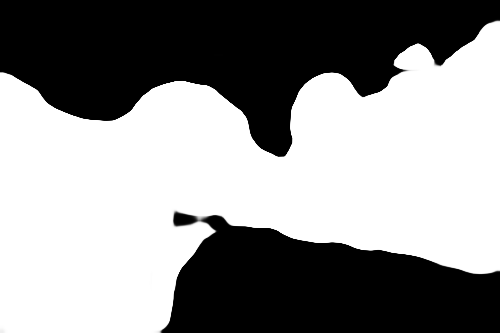}\vspace{1.2pt}
	\end{minipage}}
	\hspace{-2ex}
	\subfigure[MIN]{
		\begin{minipage}[b]{0.082\linewidth}
			\includegraphics[width=1\linewidth]{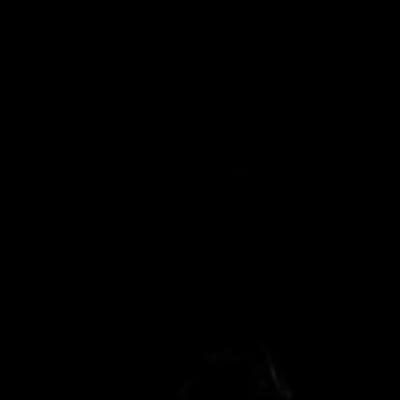}\vspace{1.2pt}
			\includegraphics[width=1\linewidth]{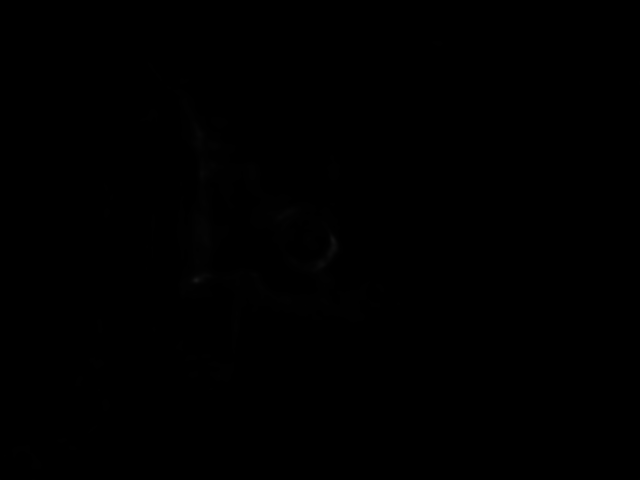}\vspace{1.2pt}
			\includegraphics[width=1\linewidth]{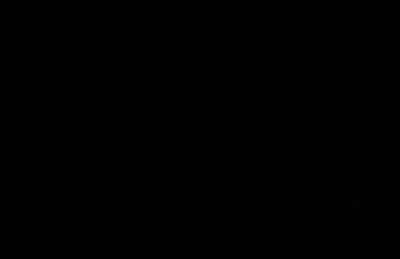}\vspace{1.2pt}
			\includegraphics[width=1\linewidth]{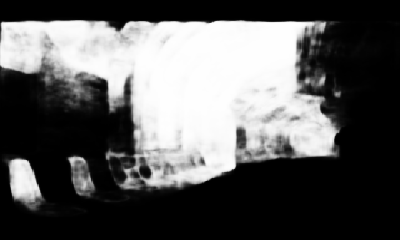}\vspace{1.2pt}
			\includegraphics[width=1\linewidth]{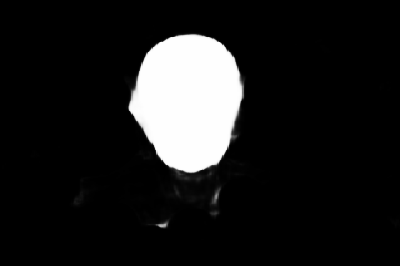}\vspace{1.2pt}
			\includegraphics[width=1\linewidth]{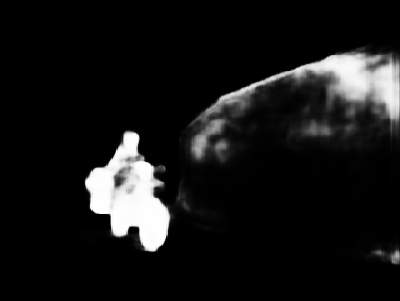}\vspace{1.2pt}
			\includegraphics[width=1\linewidth]{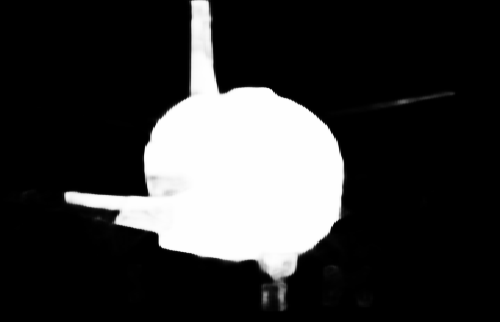}\vspace{1.2pt}
			\includegraphics[width=1\linewidth]{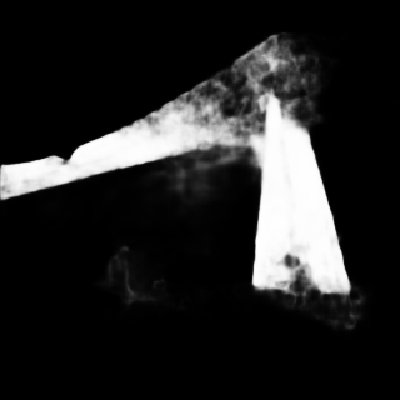}\vspace{1.2pt}
			\includegraphics[width=1\linewidth]{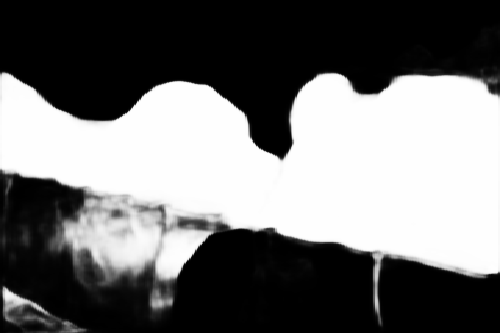}\vspace{1.2pt}
	\end{minipage}}
	\hspace{-2ex}
	\subfigure[ITSD]{
		\begin{minipage}[b]{0.082\linewidth}
			\includegraphics[width=1\linewidth]{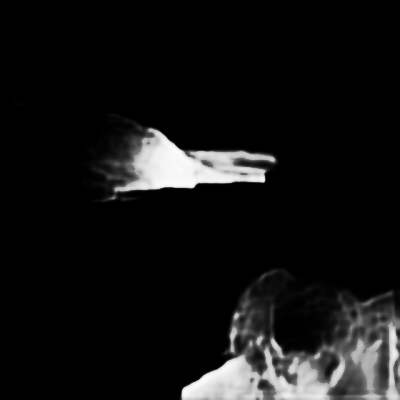}\vspace{1.2pt}
			\includegraphics[width=1\linewidth]{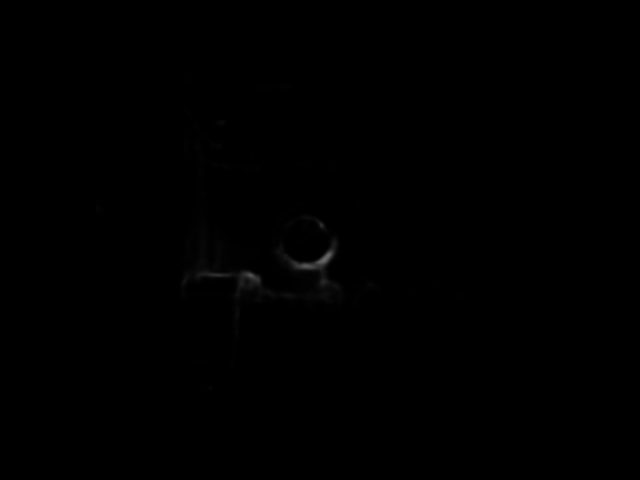}\vspace{1.2pt}
			\includegraphics[width=1\linewidth]{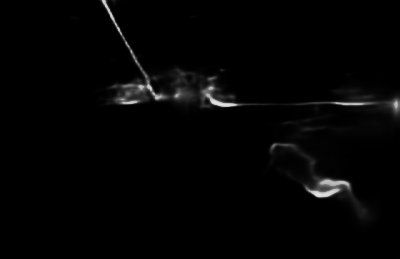}\vspace{1.2pt}
			\includegraphics[width=1\linewidth]{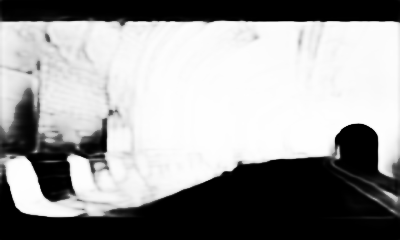}\vspace{1.2pt}
			\includegraphics[width=1\linewidth]{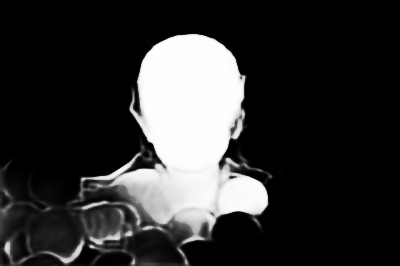}\vspace{1.2pt}
			\includegraphics[width=1\linewidth]{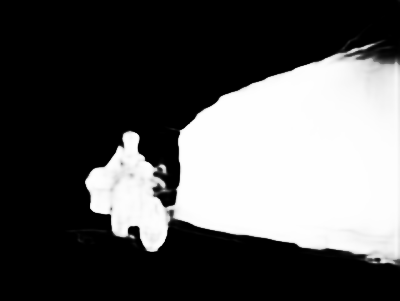}\vspace{1.2pt}
			\includegraphics[width=1\linewidth]{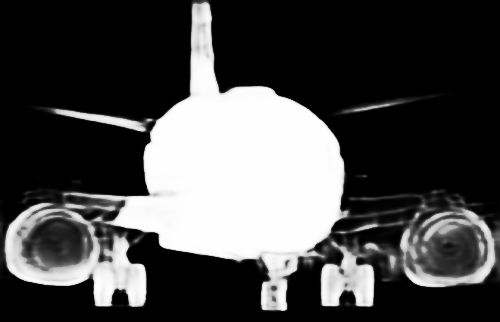}\vspace{1.2pt}
			\includegraphics[width=1\linewidth]{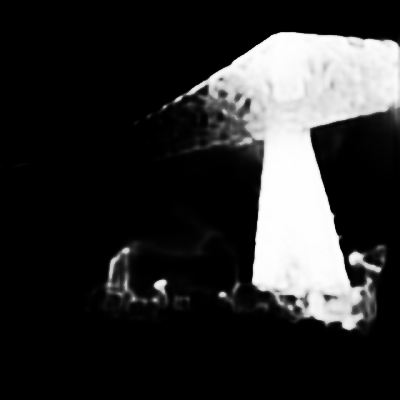}\vspace{1.2pt}
			\includegraphics[width=1\linewidth]{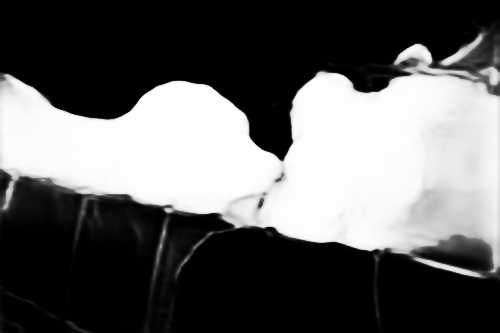}\vspace{1.2pt}
	\end{minipage}}
	\hspace{-2ex}
	\subfigure[LDF]{
		\begin{minipage}[b]{0.082\linewidth}
			\includegraphics[width=1\linewidth]{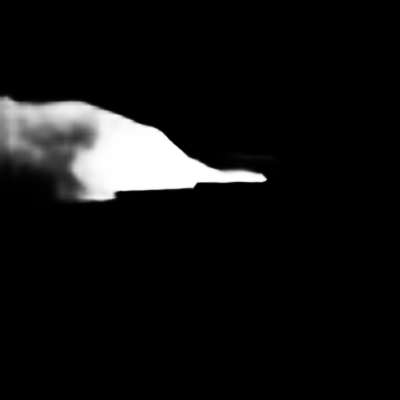}\vspace{1.2pt}
			\includegraphics[width=1\linewidth]{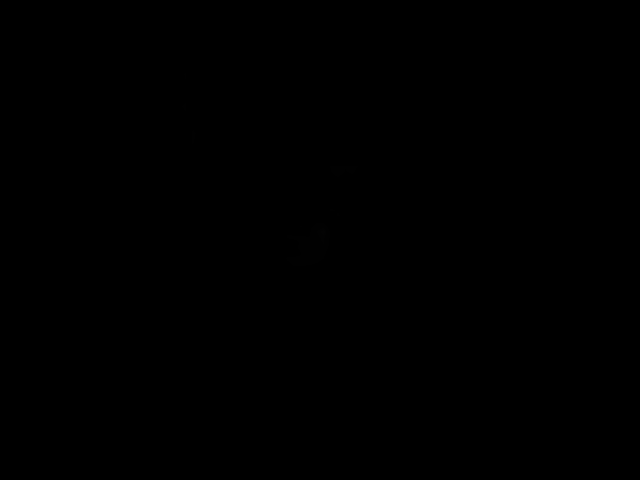}\vspace{1.2pt}
			\includegraphics[width=1\linewidth]{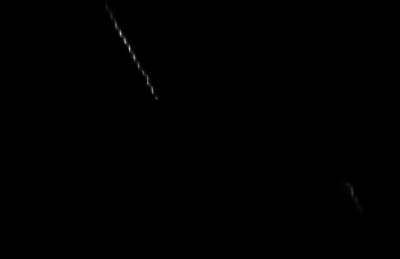}\vspace{1.2pt}
			\includegraphics[width=1\linewidth]{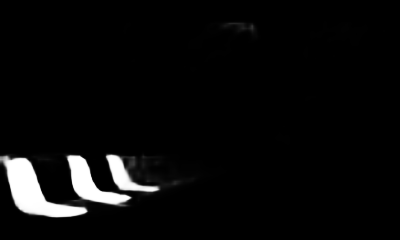}\vspace{1.2pt}
			\includegraphics[width=1\linewidth]{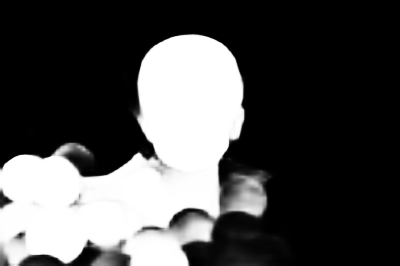}\vspace{1.2pt}
			\includegraphics[width=1\linewidth]{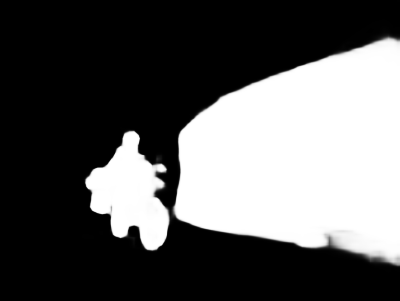}\vspace{1.2pt}
			\includegraphics[width=1\linewidth]{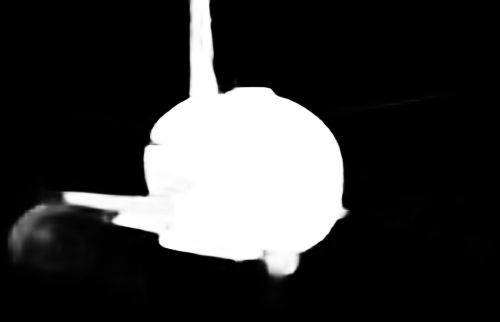}\vspace{1.2pt}
			\includegraphics[width=1\linewidth]{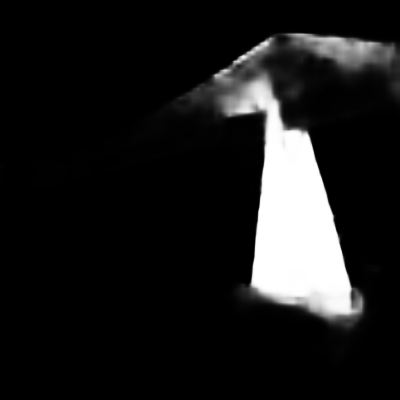}\vspace{1.2pt}
			\includegraphics[width=1\linewidth]{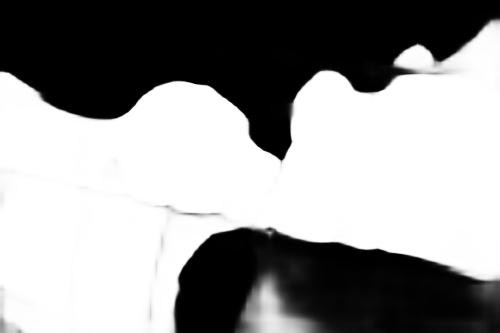}\vspace{1.2pt}
	\end{minipage}}
	\hspace{-2ex}
	\subfigure[EGN]{
		\begin{minipage}[b]{0.082\linewidth}
			\includegraphics[width=1\linewidth]{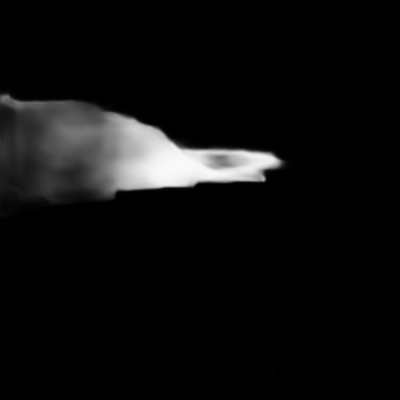}\vspace{1.2pt}
			\includegraphics[width=1\linewidth]{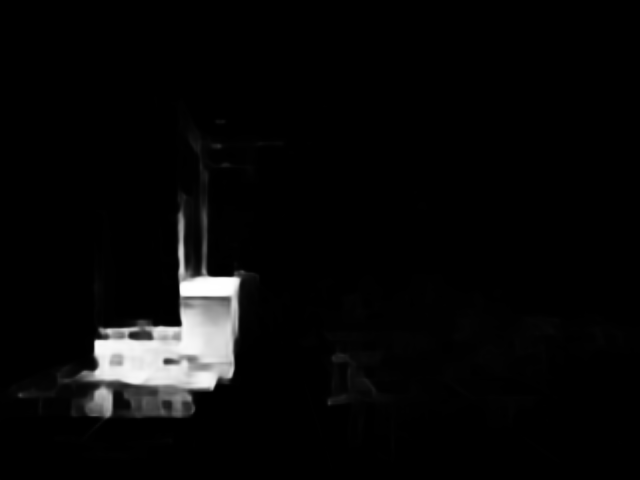}\vspace{1.2pt}
			\includegraphics[width=1\linewidth]{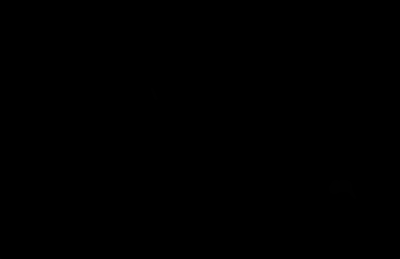}\vspace{1.2pt}
			\includegraphics[width=1\linewidth]{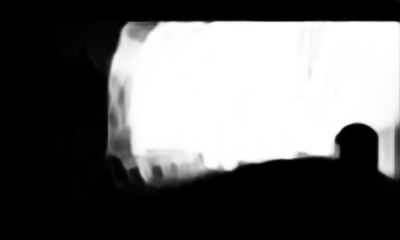}\vspace{1.2pt}
			\includegraphics[width=1\linewidth]{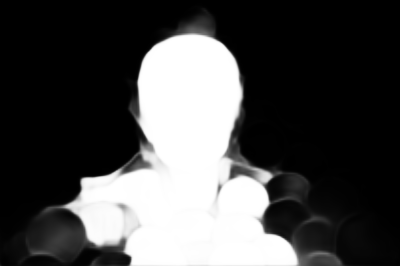}\vspace{1.2pt}
			\includegraphics[width=1\linewidth]{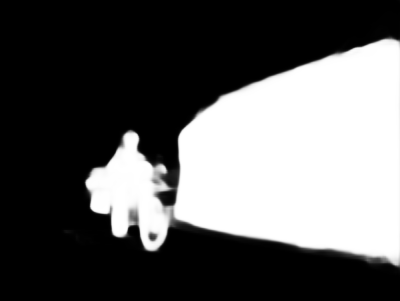}\vspace{1.2pt}
			\includegraphics[width=1\linewidth]{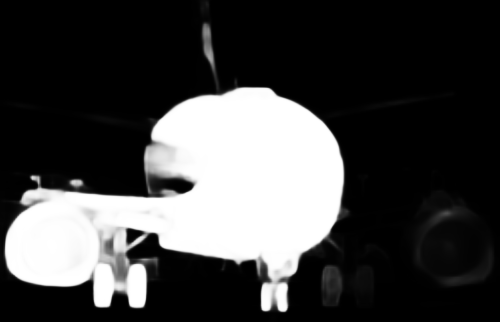}\vspace{1.2pt}
			\includegraphics[width=1\linewidth]{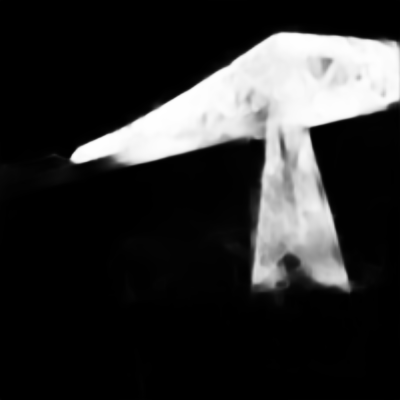}\vspace{1.2pt}
			\includegraphics[width=1\linewidth]{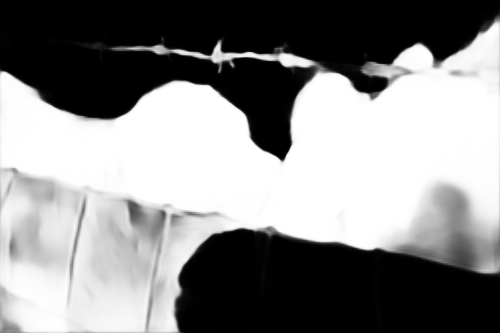}\vspace{1.2pt}
	\end{minipage}}
	
	\caption{Visual comparisons of saliency maps generated by our proposed approach and other state-of-the-art approaches.}
	\label{fig:visual_comparison}
\end{figure*}

\subsubsection{Evaluation Metrics} 
Multiple quantitative metrics are utilized to evaluate the detection performance of models.

\textbf{F-measure} ($F_{\beta}$) is employed to comprehensively balance the importance of precision and recall. We calculate the maximum and average values of F-measure, which are defined as $F^{m}_{\beta}$ and $F^{a}_{\beta}$, respectively. Besides, we report \textbf{weighted F-measure} ($F_{\beta}^{w}$)~\cite{margolin2014evaluate}. \textbf{Mean Absolute Error} ($\mathcal{M}$) directly calculates the pixel-wise average difference between the saliency map and the ground truth. We also use the common \textbf{S-measure} ($S_m$)~\cite{fan2017structure} and \textbf{E-measure} ($E_m$)~\cite{fan2018enhanced} to evaluate the performance.

\subsubsection{Implementation Details} 
We apply the Pytorch~\cite{paszke2019pytorch} framework to build our model. Following most previous work, we utilize the ResNet50 to initialize the four feature layers of the backbone. All $3\times3$ convolution layers in the whole network except for the backbone are followed by group normalization~\cite{wu2018group} and PReLU~\cite{he2015delving}. We employ the training set of DUTS to train our model. The resolutions of original input images are resized to $352\times352$. We set $\mu _R=3$ and $\mu _B=0.1$ in Eqs.~\ref{eq:sal} and~\ref{eq:jsl}, respectively. For data augmentation, the training samples are randomly rotated and flipped horizontally. During training, the batch size is set to 32 and Adam optimizer is used. We set the initial learning rate to 4e-5 and adopt the cosine annealing strategy~\cite{he2019bag} to adjust it. Network training is stopped after 50 epochs. All the experiments are conducted on a PC with one NVIDIA TESLA V-100 GPU.

\subsection{Comparison with State-of-the-arts}
We compare our detection model with recent 21 state-of-the-art methods. These methods are CPD~\cite{wu2019cascaded}, PoolN~\cite{liu2019simple}, BAS~\cite{qin2019basnet}, BAN~\cite{su2019selectivity}, SCRN~\cite{wu2019stacked}, EGN~\cite{zhao2019egnet}, F3N~\cite{wei2020f3net}, GCPA~\cite{chen2020global}, ITSD~\cite{zhou2020interactive}, MIN~\cite{pang2020multi}, LDF~\cite{wei2020label}, GateN~\cite{zhao2020suppress}, VST~\cite{liu2021visual}, MSFN~\cite{zhang2021auto}, PSG~\cite{yang2021progressive}, PurNet~\cite{li2021salient}, DCN~\cite{wu2021decomposition}, ICON~\cite{zhuge2022salient}, EDN~\cite{wu2022edn}, RCSBNet~\cite{ke2022recursive}, and ACCoNet~\cite{li2023adjacent}. For ACCoNet, we retrain it with publicly available code and the DUTS dataset to make it applicable for salient object detection in natural images. Note that our model outputs saliency maps in multiple scales. When evaluating the performance of our model, we only measure the quality of the saliency maps corresponding to the original-size images.

\subsubsection{Quantitative Evaluation} 
We record the scores for $F^{m}_{\beta}$, $F^{a}_{\beta}$, $F^w_\beta$, $S_m$, $E_m$, and $\mathcal{M}$ in Table~\ref{Scores1}. Moreover, the parameters and inference efficiency of each model are also shown in the tables. We observe that MIMONet achieves better evaluation scores than the competing methods in most cases. As for the remaining cases, the performance of MIMONet is also very close to the best performance. Overall, MIMONet has the lowest $\mathcal{M}$ scores on most datasets, proving that its saliency maps are more similar to the ground truths. It is worth noting that MIMONet has less than 30M parameters, which is a smaller number of parameters compared to most models. Therefore, our approach does not rely on a larger model to improve performance. Although MIMONet processes three different resolution images simultaneously, it still has real-time inference speed. These comparison results highlight the simplicity and practicability of MIMONet.

\subsubsection{Qualitative Evaluation} 
We list some visual comparisons of saliency maps to show the advantages of our method, as shown in Fig.~\ref{fig:visual_comparison}. These scenes contain salient objects of different sizes, including small (rows 1 to 3), middle (rows 4 to 6) and large (rows 7 to 9) objects. Compared to other methods, the saliency maps generated by MIMONet are clearer and have obvious suppression effect on the background, indicating that its anti-interference ability is stronger. When the size of the salient objects is relatively small, most competitors fail to identify them accurately and lack sufficient perception of them. When the salient regions become larger, MIMONet more completely segments them out, while the saliency maps of other methods are missing some parts. In some other cases, such as low contrast (the 3th row), complex scenes (the 4th row), and cluttered background (the 5th row), the detection effect of MIMONet is also better than that of competitors.

\subsection{Ablation Study}
\subsubsection{Effectiveness of MIMONet}
In this paper, we propose a network that takes multi-scale images as inputs and outputs their corresponding saliency maps. Compared with the current mainstream SISO framework, MIMONet has a more powerful inference capability thanks to capturing information on object size variations. Here, we analyze the construction of MIMONet to highlight its effectiveness. Therefore, we set up multiple network architectures to compare with MIMONet. Specifically, we change the number of inputs or outputs of MIMONet to build different networks, where the number of inputs is denoted by $N$. When $N=1$, a SISO network is constructed. Under this condition, we record the detection results for three different resolution images, namely low-, original-, and high-resolution images. Note that in the training stage of the SISO network, we directly treat $S^{mid}$ as $S^{c}$ and subsequently still calculate the loss of Eq.~\ref{eq:jsl}. $N=2$ means that the network has two images of different sizes as inputs. There are two cases in this condition. On the one hand, the original size image as well as the low-resolution image are used as inputs. On the other hand, the high-resolution image replaces the low-resolution image as another input. When $N=3$, we also make multiple settings. i) We explore the effect of scaling factor settings and apply multiple sets of factors for experimental comparisons. The first set of factors includes 0.75 and 1.25. The second set has a wide range, consisting of 0.5 and 2.0. ii) We use three independent backbones to extract multi-level features of each scale image separately, i.e., the backbone network does not share parameters. iii) The network only outputs the saliency map for the original resolution image, which becomes the multi-scale input and single-scale output (MISO) architecture. In this architecture, the inference operations on both low- and high-resolution images are omitted. Specifically, two branches for output $S^\text {low}$ and $S^\text {high}$ are discarded during the decoding stage. Finally, we make $N=5$, i.e., more images of different sizes are used as inputs, where the scaling factors are 0.5, 0.75, 1.25, and 1.5. The evaluation results of all the above settings are shown in Table~\ref{table:network}.

\begin{table*}  [htbp]
	\begin{center}
		\caption{Performance comparison of network architectures. $N$ represents the number of input images. ``Original''indicates an original resolution image. ``Original+Low'' means that the network inputs contain the original resolution image and the low resolution image. ``Original+High'' is defined in the same way. Note that except ``Original'' and MISO, other architectures adopt the MIMO framework. The best scores are marked in bold}
		\label{table:network}
		\resizebox{1.0\linewidth}{!}{%
			\begin{tabular}{c|cccccc|cccccc|cccccc}
				\Xhline{1pt}
				\multirow{2}{*}{Setting} & \multicolumn{6}{c|}{PASCAL-S}& \multicolumn{6}{c|}{DUT-OMRON} & \multicolumn{6}{c}{SOC-test}  \\
				
				\cline{2-19}
				& $F^{m}_\beta\uparrow$ & $F^{a}_\beta\uparrow$ & $F^{w}_\beta\uparrow$  & $S_m\uparrow$ & $E_m\uparrow$ & $\mathcal{M}\downarrow$
			& $F^{m}_\beta\uparrow$ & $F^{a}_\beta\uparrow$ & $F^{w}_\beta\uparrow$  & $S_m\uparrow$ & $E_m\uparrow$ & $\mathcal{M}\downarrow$
			& $F^{m}_\beta\uparrow$ & $F^{a}_\beta\uparrow$ & $F^{w}_\beta\uparrow$  & $S_m\uparrow$ & $E_m\uparrow$ & $\mathcal{M}\downarrow$ \\
				\Xhline{1pt}
                \multicolumn{19}{c}{$N=1$} \\
				\Xhline{1pt}

                Low & 
				0.867 & 0.811 & 0.805 & 0.849 & 0.888 & 0.065 & 0.797 & 0.736 & 0.725 & 0.821 & 0.866 & 0.060 & 0.751 & 0.663 & 0.651 & 0.759 & 0.802 & 0.098 \\
    
				Original & 
				0.881 & 0.835 & 0.826 & 0.861 & 0.897 & 0.061 & 0.802 & 0.739 & 0.729 & 0.823 & 0.861 & 0.063 & 0.766 & 0.693 & 0.669 & 0.768 & 0.811 & 0.096 \\

                High & 
				0.880 & 0.840 & 0.823 & 0.859 & 0.900 & 0.062 & 0.802 & 0.741 & 0.727 & 0.821 & 0.861 & 0.063 & 0.771 & 0.700 & 0.671 & 0.772 & 0.813 & 0.095 \\

                \Xhline{1pt}
                \multicolumn{19}{c}{$N=2$} \\
				\Xhline{1pt}

				Original+Low &
				0.887 & 0.839 & 0.833 & 0.866 & 0.903 & 0.057 & 0.811 & 0.756 & 0.745 & 0.832 & 0.871 & 0.059 & 0.776 & 0.706 & 0.680  & 0.777 & 0.819 & 0.092 \\

				Original+High & 
				0.889 & 0.844 & 0.832 & 0.865 & 0.905 & 0.058 & 0.815 & 0.760  & 0.749 & 0.834 & 0.873 & 0.058 & 0.774 & 0.706 & 0.685 & 0.780  & 0.826 & 0.090 \\

                \Xhline{1pt}
                \multicolumn{19}{c}{$N=3$} \\
				\Xhline{1pt}
				Factors$=0.75, 1.25$ & 
				\textbf{0.890} & 0.854 & 0.841 & \textbf{0.869} & 0.913 & \textbf{0.054} & \textbf{0.825} & 0.782 & 0.768 & \textbf{0.845} & 0.887 & 0.052 & 0.787 & 0.727 & 0.700   & 0.788 & 0.833 & 0.088 \\

                Factors$=0.5, 2$ & 
				0.888 & 0.849 & 0.838 & 0.870 & 0.909 &0.055 & 0.820 & 0.769 & 0.756 & 0.838 & 0.878 & 0.055 & 0.784 & 0.714 & 0.690   & 0.780 & 0.823 & 0.091 \\

                Independent backbones & 
				0.885 & 0.843 & 0.832 & 0.866 & 0.906 & 0.056 & 0.818 & 0.764 & 0.752 & 0.835 & 0.871 & 0.060 & 0.780  & 0.711 & 0.686 & 0.778 & 0.822 & 0.092 \\

                MISO & 
				0.886 & 0.840  & 0.830  & 0.866 & 0.904 & 0.058 & 0.815 & 0.755 & 0.745 & 0.833 & 0.868 & 0.060 & 0.773 & 0.699 & 0.676 & 0.774 & 0.811 & 0.094 \\

                MIMO (Ours) & 
				\textbf{0.890}  & \textbf{0.859} & \textbf{0.843} & \textbf{0.869} & 0.913 & \textbf{0.054} & \textbf{0.825} & \textbf{0.784} & \textbf{0.769} & \textbf{0.845} & \textbf{0.888} & \textbf{0.048} & 0.790  & 0.736 & 0.707 & 0.790  & \textbf{0.846} & \textbf{0.081} \\

				\Xhline{1pt}
                \multicolumn{19}{c}{$N=5$} \\
				\Xhline{1pt}
				Factors$=0.5, 0.75, 1.25, 1.5$ & 
				\textbf{0.890}  & 0.858 & \textbf{0.843} & \textbf{0.869} & \textbf{0.915} & \textbf{0.054} & 0.822 & 0.778 & 0.767 & 0.844 & 0.886 & 0.050 & \textbf{0.792} & \textbf{0.737} & \textbf{0.710}  & \textbf{0.792} & 0.843 & 0.082 \\
				
				\Xhline{1pt}

		\end{tabular} }
	\end{center}
\end{table*}

As can be seen from Table~\ref{table:network}, the detection performance of the network can be improved by adding images of different sizes as inputs, with reference to the case when $N>1$. Comparing the results for $N=2$ and $N=3$, we observe that the simultaneous use of low-, raw-, and high-resolution images as inputs enables the network to learn the size variation pattern of the objects more effectively, resulting in higher accuracy rates. We present some saliency maps in Fig.~\ref{fig:sample2} to visually compare the processing effects of these settings. When the number of inputs is 3, the model can better segment salient objects of various sizes using knowledge of size variation, and clearer saliency maps are generated. However, continuously increasing the number of input images does not guarantee to harvest better evaluation scores. When $N=5$, the performance obtained is worse than that at $N=3$ (MIMONet) in some cases. This is because more parameters are introduced into the network, which increases the difficulty of optimization, and the global optimal solution is not obtained. 

\begin{figure}[htbp]
	\centering
	\subfigure[Input]{
		\begin{minipage}[b]{0.115\linewidth}
			\includegraphics[width=1\linewidth]{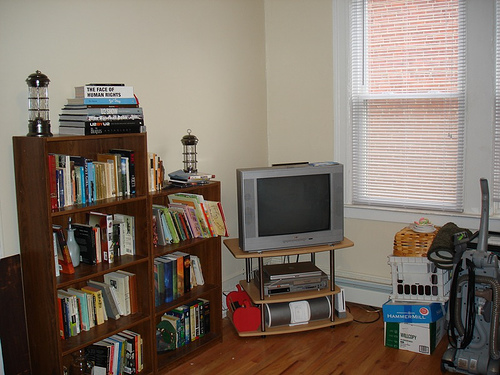}\vspace{1.2pt}
			\includegraphics[width=1\linewidth]{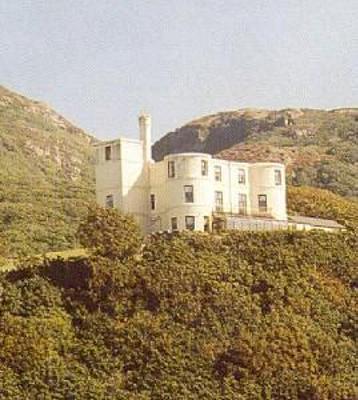}\vspace{1.2pt}
			\includegraphics[width=1\linewidth]{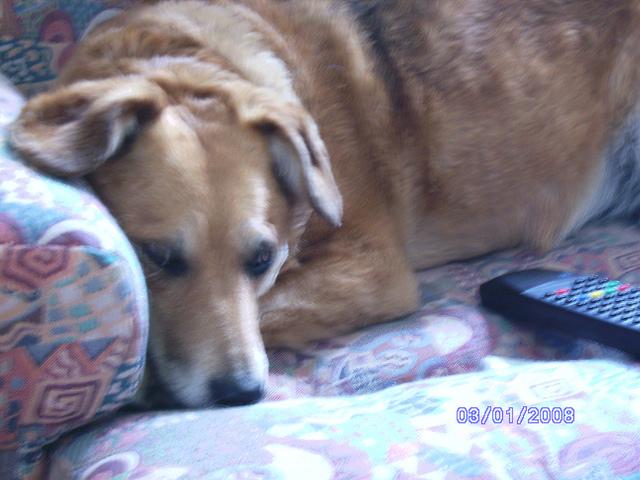}\vspace{1.2pt}
	\end{minipage}}
	\hspace{-2ex}
	\subfigure[GT]{
		\begin{minipage}[b]{0.115\linewidth}
			\includegraphics[width=1\linewidth]{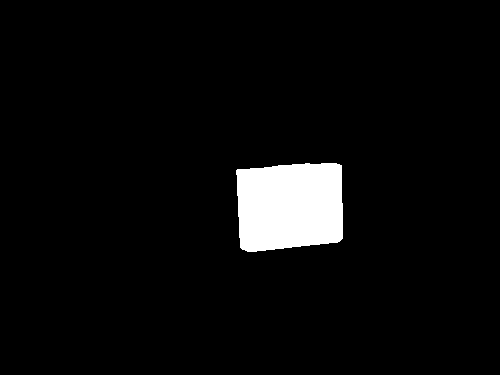}\vspace{1.2pt}
			\includegraphics[width=1\linewidth]{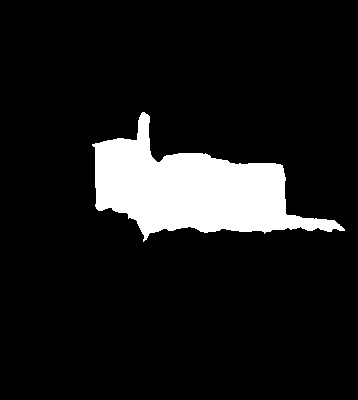}\vspace{1.2pt}
			\includegraphics[width=1\linewidth]{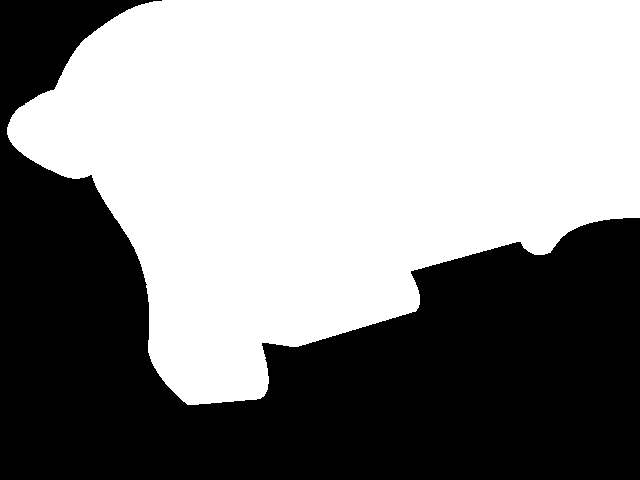}\vspace{1.2pt}
	\end{minipage}}
	\hspace{-2ex}
	\subfigure[L.O.H]{
		\begin{minipage}[b]{0.115\linewidth}
			\includegraphics[width=1\linewidth]{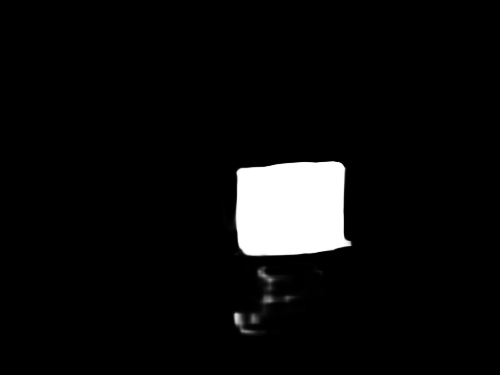}\vspace{1.2pt}
			\includegraphics[width=1\linewidth]{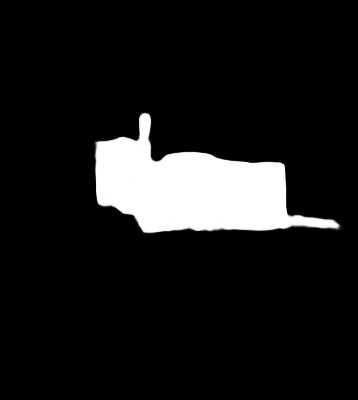}\vspace{1.2pt}
			\includegraphics[width=1\linewidth]{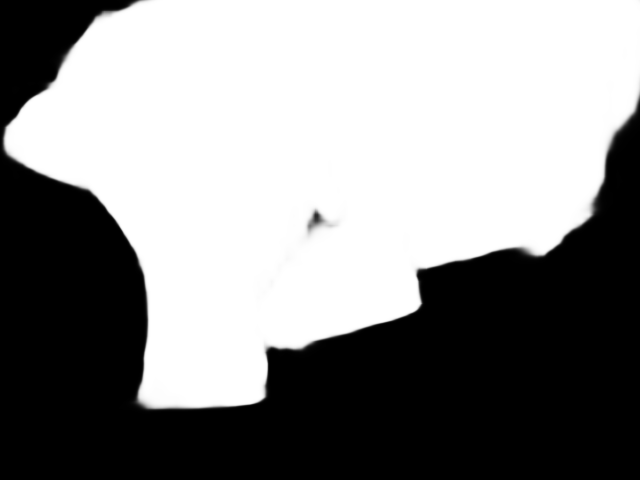}\vspace{1.2pt}
	\end{minipage}}
	\hspace{-2ex}
	\subfigure[O.H]{
		\begin{minipage}[b]{0.115\linewidth}
			\includegraphics[width=1\linewidth]{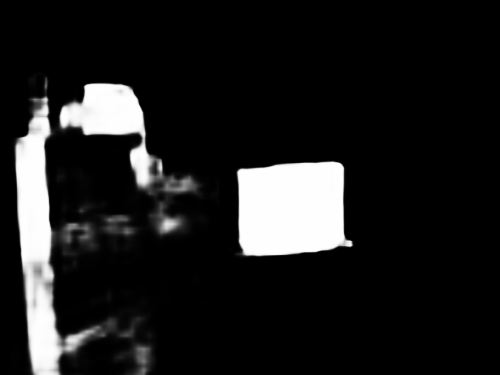}\vspace{1.2pt}
			\includegraphics[width=1\linewidth]{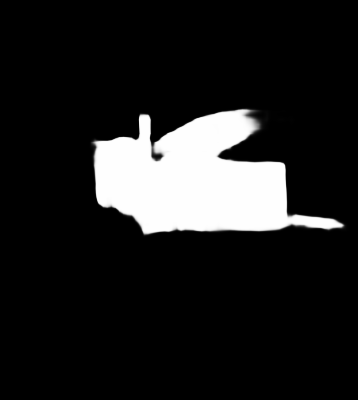}\vspace{1.2pt}
			\includegraphics[width=1\linewidth]{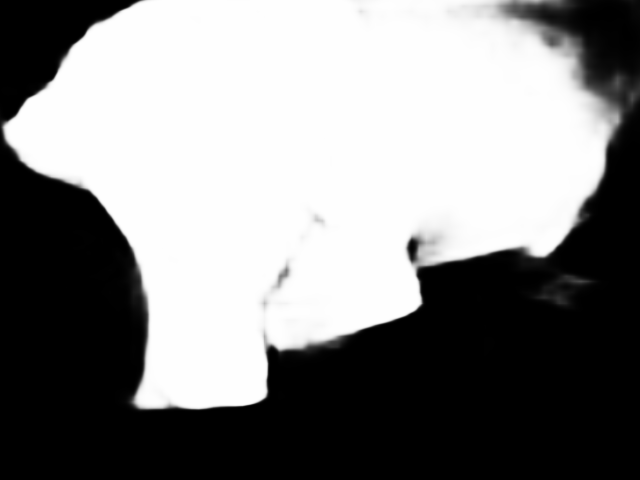}\vspace{1.2pt}
	\end{minipage}}
	\hspace{-2ex}
	\subfigure[L.O]{
		\begin{minipage}[b]{0.115\linewidth}
			\includegraphics[width=1\linewidth]{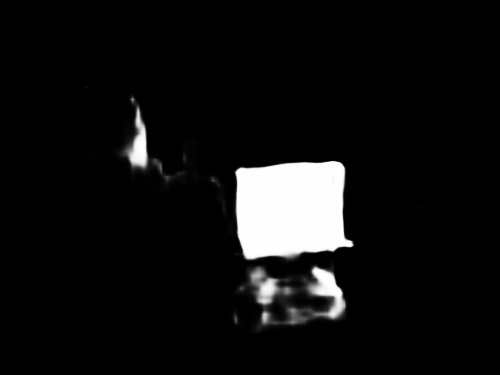}\vspace{1.2pt}
			\includegraphics[width=1\linewidth]{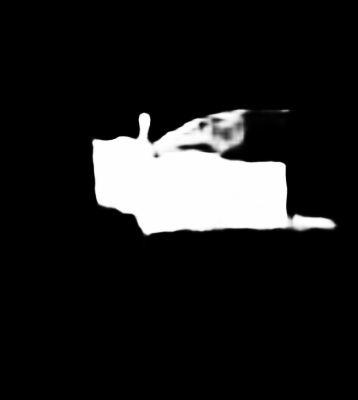}\vspace{1.2pt}
			\includegraphics[width=1\linewidth]{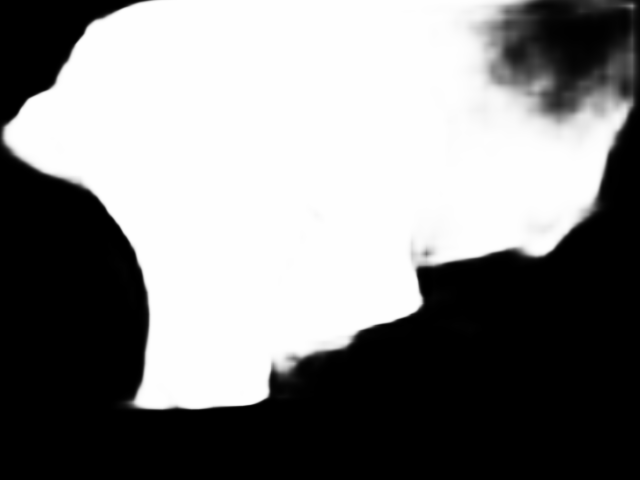}\vspace{1.2pt}
	\end{minipage}}
	\hspace{-2ex}
	\subfigure[H]{
		\begin{minipage}[b]{0.115\linewidth}
			\includegraphics[width=1\linewidth]{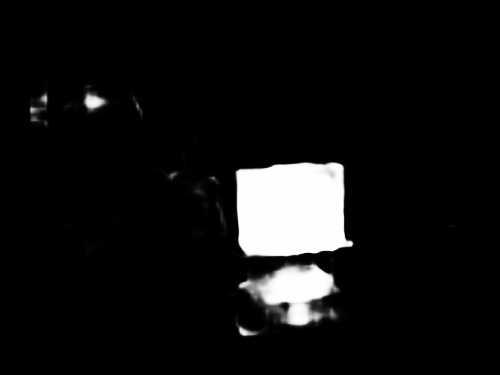}\vspace{1.2pt}
			\includegraphics[width=1\linewidth]{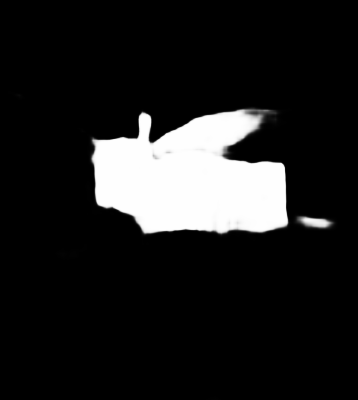}\vspace{1.2pt}
			\includegraphics[width=1\linewidth]{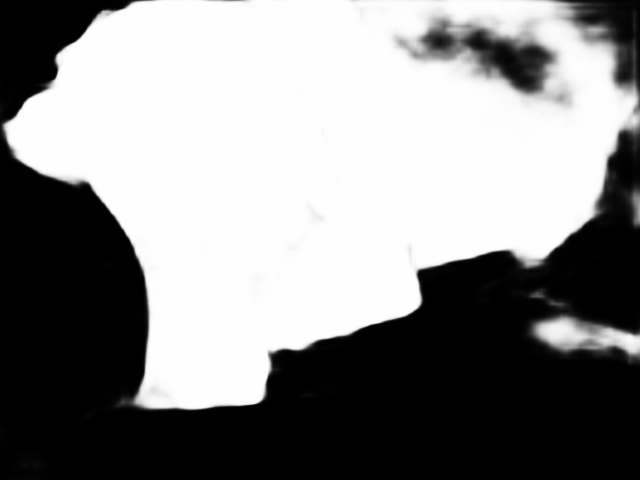}\vspace{1.2pt}
	\end{minipage}}
	\hspace{-2ex}
	\subfigure[O]{
		\begin{minipage}[b]{0.115\linewidth}
			\includegraphics[width=1\linewidth]{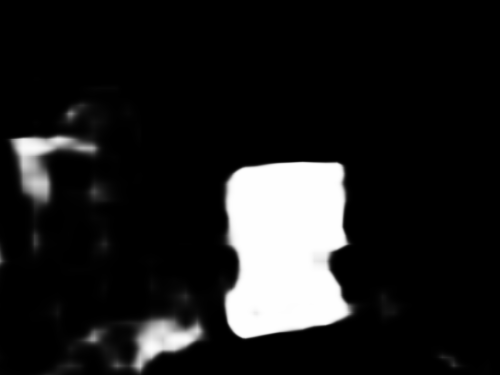}\vspace{1.2pt}
			\includegraphics[width=1\linewidth]{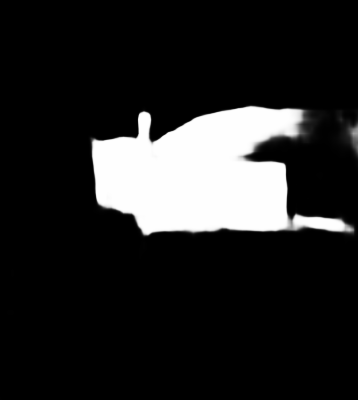}\vspace{1.2pt}
			\includegraphics[width=1\linewidth]{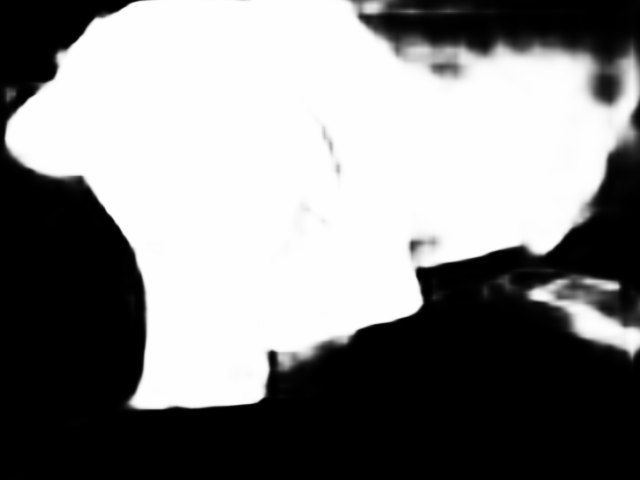}\vspace{1.2pt}
	\end{minipage}}
	\hspace{-2ex}
	\subfigure[L]{
		\begin{minipage}[b]{0.115\linewidth}
			\includegraphics[width=1\linewidth]{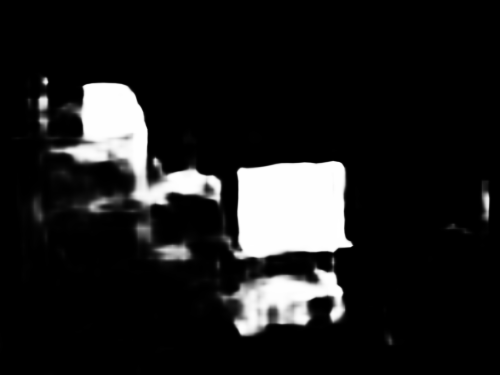}\vspace{1.2pt}
			\includegraphics[width=1\linewidth]{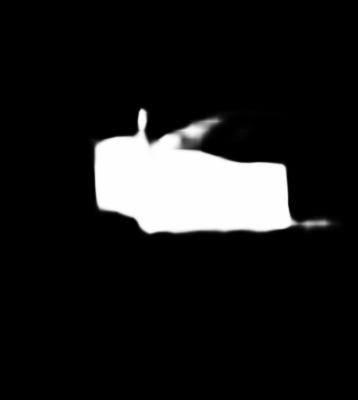}\vspace{1.2pt}
			\includegraphics[width=1\linewidth]{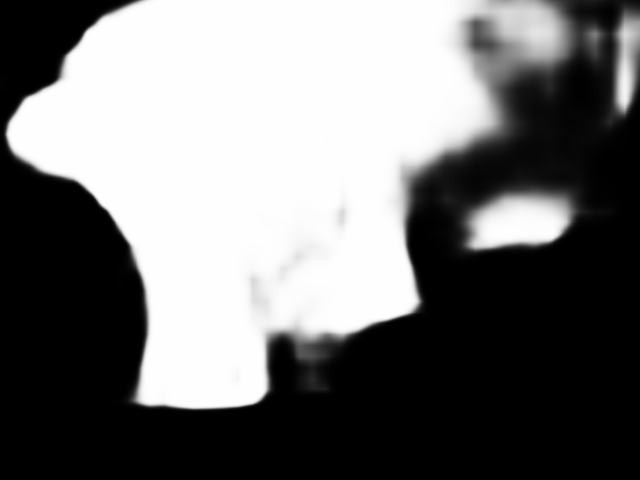}\vspace{1.2pt}
	\end{minipage}}
	\caption{Detection results produced by different combinations of input size and quantity. ``L.O.H'' represents a combination of low-, original- and high-resolution images. ``O.H'' indicates original- and high-resolution images are input. ``L.O'' means the inputs of the model are low- and original-resolution images. ``H'', ``O'' and ``L'' represent using only high, original, and low resolution images, respectively.}
	\label{fig:sample2}
\end{figure}

The setting of scaling factors is also a key point affecting performance. On the one hand, for the first set of scaling factors, their small range suggests a little change in the object size. This limited range condition may fail to provide sufficient useful knowledge of size variations for model learning. When the range of scaling factors is increased to a certain extent (scaling factors are 0.5 and 1.5), the size of the objects can change gradually and steadily from small to large, which is more suitable for the model to collect effective size change cues of the objects from multi-scale images. Therefore, the inference ability of the model is enhanced. On the other hand, after setting scaling factors with a wider range, the performance is not further improved, but rather weakened. The large span between the factors indicates a dramatic increase in the variation of the object size as well. The lack of a relatively stable scale change process makes it unfavorable for the model to accurately capture the change pattern of the objects.

Moreover, using multiple independent backbones not only increases the parameters and complexity of the network, but also reduces performance. Because the parameters among the backbone are not shared, the multi-level features of the three branches may not consistently express the information about the same objects, thus failing to exchange accurate object size cues. As for the MISO model, it does not output low- and high-resolution saliency maps. Without the supervisions for them, the network cannot effectively extract the features of small and large objects. Therefore, the performance of the MISO model is not as good as that of the MIMO model.

\begin{table*}  [htbp]
	\begin{center}
		\caption{Evaluation results of MSP and other alternatives. ``2:1:1'', ``1:1:2'' and ``1:2:1'' denote three different proportions for splitting the feature layer into low-, middle- and high-resolution sub-layers, respectively. The best scores are marked in bold}
		\label{table:MSP}
		\resizebox{1.0\linewidth}{!}{%
			\begin{tabular}{c|cccccc|cccccc|cccccc}
				\Xhline{1pt}
				\multirow{2}{*}{Setting} & \multicolumn{6}{c|}{PASCAL-S}& \multicolumn{6}{c|}{DUT-OMRON} & \multicolumn{6}{c}{SOC-test}  \\
				
				\cline{2-19}
				& $F^{m}_\beta\uparrow$ & $F^{a}_\beta\uparrow$ & $F^{w}_\beta\uparrow$  & $S_m\uparrow$ & $E_m\uparrow$ & $\mathcal{M}\downarrow$
			& $F^{m}_\beta\uparrow$ & $F^{a}_\beta\uparrow$ & $F^{w}_\beta\uparrow$  & $S_m\uparrow$ & $E_m\uparrow$ & $\mathcal{M}\downarrow$
			& $F^{m}_\beta\uparrow$ & $F^{a}_\beta\uparrow$ & $F^{w}_\beta\uparrow$  & $S_m\uparrow$ & $E_m\uparrow$ & $\mathcal{M}\downarrow$ \\
				\Xhline{1pt}
				2:1:1 & 
				0.888 & 0.843 & 0.834 & 0.868 & 0.906 & 0.057 & 0.815 & 0.756 & 0.745 & 0.833 & 0.868 & 0.060 & 0.777 & 0.700   & 0.678 & 0.775 & 0.814 & 0.095 \\
                1:1:2 &
                0.889 & 0.850  & 0.836 & \textbf{0.869} & 0.910  & 0.056 & 0.816 & 0.762 & 0.751 & 0.836 & 0.874 & 0.058 & 0.777 & 0.707 & 0.683 & 0.777 & 0.819 & 0.093 \\
                1:2:1 &
                \textbf{0.890}  & \textbf{0.859} & \textbf{0.843} & \textbf{0.869} & \textbf{0.913} & \textbf{0.054} & \textbf{0.825} & \textbf{0.784} & \textbf{0.769} & \textbf{0.845} & \textbf{0.888} & \textbf{0.048} & \textbf{0.790}  & \textbf{0.736} & \textbf{0.707} & \textbf{0.790}  & \textbf{0.846} & \textbf{0.081} \\
				\hline
                \hline
                Conv &
                0.886 & 0.845 & 0.833 & 0.867 & 0.906 & 0.056 & 0.813 & 0.754 & 0.744 & 0.833 & 0.871 & 0.058 & 0.771 & 0.702 & 0.677 & 0.774 & 0.815 & 0.093 \\
                PPM &
                0.888 & 0.846 & 0.834 & 0.867 & 0.905 & 0.058 & 0.819 & 0.761 & 0.750  & 0.837 & 0.874 & 0.057 & 0.779 & 0.703 & 0.684 & 0.779 & 0.815 & 0.092 \\
                ASPP &
                0.889 & 0.845 & 0.835 & 0.869 & 0.906 & 0.056 & 0.817 & 0.755 & 0.747 & 0.836 & 0.872 & 0.056 & 0.780  & 0.701 & 0.685 & 0.780  & 0.818 & 0.089 \\
                MSP &
                \textbf{0.890}  & \textbf{0.859} & \textbf{0.843} & \textbf{0.869} & \textbf{0.913} & \textbf{0.054} & \textbf{0.825} & \textbf{0.784} & \textbf{0.769} & \textbf{0.845} & \textbf{0.888} & \textbf{0.048} & \textbf{0.790}  & \textbf{0.736} & \textbf{0.707} & \textbf{0.790}  & \textbf{0.846} & \textbf{0.081} \\

                \hline
                \hline
                w/o ST &
                0.884 & 0.843 & 0.833 & 0.866 & 0.905 & 0.057 & 0.816 & 0.756 & 0.746 & 0.835 & 0.870 & 0.058 & 0.775 & 0.707 & 0.681 & 0.776 & 0.818 & 0.093 \\
                
				\Xhline{1pt}
		\end{tabular} }
	\end{center}
\end{table*}

\subsubsection{Importance of MSP}
The MSP module resizes the input feature layers in a hierarchical manner, aiming at making the feature layers capable of perceiving targets of different sizes simultaneously. Corresponding to Eq.~\ref{eq:msp}, in the MSP module, the input feature layer is divided into three sub-feature layers with different resolutions. In this case, the feature layer learns the object structure information of low, middle and high scales simultaneously, thus it has more powerful expression ability. Here, we explore the impact of split ratio on performance. We set three split ratios, including 2:1:1, 1:1:2, and 1:2:1. The results obtained from these settings are shown in Table~\ref{table:MSP}. The combination of 1:2:1 has more significant performance advantages than the other two combinations. Over-reliance on the learning of the other two resolution features may lead to insufficient representation of original size structural information. Splitting the feature layers in a ratio of 1:2:1 allows the network to allocate more computational resources and attention to original resolution features, while also absorbing the benefits of both low- and high-resolution features. This balance promotes the model to more effectively learn hierarchical structural information and improve its inference ability.

For the construction of MSP modules, we further investigate the impact of removing scale transformation (ST) operations (up-sampling and down-sampling), which means that all three branches process features at the original scale. The experimental results show that the detection performance becomes poor with the missing scale transformation step. The scale transformation simulates the potential scale variation of the objects, which is beneficial for the model to perceive small and large regions, thereby improving the its detection ability.

\begin{figure}[htbp]
	\centering
	\subfigure[Input]{
		\begin{minipage}[b]{0.162\linewidth}
			\includegraphics[width=1\linewidth]{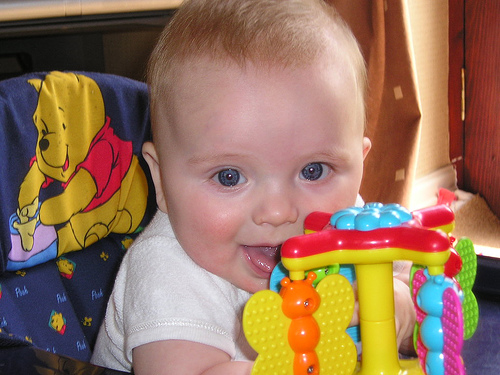}\vspace{1.2pt}
			\includegraphics[width=1\linewidth]{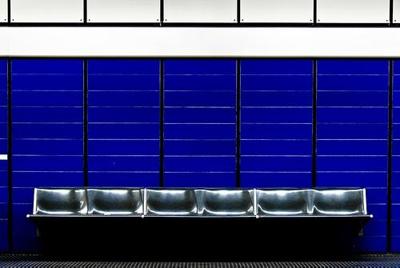}\vspace{1.2pt}
			\includegraphics[width=1\linewidth]{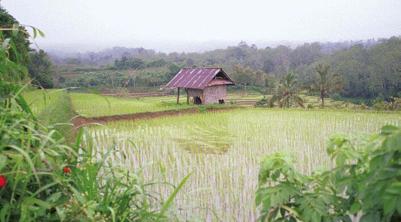}\vspace{1.2pt}
	\end{minipage}}
	\hspace{-2ex}
	\subfigure[GT]{
		\begin{minipage}[b]{0.162\linewidth}
			\includegraphics[width=1\linewidth]{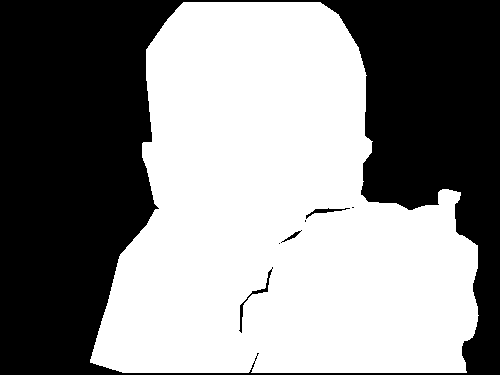}\vspace{1.2pt}
			\includegraphics[width=1\linewidth]{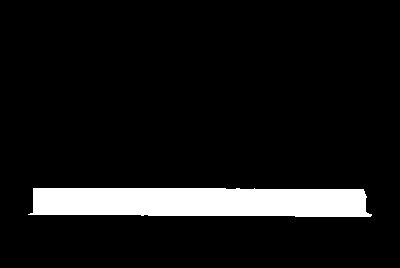}\vspace{1.2pt}
			\includegraphics[width=1\linewidth]{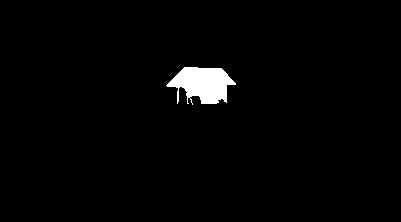}\vspace{1.2pt}
	\end{minipage}}
	\hspace{-2ex}
	\subfigure[MSP]{
		\begin{minipage}[b]{0.162\linewidth}
			\includegraphics[width=1\linewidth]{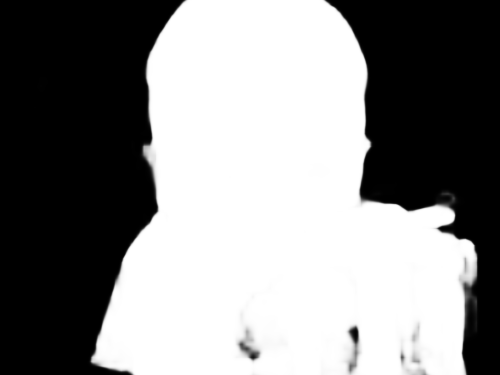}\vspace{1.2pt}
			\includegraphics[width=1\linewidth]{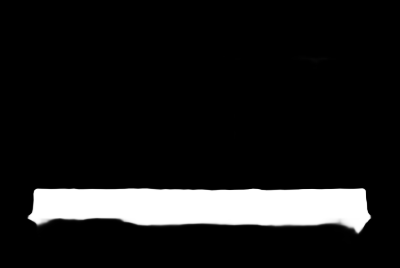}\vspace{1.2pt}
			\includegraphics[width=1\linewidth]{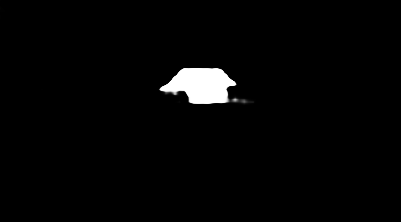}\vspace{1.2pt}
	\end{minipage}}
	\hspace{-2ex}
	\subfigure[ASPP]{
		\begin{minipage}[b]{0.162\linewidth}
			\includegraphics[width=1\linewidth]{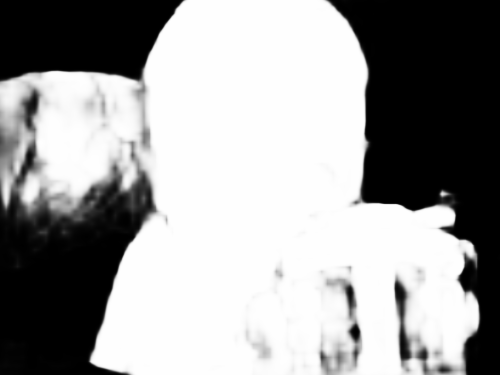}\vspace{1.2pt}
			\includegraphics[width=1\linewidth]{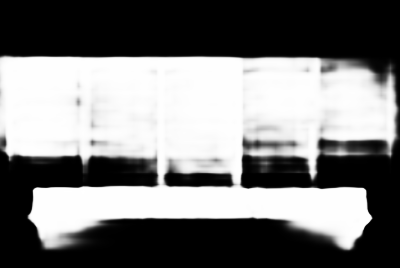}\vspace{1.2pt}
			\includegraphics[width=1\linewidth]{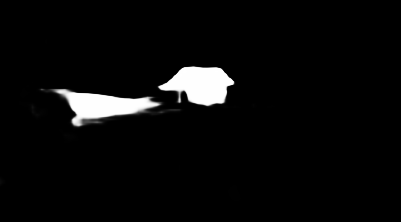}\vspace{1.2pt}
	\end{minipage}}
	\hspace{-2ex}
	\subfigure[PPM]{
		\begin{minipage}[b]{0.162\linewidth}
			\includegraphics[width=1\linewidth]{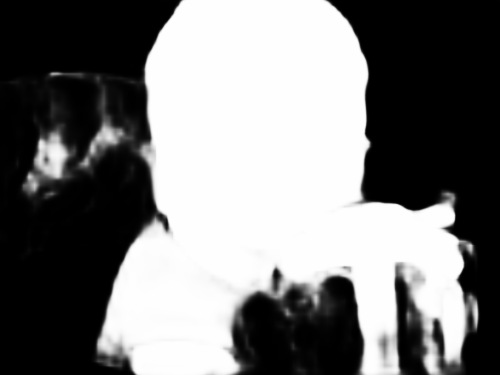}\vspace{1.2pt}
			\includegraphics[width=1\linewidth]{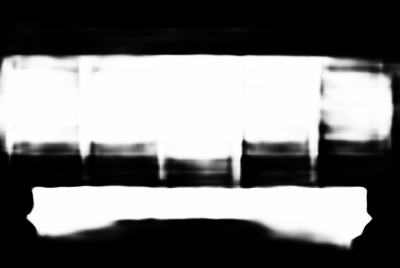}\vspace{1.2pt}
			\includegraphics[width=1\linewidth]{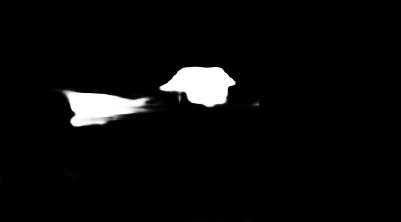}\vspace{1.2pt}
	\end{minipage}}
	\hspace{-2ex}
	\subfigure[Conv]{
		\begin{minipage}[b]{0.162\linewidth}
			\includegraphics[width=1\linewidth]{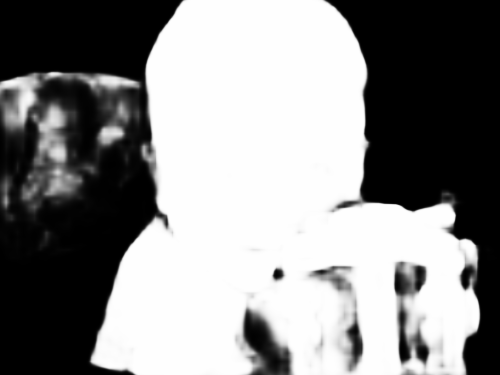}\vspace{1.2pt}
			\includegraphics[width=1\linewidth]{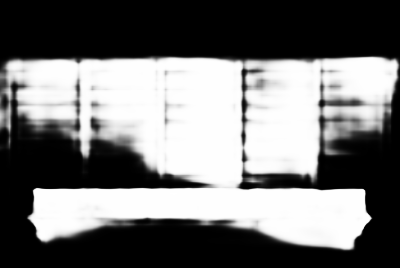}\vspace{1.2pt}
			\includegraphics[width=1\linewidth]{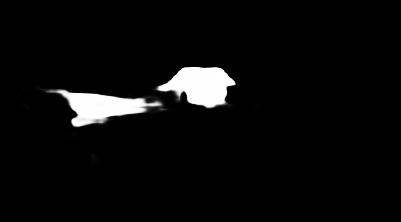}\vspace{1.2pt}
	\end{minipage}}
	\caption{Saliency maps resulting from MIMONet equipped with different modules.}
	\label{fig:msp}
\end{figure}

We also compare the MSP module with the common PPM~\cite{zhao2017pyramid} and ASPP~\cite{chen2018deeplab} modules. The PPM module mainly fuses the information of various down-sampled features. However, using only down-sampled operation tends to lose some detail information of the original feature layer. For the saliency detection task, the loss of detail information is detrimental to the recovery of region boundaries. The ASPP module uses different dilation rates to change the perceptual range of the convolution kernel without resizing the size of the feature layer. This operation tends to extract the structural information of fixed size regions. In contrast to them, the MSP module generates sub-feature layers with low-, middle- and high-resolutions through different scale transformations of feature layers, which helps to subsequently obtain multi-scale structural information of the objects. Therefore, MSP module can achieve better detection results in MIMONet than PPM and ASPP modules. In addition, we directly remove the whole MSP module and replace it with a $3\times3$ convolution (Conv) layer. From Table~\ref{table:MSP}, the detection capability of the network is severely weakened without the assistance of the MSP modules. Some detection results produced by the different modules are shown in Fig.~\ref{fig:msp}. We observe that the network equipped with the MSP modules outputs more accurate and smoother salient regions. This indicates that the MSP module plays an important role in the MIMONet.

\subsubsection{Contribution of JSL}
In JSL, we calculate the loss of boundary information. This approach not only constrains the multiple predictions to have the same salient regions, but also further maintains the region boundaries. Here, we use different structures to generate boundary maps and compare the performance of different structures. First, we remove JSL and record the results of training network without JSL. Then, we change the number of branches $R$ in the process of generating the boundary map, and set $R=1, 2, 3$ and 4 respectively. The evaluation results of these settings are shown in Table~\ref{table:JSL}. Comparing the first and second rows, we can see that adding JSL significantly improves the detection performance. Furthermore, the boundary maps generated by different structures are shown in Fig.~\ref{fig:en}. It can be seen that using more branches can generate more accurate boundary map, and thus obtain better evaluation scores than the case of a single branch. When $R=4$, the boundary structures of the salient regions are more accurately portrayed. By increasing the number of branches, more branches can be used to collect diverse structural information simultaneously. The joint constraint of these structural information provides the model with more effective support for identifying the object boundaries. After comparison, we set $R=4$ to get the maximum benefit.

\begin{table*}  [htbp]
	\begin{center}
		\caption{Performance analysis on JSL. $R$ indicates the number of branches required to generate the common boundaries. The best scores are marked in bold}
		\label{table:JSL}
		\resizebox{1.0\linewidth}{!}{%
			\begin{tabular}{c|cccccc|cccccc|cccccc}
				\Xhline{1pt}
				\multirow{2}{*}{Setting} & \multicolumn{6}{c|}{PASCAL-S}& \multicolumn{6}{c|}{DUT-OMRON} & \multicolumn{6}{c}{SOC-test}  \\
				
				\cline{2-19}
				& $F^{m}_\beta\uparrow$ & $F^{a}_\beta\uparrow$ & $F^{w}_\beta\uparrow$  & $S_m\uparrow$ & $E_m\uparrow$ & $\mathcal{M}\downarrow$
			& $F^{m}_\beta\uparrow$ & $F^{a}_\beta\uparrow$ & $F^{w}_\beta\uparrow$  & $S_m\uparrow$ & $E_m\uparrow$ & $\mathcal{M}\downarrow$
			& $F^{m}_\beta\uparrow$ & $F^{a}_\beta\uparrow$ & $F^{w}_\beta\uparrow$  & $S_m\uparrow$ & $E_m\uparrow$ & $\mathcal{M}\downarrow$ \\
				\Xhline{1pt}
				
				w/o JSL & 
				0.886 & 0.839 & 0.833 & 0.866 & 0.901 & 0.057 & 0.812 & 0.754 & 0.743 & 0.832 & 0.867 & 0.060 & 0.774 & 0.702 & 0.677 & 0.775 & 0.815 & 0.093 \\
                w/ JSL &
                \textbf{0.890}  & \textbf{0.859} & \textbf{0.843} & \textbf{0.869} & \textbf{0.913} & \textbf{0.054} & \textbf{0.825} & \textbf{0.784} & \textbf{0.769} & \textbf{0.845} & \textbf{0.888} & \textbf{0.048} & \textbf{0.790}  & \textbf{0.736} & \textbf{0.707} & \textbf{0.790}  & \textbf{0.846} & \textbf{0.081} \\
                
				\hline
                \hline
                $R=1$ &
                0.886 & 0.841 & 0.832 & 0.866 & 0.903 & 0.058 & 0.815 & 0.755 & 0.745 & 0.834 & 0.868 & 0.060 & 0.777 & 0.703 & 0.684 & 0.779 & 0.815 & 0.092 \\
                $R=2$ &
                0.886 & 0.844 & 0.832 & 0.867 & 0.905 & 0.057 & 0.816 & 0.759 & 0.750  & 0.836 & 0.871 & 0.058 & 0.779 & 0.705 & 0.682 & 0.777 & 0.817 & 0.093 \\
                $R=3$ &
                0.887 & 0.851 & 0.838 & 0.867 & 0.910  & 0.055 & 0.820  & 0.774 & 0.763 & 0.843 & 0.880  & 0.052 & 0.785 & 0.728 & 0.702 & 0.787 & 0.837 & 0.084 \\
                $R=4$ &
                \textbf{0.890}  & \textbf{0.859} & \textbf{0.843} & \textbf{0.869} & \textbf{0.913} & \textbf{0.054} & \textbf{0.825} & \textbf{0.784} & \textbf{0.769} & \textbf{0.845} & \textbf{0.888} & \textbf{0.048} & \textbf{0.790}  & \textbf{0.736} & \textbf{0.707} & \textbf{0.790}  & \textbf{0.846} & \textbf{0.081} \\
				
				\Xhline{1pt}
		\end{tabular} }
	\end{center}
\end{table*}

\begin{figure}[htbp]
	\centering
	\subfigure[Image]{
		\begin{minipage}[b]{0.162\linewidth}
			\includegraphics[width=1\linewidth]{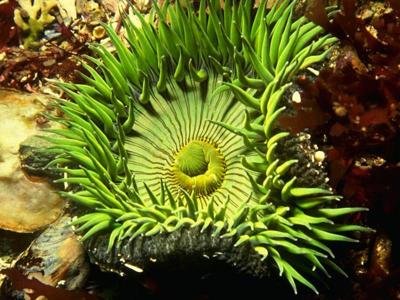}\vspace{1.2pt}
			\includegraphics[width=1\linewidth]{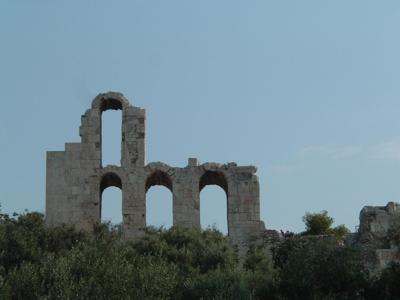}\vspace{1.2pt}
            \includegraphics[width=1\linewidth]{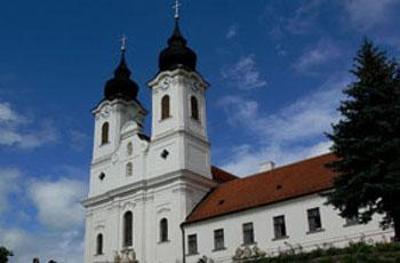}\vspace{1.2pt}
	\end{minipage}}
	\hspace{-2ex}
	\subfigure[$R=1$]{
		\begin{minipage}[b]{0.162\linewidth}
			\includegraphics[width=1\linewidth]{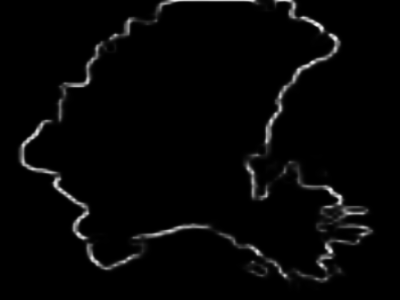}\vspace{1.2pt}
			\includegraphics[width=1\linewidth]{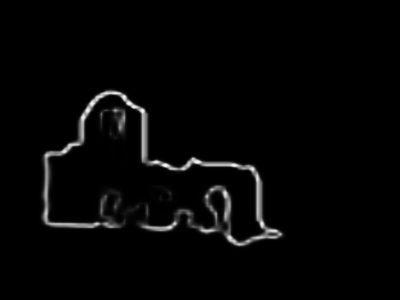}\vspace{1.2pt}
            \includegraphics[width=1\linewidth]{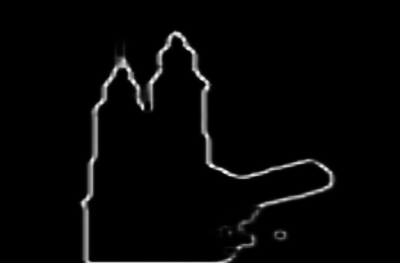}\vspace{1.2pt}
	\end{minipage}}
	\hspace{-2ex}
	\subfigure[$R=2$]{
		\begin{minipage}[b]{0.162\linewidth}
			\includegraphics[width=1\linewidth]{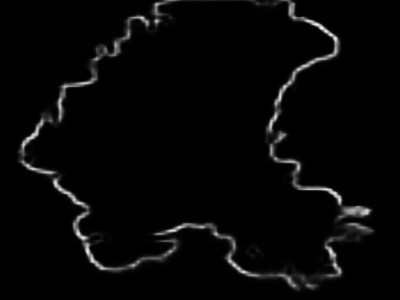}\vspace{1.2pt}
			\includegraphics[width=1\linewidth]{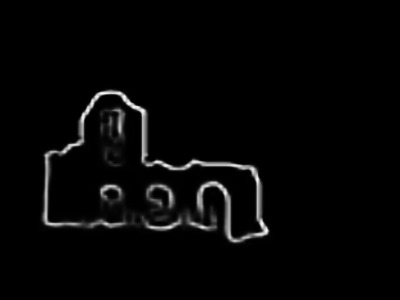}\vspace{1.2pt}
            \includegraphics[width=1\linewidth]{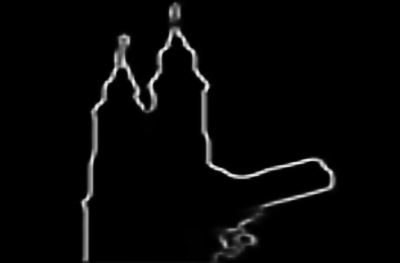}\vspace{1.2pt}
	\end{minipage}}
	\hspace{-2ex}
	\subfigure[$R=3$]{
		\begin{minipage}[b]{0.162\linewidth}
			\includegraphics[width=1\linewidth]{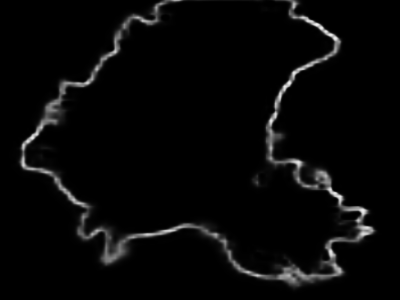}\vspace{1.2pt}
			\includegraphics[width=1\linewidth]{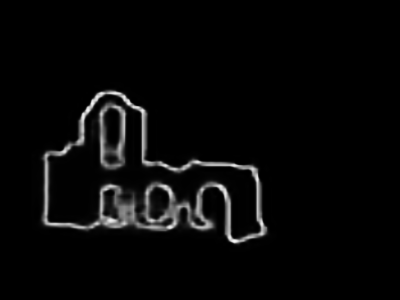}\vspace{1.2pt}
            \includegraphics[width=1\linewidth]{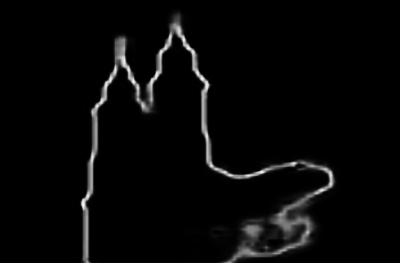}\vspace{1.2pt}
	\end{minipage}}
	\hspace{-2ex}
	\subfigure[$R=4$]{
		\begin{minipage}[b]{0.162\linewidth}
			\includegraphics[width=1\linewidth]{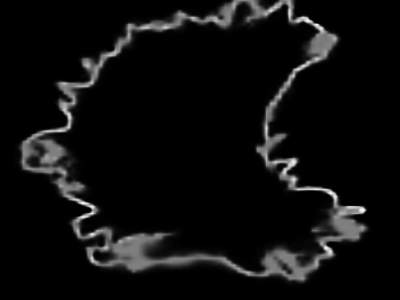}\vspace{1.2pt}
			\includegraphics[width=1\linewidth]{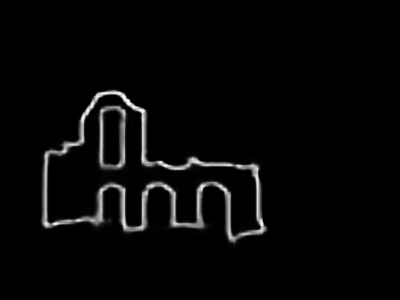}\vspace{1.2pt}
            \includegraphics[width=1\linewidth]{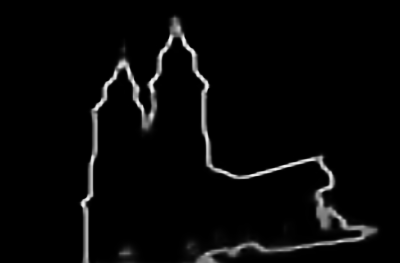}\vspace{1.2pt}
	\end{minipage}}
	\hspace{-2ex}
	\subfigure[GT]{
		\begin{minipage}[b]{0.162\linewidth}
			\includegraphics[width=1\linewidth]{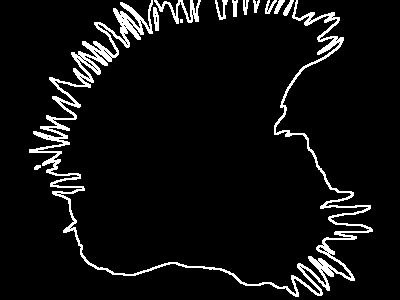}\vspace{1.2pt}
			\includegraphics[width=1\linewidth]{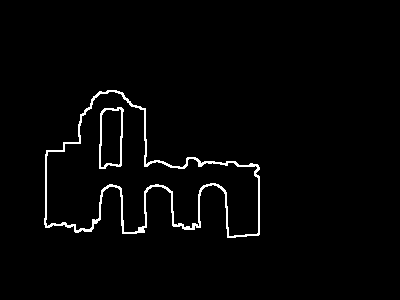}\vspace{1.2pt}
            \includegraphics[width=1\linewidth]{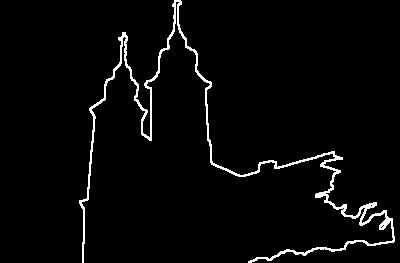}\vspace{1.2pt}
	\end{minipage}}
	\caption{The boundary effects generated by different branch structures.}
	\label{fig:en}
\end{figure}

\section{Conclusion}
In this paper, we propose a multi-scale input and multi-scale output network (MIMONet) to further improve the accuracy of detecting various scale targets. On the one hand, the mode of inputting multi-scale image solves the information limitation of the original image, and provides convenience for understanding the rule of object size change. On the other hand, by outputting multi-scale saliency maps and supervising them, the network can more effectively extract precise common saliency cues from multi-scale images. Besides, we also design a multi-scale perception (MSP) module and embed several such modules into MIMONet. The MSP modules adopt scale transformations to capture the multi-scale structure information of the objects, which facilitates the network to discriminate more accurate targets. To make saliency maps with different resolutions further focus on the same salient regions simultaneously, we introduce a joint saliency loss (JSL) to train the network. With the help of JSL, the region boundaries can also be more clearly preserved. The experimental results show that MIMONet achieves state-ofthe-art performance on multiple datasets with relatively fewer parameters. In the future, we will focus on saliency detection of point cloud~\cite{fan2022salient} data. In addition to providing RGB information, point clouds also contain precise scene geometry information, which is of great help in the correct recognition of salient objects.


\bibliographystyle{IEEEtran}
\bibliography{mybibfile}

@inproceedings{hou2017deeply,
  title={Deeply supervised salient object detection with short connections},
  author={Hou, Qibin and Cheng, Ming-Ming and Hu, Xiaowei and Borji, Ali and Tu, Zhuowen and Torr, Philip HS},
  booktitle={Proceedings of the IEEE Conference on Computer Vision and Pattern Recognition},
  pages={3203--3212},
  year={2017}
}

@inproceedings{zhang2017amulet,
  title={Amulet: Aggregating multi-level convolutional features for salient object detection},
  author={Zhang, Pingping and Wang, Dong and Lu, Huchuan and Wang, Hongyu and Ruan, Xiang},
  booktitle={Proceedings of the IEEE International Conference on Computer Vision},
  pages={202--211},
  year={2017}
}

@inproceedings{zhang2018bi,
  title={A bi-directional message passing model for salient object detection},
  author={Zhang, Lu and Dai, Ju and Lu, Huchuan and He, You and Wang, Gang},
  booktitle={Proceedings of the IEEE Conference on Computer Vision and Pattern Recognition},
  pages={1741--1750},
  year={2018}
}

@inproceedings{yan2013hierarchical,
  title={Hierarchical saliency detection},
  author={Yan, Qiong and Xu, Li and Shi, Jianping and Jia, Jiaya},
  booktitle={Proceedings of the IEEE Conference on Computer Vision and Pattern Recognition},
  pages={1155--1162},
  year={2013}
}

@inproceedings{wang2017learning,
  title={Learning to detect salient objects with image-level supervision},
  author={Wang, Lijun and Lu, Huchuan and Wang, Yifan and Feng, Mengyang and Wang, Dong and Yin, Baocai and Ruan, Xiang},
  booktitle={Proceedings of the IEEE Conference on Computer Vision and Pattern Recognition},
  pages={136--145},
  year={2017}
}

@inproceedings{li2015visual,
  title={Visual saliency based on multiscale deep features},
  author={Li, Guanbin and Yu, Yizhou},
  booktitle={Proceedings of the IEEE conference on computer vision and pattern recognition},
  pages={5455--5463},
  year={2015}
}

@inproceedings{liu2018picanet,
  title={PiCANet: Learning pixel-wise contextual attention for saliency detection},
  author={Liu, Nian and Han, Junwei and Yang, Ming-Hsuan},
  booktitle={Proceedings of the IEEE Conference on Computer Vision and Pattern Recognition},
  pages={3089--3098},
  year={2018}
}

@inproceedings{luo2017non,
  title={Non-local deep features for salient object detection},
  author={Luo, Zhiming and Mishra, Akshaya and Achkar, Andrew and Eichel, Justin and Li, Shaozi and Jodoin, Pierre-Marc},
  booktitle={Proceedings of the IEEE Conference on Computer Vision and Pattern Recognition},
  pages={6609--6617},
  year={2017}
}

@inproceedings{fan2017structure,
  title={Structure-measure: A new way to evaluate foreground maps},
  author={Fan, Deng-Ping and Cheng, Ming-Ming and Liu, Yun and Li, Tao and Borji, Ali},
  booktitle={Proceedings of the IEEE international conference on computer vision},
  pages={4548--4557},
  year={2017}
}

@inproceedings{feng2019attentive,
  title={Attentive feedback network for boundary-aware salient object detection},
  author={Feng, Mengyang and Lu, Huchuan and Ding, Errui},
  booktitle={Proceedings of the IEEE Conference on Computer Vision and Pattern Recognition},
  pages={1623--1632},
  year={2019}
}

@inproceedings{li2014secrets,
  title={The secrets of salient object segmentation},
  author={Li, Yin and Hou, Xiaodi and Koch, Christof and Rehg, James M and Yuille, Alan L},
  booktitle={Proceedings of the IEEE Conference on Computer Vision and Pattern Recognition},
  pages={280--287},
  year={2014}
}

@inproceedings{hu18recurrently,
     author = {Hu, Xiaowei and Zhu, Lei and Qin, Jing and Fu, Chi-Wing and Heng, Pheng-Ann},
     title = {Recurrently Aggregating Deep Features for Salient Object Detection},
     booktitle = {AAAI},
     pages={6943--6950},
     year = {2018}
}

@inproceedings{wu2019cascaded,
  title={Cascaded partial decoder for fast and accurate salient object detection},
  author={Wu, Zhe and Su, Li and Huang, Qingming},
  booktitle={Proceedings of the IEEE Conference on Computer Vision and Pattern Recognition},
  pages={3907--3916},
  year={2019}
}

@inproceedings{yang2013saliency,
  title={Saliency detection via graph-based manifold ranking},
  author={Yang, Chuan and Zhang, Lihe and Lu, Huchuan and Ruan, Xiang and Yang, Ming-Hsuan},
  booktitle={Proceedings of the IEEE conference on computer vision and pattern recognition},
  pages={3166--3173},
  year={2013}
}

@inproceedings{wu2018group,
  title={Group normalization},
  author={Wu, Yuxin and He, Kaiming},
  booktitle={Proceedings of the European Conference on Computer Vision},
  pages={3--19},
  year={2018}
}

@inproceedings{he2015delving,
  title={Delving deep into rectifiers: Surpassing human-level performance on imagenet classification},
  author={He, Kaiming and Zhang, Xiangyu and Ren, Shaoqing and Sun, Jian},
  booktitle={Proceedings of the IEEE international conference on computer vision},
  pages={1026--1034},
  year={2015}
}

@inproceedings{su2019selectivity,
  title={Selectivity or invariance: Boundary-aware salient object detection},
  author={Su, Jinming and Li, Jia and Zhang, Yu and Xia, Changqun and Tian, Yonghong},
  booktitle={Proceedings of the IEEE International Conference on Computer Vision},
  pages={3799--3808},
  year={2019}
}

@inproceedings{qin2019basnet,
  title={Basnet: Boundary-aware salient object detection},
  author={Qin, Xuebin and Zhang, Zichen and Huang, Chenyang and Gao, Chao and Dehghan, Masood and Jagersand, Martin},
  booktitle={Proceedings of the IEEE Conference on Computer Vision and Pattern Recognition},
  pages={7479--7489},
  year={2019}
}

@inproceedings{paszke2019pytorch,
  title={PyTorch: An imperative style, high-performance deep learning library},
  author={Paszke, Adam and Gross, Sam and Massa, Francisco and Lerer, Adam and Bradbury, James and Chanan, Gregory and Killeen, Trevor and Lin, Zeming and Gimelshein, Natalia and Antiga, Luca and others},
  booktitle={Advances in Neural Information Processing Systems},
  pages={8024--8035},
  year={2019}
}

@inproceedings{liu2019simple,
  title={A simple pooling-based design for real-time salient object detection},
  author={Liu, Jiang-Jiang and Hou, Qibin and Cheng, Ming-Ming and Feng, Jiashi and Jiang, Jianmin},
  booktitle={Proceedings of the IEEE Conference on Computer Vision and Pattern Recognition},
  pages={3917--3926},
  year={2019}
}

@inproceedings{he2016deep,
  title={Deep residual learning for image recognition},
  author={He, Kaiming and Zhang, Xiangyu and Ren, Shaoqing and Sun, Jian},
  booktitle={Proceedings of the IEEE conference on computer vision and pattern recognition},
  pages={770--778},
  year={2016}
}

@inproceedings{fan2018enhanced,
  title={Enhanced-alignment measure for binary foreground map evaluation},
  author={Fan, Deng-Ping and Gong, Cheng and Cao, Yang and Ren, Bo and Cheng, Ming-Ming and Borji, Ali},
  booktitle={Proceedings of the 27th International Joint Conference on Artificial Intelligence},
  pages={698--704},
  year={2018}
}

@inproceedings{margolin2014evaluate,
  title={How to evaluate foreground maps?},
  author={Margolin, Ran and Zelnik-Manor, Lihi and Tal, Ayellet},
  booktitle={Proceedings of the IEEE conference on computer vision and pattern recognition},
  pages={248--255},
  year={2014}
}

@inproceedings{zhao2020suppress,
  title={Suppress and balance: A simple gated network for salient object detection},
  author={Zhao, Xiaoqi and Pang, Youwei and Zhang, Lihe and Lu, Huchuan and Zhang, Lei},
  booktitle={European Conference on Computer Vision},
  pages={35--51},
  year={2020},
  organization={Springer}
}

@inproceedings{chen2020global,
  title={Global context-aware progressive aggregation network for salient object detection},
  author={Chen, Zuyao and Xu, Qianqian and Cong, Runmin and Huang, Qingming},
  booktitle={Proceedings of the AAAI Conference on Artificial Intelligence},
  pages={10599--10606},
  year={2020}
}

@inproceedings{wei2020f3net,
  title={F$^3$Net: Fusion, Feedback and Focus for Salient Object Detection},
  author={Wei, Jun and Wang, Shuhui and Huang, Qingming},
  booktitle={Proceedings of the AAAI Conference on Artificial Intelligence},
  pages={12321--12328},
  year={2020}
}

@inproceedings{wu2019stacked,
  title={Stacked Cross Refinement Network for Edge-Aware Salient Object Detection},
  author={Wu, Zhe and Su, Li and Huang, Qingming},
  booktitle={Proceedings of the IEEE International Conference on Computer Vision},
  pages={7264--7273},
  year={2019}
}

@inproceedings{zhao2019egnet,
  title={EGNet: Edge guidance network for salient object detection},
  author={Zhao, Jia-Xing and Liu, Jiang-Jiang and Fan, Deng-Ping and Cao, Yang and Yang, Jufeng and Cheng, Ming-Ming},
  booktitle={Proceedings of the IEEE International Conference on Computer Vision},
  pages={8779--8788},
  year={2019}
}

@inproceedings{zhou2020interactive,
  title={Interactive Two-Stream Decoder for Accurate and Fast Saliency Detection},
  author={Zhou, Huajun and Xie, Xiaohua and Lai, Jian-Huang and Chen, Zixuan and Yang, Lingxiao},
  booktitle={Proceedings of the IEEE/CVF Conference on Computer Vision and Pattern Recognition},
  pages={9141--9150},
  year={2020}
}

@inproceedings{he2019bag,
  title={Bag of tricks for image classification with convolutional neural networks},
  author={He, Tong and Zhang, Zhi and Zhang, Hang and Zhang, Zhongyue and Xie, Junyuan and Li, Mu},
  booktitle={Proceedings of the IEEE Conference on Computer Vision and Pattern Recognition},
  pages={558--567},
  year={2019}
}

@inproceedings{pang2020multi,
  title={Multi-scale interactive network for salient object detection},
  author={Pang, Youwei and Zhao, Xiaoqi and Zhang, Lihe and Lu, Huchuan},
  booktitle={Proceedings of the IEEE/CVF Conference on Computer Vision and Pattern Recognition},
  pages={9413--9422},
  year={2020}
}

@inproceedings{wei2020label,
  title={Label decoupling framework for salient object detection},
  author={Wei, Jun and Wang, Shuhui and Wu, Zhe and Su, Chi and Huang, Qingming and Tian, Qi},
  booktitle={Proceedings of the IEEE/CVF Conference on Computer Vision and Pattern Recognition},
  pages={13025--13034},
  year={2020}
}

@INPROCEEDINGS{7298965, 
author={J. {Long} and E. {Shelhamer} and T. {Darrell}}, 
booktitle={2015 IEEE Conference on Computer Vision and Pattern Recognition}, 
title={Fully convolutional networks for semantic segmentation}, 
year={2015}, 
volume={},
number={}, 
pages={3431-3440},
}

@ARTICLE{liu2020dynamic,
  author={J. -J. {Liu} and Q. {Hou} and M. -M. {Cheng}},
  journal={IEEE Transactions on Image Processing}, 
  title={Dynamic Feature Integration for Simultaneous Detection of Salient Object, Edge, and Skeleton}, 
  year={2020},
  volume={29},
  number={},
  pages={8652-8667},}

@inproceedings{fan2018salient,
  title={Salient objects in clutter: Bringing salient object detection to the foreground},
  author={Fan, Deng-Ping and Cheng, Ming-Ming and Liu, Jiang-Jiang and Gao, Shang-Hua and Hou, Qibin and Borji, Ali},
  booktitle={European conference on computer vision},
  pages={186--202},
  year={2018}
}

@inproceedings{liu2021visual,
  title={Visual saliency transformer},
  author={Liu, Nian and Zhang, Ni and Wan, Kaiyuan and Shao, Ling and Han, Junwei},
  booktitle={Proceedings of the IEEE/CVF international conference on computer vision},
  pages={4722--4732},
  year={2021}
}

@inproceedings{zhang2021auto,
  title={Auto-msfnet: Search multi-scale fusion network for salient object detection},
  author={Zhang, Miao and Liu, Tingwei and Piao, Yongri and Yao, Shunyu and Lu, Huchuan},
  booktitle={Proceedings of the 29th ACM international conference on multimedia},
  pages={667--676},
  year={2021}
}

@article{wu2021decomposition,
  title={Decomposition and completion network for salient object detection},
  author={Wu, Zhe and Su, Li and Huang, Qingming},
  journal={IEEE Transactions on Image Processing},
  volume={30},
  pages={6226--6239},
  year={2021},
  publisher={IEEE}
}

@article{zhuge2022salient,
  title={Salient object detection via integrity learning},
  author={Zhuge, Mingchen and Fan, Deng-Ping and Liu, Nian and Zhang, Dingwen and Xu, Dong and Shao, Ling},
  journal={IEEE Transactions on Pattern Analysis and Machine Intelligence},
  year={2022},
  publisher={IEEE}
}

@article{wu2022edn,
  title={EDN: Salient object detection via extremely-downsampled network},
  author={Wu, Yu-Huan and Liu, Yun and Zhang, Le and Cheng, Ming-Ming and Ren, Bo},
  journal={IEEE Transactions on Image Processing},
  volume={31},
  pages={3125--3136},
  year={2022},
  publisher={IEEE}
}

@article{tu2022rgbt,
  title={RGBT salient object detection: A large-scale dataset and benchmark},
  author={Tu, Zhengzheng and Ma, Yan and Li, Zhun and Li, Chenglong and Xu, Jieming and Liu, Yongtao},
  journal={IEEE Transactions on Multimedia},
  year={2022},
  publisher={IEEE}
}

@inproceedings{zhao2017pyramid,
  title={Pyramid scene parsing network},
  author={Zhao, Hengshuang and Shi, Jianping and Qi, Xiaojuan and Wang, Xiaogang and Jia, Jiaya},
  booktitle={Proceedings of the IEEE conference on computer vision and pattern recognition},
  pages={2881--2890},
  year={2017}
}

@article{chen2018deeplab,
  title={Deeplab: Semantic image segmentation with deep convolutional nets, atrous convolution, and fully connected crfs},
  author={Chen, Liang-Chieh and Papandreou, George and Kokkinos, Iasonas and Murphy, Kevin and Yuille, Alan L},
  journal={IEEE transactions on pattern analysis and machine intelligence},
  volume={40},
  number={4},
  pages={834--848},
  year={2018},
  publisher={IEEE}
}

@article{liao2022cross,
  title={Cross-collaborative fusion-encoder network for robust RGB-thermal salient object detection},
  author={Liao, Guibiao and Gao, Wei and Li, Ge and Wang, Junle and Kwong, Sam},
  journal={IEEE Transactions on Circuits and Systems for Video Technology},
  volume={32},
  number={11},
  pages={7646--7661},
  year={2022},
  publisher={IEEE}
}

@article{gao2021unified,
  title={Unified information fusion network for multi-modal RGB-D and RGB-T salient object detection},
  author={Gao, Wei and Liao, Guibiao and Ma, Siwei and Li, Ge and Liang, Yongsheng and Lin, Weisi},
  journal={IEEE Transactions on Circuits and Systems for Video Technology},
  volume={32},
  number={4},
  pages={2091--2106},
  year={2021},
  publisher={IEEE}
}

@inproceedings{fan2022salient,
  title={Salient object detection for point clouds},
  author={Fan, Songlin and Gao, Wei and Li, Ge},
  booktitle={European Conference on Computer Vision},
  pages={1--19},
  year={2022},
  organization={Springer}
}

@article{li2021salient,
  title={Salient object detection with purificatory mechanism and structural similarity loss},
  author={Li, Jia and Su, Jinming and Xia, Changqun and Ma, Mingcan and Tian, Yonghong},
  journal={IEEE Transactions on Image Processing},
  volume={30},
  pages={6855--6868},
  year={2021},
  publisher={IEEE}
}

@article{yang2021progressive,
  title={Progressive self-guided loss for salient object detection},
  author={Yang, Sheng and Lin, Weisi and Lin, Guosheng and Jiang, Qiuping and Liu, Zichuan},
  journal={IEEE Transactions on Image Processing},
  volume={30},
  pages={8426--8438},
  year={2021},
  publisher={IEEE}
}

@article{li2023adjacent,
  title={Adjacent context coordination network for salient object detection in optical remote sensing images},
  author={Li, Gongyang and Liu, Zhi and Zeng, Dan and Lin, Weisi and Ling, Haibin},
  journal={IEEE Transactions on Cybernetics},
  volume={53},
  number={1},
  pages={526--538},
  year={2023},
  publisher={IEEE}
}

@article{fu2022siamese,
  title={Siamese network for RGB-D salient object detection and beyond},
  author={Fu, Keren and Fan, Deng-Ping and Ji, Ge-Peng and Zhao, Qijun and Shen, Jianbing and Zhu, Ce},
  journal={IEEE transactions on pattern analysis and machine intelligence},
  volume={44},
  number={9},
  pages={5541--5559},
  year={2022},
  publisher={IEEE}
}

@inproceedings{ke2022recursive,
  title={Recursive contour-saliency blending network for accurate salient object detection},
  author={Ke, Yun Yi and Tsubono, Takahiro},
  booktitle={Proceedings of the IEEE/CVF Winter Conference on Applications of Computer Vision},
  pages={2940--2950},
  year={2022}
}

@article{liu2022dna,
  title={DNA: Deeply supervised nonlinear aggregation for salient object detection},
  author={Liu, Yun and Cheng, Ming-Ming and Zhang, Xin-Yu and Nie, Guang-Yu and Wang, Meng},
  journal={IEEE Transactions on Cybernetics},
  volume={52},
  number={7},
  pages={6131--6142},
  year={2022},
  publisher={IEEE}
}

@article{zhou2023edge,
  title={Edge-guided recurrent positioning network for salient object detection in optical remote sensing images},
  author={Zhou, Xiaofei and Shen, Kunye and Weng, Li and Cong, Runmin and Zheng, Bolun and Zhang, Jiyong and Yan, Chenggang},
  journal={IEEE Transactions on Cybernetics},
  volume={53},
  number={1},
  pages={539--552},
  year={2023},
  publisher={IEEE}
}

@article{li2022complementarity,
  title={Complementarity-aware attention network for salient object detection},
  author={Li, Junxia and Pan, Zefeng and Liu, Qingshan and Cui, Ying and Sun, Yubao},
  journal={IEEE Transactions on Cybernetics},
  volume={52},
  number={2},
  pages={873--886},
  year={2022},
  publisher={IEEE}
}

\end{document}